\documentclass[11pt]{article}
\usepackage[table]{xcolor}

\usepackage[final]{acl}

\usepackage{times}
\usepackage{latexsym}

\usepackage[T1]{fontenc}

\usepackage[utf8]{inputenc}

\usepackage{microtype}

\usepackage{inconsolata}

\usepackage{graphicx}

\usepackage{longtable}
\usepackage{tabularx}
\usepackage{todonotes}
\usepackage{enumitem}
\usepackage{float}

\title{Do Persona-Infused LLMs Affect Performance in a Strategic Reasoning Game?}

\author{John Licato and Stephen Steinle \\ Bellini College of Artificial Intelligence,\\ Cybersecurity and Computing \\  University of South Florida
        \And
        Brayden Hollis \\ Information Directorate \\ Air Force Research Library}

\begin{document}
\maketitle
\begin{abstract}
    Although persona prompting in large language models appears to trigger different styles of generated text, it is unclear whether these translate into measurable behavioral differences, much less whether they affect decision-making in an adversarial strategic environment that we provide as open-source. We investigate the impact of persona prompting on strategic performance in PERIL, a world-domination board game. Specifically, we compare the effectiveness of persona-derived heuristic strategies to those chosen manually. Our findings reveal that certain personas associated with strategic thinking improve game performance, but only when a mediator is used to translate personas into heuristic values. We introduce this mediator as a structured translation process, inspired by exploratory factor analysis, that maps LLM-generated inventory responses into heuristics. Results indicate our method enhances heuristic reliability and face validity compared to directly inferred heuristics, allowing us to better study the effect of persona types on decision-making. These insights advance our understanding of how persona prompting influences LLM-based decision-making and propose a heuristic generation method that applies psychometric principles to LLMs.

\end{abstract}

\section{Introduction}

    ``If you would read a [person's] Disposition, see him Game, you will then learn more of him in one hour, than in seven Years Conversation,'' according to a letter of advice written over 300 years ago \cite{lingard1907letter}. If this advice is correct, perhaps nowhere is one's personality more apparent than in strategic adversarial games, where individual behavioral tendencies such as aggression, patience, caution, and others dictate the heuristics that guide players' decision-making. Such settings present a unique opportunity to study the relationship between how modern large language models (LLMs) relate personality descriptions (often called \textit{personas}) and decision-making in strategic environments.
    
    In this paper, we investigate whether personality traits inferred from prompts reliably translate into actionable heuristics in a strategy board game. Strategic reasoning is a critical capability for advancing AI in decision-making and human-machine collaboration. Beyond gaming, the findings have broader implications for simulation, training, and the development of automated systems requiring strategic adaptability. Such work contributes to advancing AI's role in team-based environments, military simulations, and other domains where human-like variability and strategic decision-making are essential. 

    \begin{figure}[t!]
        \centering
        \includegraphics[width=1\linewidth]{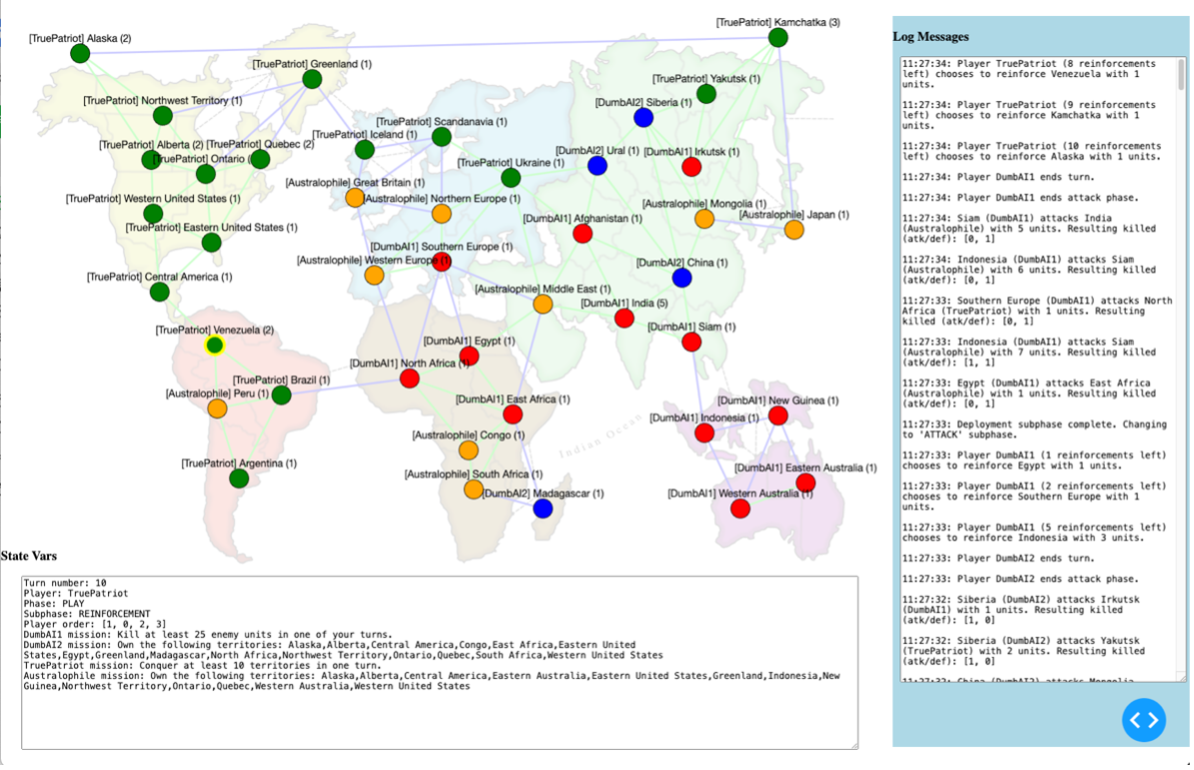}
        \caption{PERIL, our implementation inspired by a popular strategic world conquest board game, was used to study the effects of persona-based prompting in LLMs.}
        \label{fig:Risk}
    \end{figure}

    Isolating strategic reasoning ability is difficult, especially when that strategy must apply to an environment that has a large search space, nondeterministic outcomes, and high-pressure conditions that require rapid decision-making under time and computation constraints. In such environments, strategic actors may benefit from \textit{precommitment}, or the fixing of rules and heuristics that will constrain one's behavior ahead of time, so that cognitive overhead, behavioral inconsistencies, and the effect of time pressure on decisions will be minimized later. In this paper, we will focus on this aspect of strategic thinking by studying the performance of LLMs in a variant of a popular strategy board game. More specifically, we will study the effect of \textit{persona prompting}, a prompting strategy in which a pre-trained LLM is prompted with a description of a personality and asked to act in accordance with it. We do this through a fixed set of heuristics tailored to the PERIL environment, which should be understood as design choices rather than a comprehensive taxonomy of player attributes. Despite its promise, the effect of persona prompting on tasks requiring strategic reasoning, particularly in dynamic and uncertain environments, remains underexplored.

    We primarily address two research questions: \textbf{(1)} Does persona prompting using personalities with traits associated with strategic reasoning improve performance on strategic games? \textbf{(2)} Does using a personality inventory to translate persona descriptions into heuristics lead to decision-making heuristics with more face validity?
    
    \paragraph{Novel Contributions and Summary of Findings} 
    \begin{itemize}
        \item This is the first work to specifically study the effect of persona prompting on decision-making in a strategic reasoning game. We found a positive relationship between personality traits that intuitively would lead to better performance in the game and actual game performance, thus contributing to the ongoing research on how and when to use persona prompting. 
        \item We introduce PERIL, a new platform for evaluating strategic decision-making capabilities of AI players. In this paper, we compare the effects of persona prompting on players with the same mission, but the platform we implemented allows for multiple missions (which can in the future be used for studying strategic deception). We will make our full source code and platform available.
        \item We introduce the use of personality inventory questionnaires to translate personas into heuristic choices in an end-to-end fashion. We observe that this method results in heuristics that align with features of those personalities (much more so than when the questionnaire is not used). Without it the variation in how each persona translates into heuristics is small, suggesting that persona prompting alone does not lead to significant behavior differences.
    \end{itemize}

\section{Related Work}

    In recent years, pre-trained LLMs have become increasingly difficult to fine-tune, due to a combination of model size, computation requirements, and reduced access to pre-trained models' weights. As a result, many researchers have turned to strategies exploring the extent to which prompts can be adjusted to improve performance. An approach rapidly gaining popularity is based on the concept of the \textit{persona}, where a personality description is provided to the LLM, and it is asked to act in accordance with that personality \cite{Tseng2024,Zhang2024, bhandari2025can}. New frameworks are rapidly emerging to compare different persona prompts on a variety of tasks \cite{Pan2024,Lin2024,Samuel2024,Liu2024, poterti2025designing}, and datasets of high-quality LLM-generated personas are now available \cite{Schuller2024,Chan2024, wang2025opera}.

    However, it is unclear how well persona prompting can affect performance in action spaces. There are early results showing how LLMs can play a role in the command and conquer domain space \cite{Goecks2024} and in strategic board games \cite{meta2022human, Lore2024, hu2024surveylargelanguagemodelbased, belle2025agents}, but very little studying the effects of persona prompting. There are mixed results on its general effectiveness: \cite{Liu2024} used multiple personas to improve scientific ideation. \cite{Kamruzzaman2024} found that nationality-based persona prompting introduced preference biases towards the assigned country of origin. And other authors have found that multi-agent systems consisting of multiple personas working together can produce differences in reasoning, hallucination rate, and more \cite{Sreedhar2024,Wang2024,Olea2024,jiang2025harbor}.
    
    Strangely, not all recent work agrees that persona prompting has a significant effect, or even an effect that would be expected given the persona. For example, \cite{Hu2024} find that persona variables accounted for less than 10\% of annotation variance in subjective NLP datasets. \cite{Kim2024} found that role-playing prompts \textit{decreased} reasoning abilities in some datasets. And \cite{Zheng2024} find that on a subset of the MMLU benchmark tasks \cite{hendrycks2021measuring}, the use of personas does not seem to improve performance beyond random chance. Conflicting results can appear even in the same study, suggesting that persona prompting results are very sensitive to prompt structure \cite{phelps2025machine}. Further clouding the effective application of personas is their use in so-called psychological inventories \cite{li2025llm, wu2025personas}. While there is plenty of quality literature on the reliability of using personas in inventories \cite{bhandari2025evaluating, frisch2024llm}, there is significantly less work exploring the validity of such inventory results \cite{maharjan2025large}.    Another relevant line of work examines how persona prompting interacts with broader tendencies to anthropomorphize LLMs. Researchers have proposed multi-level frameworks describing how human-like qualities are attributed across perceptual, linguistic, and cognitive dimensions \cite{xiao2025humanizing}, as well as taxonomies of linguistic expressions that make language technologies appear more human-like \cite{devrio2025taxonomy}. These works address that part of the difficulty in interpreting persona effects may stem from how human expectations shape prompts. This is without even addressing the central concern of what theoretical foundation such results rest upon (even assuming reliability and validity) since the responses generated by LLMs do not conform to human distributions \cite{pratelli2025evaluating}.
    
    Given this gap in the literature, it is clear more work is needed to observe the effects of persona prompting in new scenarios, so that a deeper theory can be developed. Drawing on work showing that the use of personality questionnaires can improve LLM performance \cite{Bodroza2024,jiang-etal-2024-personallm},  we set out to study the extent to which persona-based prompting, augmented with personality questionnaires, can affect performance in a game of strategic reasoning, as well as gain deeper insight into how LLMs translate personas into decisions. As discussed in the previous paragraph, we do not make claims about the theory behind \textit{why} the LLM makes generations as it does, but rather to show that personas \textit{do} result in meaningful changes and that an exploratory factor analysis method bears fruit when applied to LLM reasoning.
    
\section{Experimental Setup}
        Our experimental design has four components: game environment, persona selection, heuristic generation, and tournament. The game environment introduces our PERIL software and the set of heuristics used to guide play. Persona selection explains how we sampled personas and identified the subset used in experiments. Heuristic generation describes how we constructed a questionnaire with GPT-4 and used it to map persona responses into heuristic values. Finally, tournament games outlines how matches were run and how performance metrics were collected.

\begin{table*}[ht!]
\footnotesize
\centering
\caption{Decision-Making Heuristics Available in PERIL}
\label{tbl:heuristics}

\begin{tabularx}{\textwidth}{|p{0.16\textwidth}|X|}
\hline
\rowcolor{gray} \multicolumn{2}{|c|}{\textbf{PHASE}} \\ \hline

\rowcolor{lightgray} \multicolumn{2}{|c|}{\textbf{Initialization / Deployment Phase}} \\ \hline
PTM/PTL & Place a unit in a region \( T_1 \) that is adjacent to region \( T_2 \) if \( T_2 \) is owned by the player with the most (PTM) or least (PTL) regions. \\ \hline
PUM/PUL & Place a unit in a region \( T_1 \) that is adjacent to region \( T_2 \) if \( T_2 \) is owned by the player with the most (PUM) or least (PUL) units. \\ \hline
PCM/PCL & Place a unit in a region \( T_1 \) that is adjacent to region \( T_2 \) if \( T_2 \) is owned by the player with the most (PCM) or least (PCL) zones owned (measured by total zone bonuses). \\ \hline
ETE/ETN & Place a unit in a region if it is adjacent to an enemy region (ETE) / not adjacent to any enemy regions (ETN). \\ \hline
EAC & Place a unit in a region if it is on a zone boundary. \\ \hline
EACM/EACL & Place a unit in a region if it is adjacent to the largest (EACM) / smallest (EACL) zone. \\ \hline
EACO & Place a unit in a region if it is adjacent to a zone that is completely owned by an enemy player. \\ \hline

\rowcolor{lightgray} \multicolumn{2}{|c|}{\textbf{Attack Phase}} \\ \hline
PTM/PTL & Attack from region \( T_1 \) to \( T_2 \) if \( T_2 \) is owned by the player with the most (PTM) or least (PTL) regions. \\ \hline
PUM/PUL & Attack from region \( T_1 \) to \( T_2 \) if \( T_2 \) is owned by the player with the most (PUM) or least (PUL) units. \\ \hline
PCM/PCL & Attack from region \( T_1 \) to \( T_2 \) if \( T_2 \) is owned by the highest (PCM) or lowest (PCL) number of zones owned (measured by total zone bonuses). \\ \hline
ONM/ONL & Attack if the units in \( T_1 \) are greater (ONM) / fewer (ONL) than the units in \( T_2 \). \\ \hline
ON2 & Attack if the units in \( T_1 \) are at least twice the number of units in \( T_2 \). \\ \hline
ICD/ICS & Attack from \(T_1\) to \(T_2\) if they are (ICD) / are not (ICS) in different zones. \\ \hline
L & Attack from \(T_1\) to \(T_2\) if \(T_2\) connects \(T_1\) to a region you own that it isn't currently connected to. \\ \hline
PASS & Likelihood of passing, ending your turn without more attacks. If set to 100, you will never attack; if set to 0, you will always attack. \\ \hline

\rowcolor{lightgray} \multicolumn{2}{|c|}{\textbf{Redeployment Phase}} \\ \hline
OBTM/OBTL & Move from \( T_1 \) to \( T_2 \) if \( T_2 \) is adjacent to more regions occupied by the player with the most (OBTM) or least (OBTL) regions. \\ \hline
OBUM/OBUL & Move from \( T_1 \) to \( T_2 \) if \( T_2 \) is adjacent to more regions occupied by the player with the most (OBUM) or least (OBUL) units. \\ \hline
OBCM/OBCL & Move from \( T_1 \) to \( T_2 \) if \( T_2 \) is adjacent to more regions occupied by the player with the most (OBCM) or least (OBCL) zones owned (measured by total zone bonuses). \\ \hline
CNM/CNL & Move from \( T_1 \) to \( T_2 \) if \( T_2 \) is connected to more (CNM) / fewer (CNL) regions. \\ \hline
CB & Move from \( T_1 \) to \( T_2 \) if \( T_2 \) is on a zone boundary and \(T_1\) is not. \\ \hline
CA & Move from \( T_1 \) to \( T_2 \) if \( T_2 \) is adjacent to at least one region occupied by an enemy player and \(T_1\) is not. \\ \hline
CAC & Move from \( T_1 \) to \( T_2 \) if \( T_2 \) is on a zone boundary and \(T_1\) is not. \\ \hline
M/L & Move from \( T_1 \) to \( T_2 \) if \( T_2 \) has more (M) / fewer (L) units on it. \\ \hline
SI & Move from \( T_1 \) to \( T_2 \) if \( T_2 \) is adjacent to a region that has a higher chance of successful invasion than those connected to \( T_1 \), calculated using the ratio of available troops from attacking region to troops on target region. \\ \hline
PASS & Likelihood of passing, ending your turn without more redeployments. If set to 100, you will never redeploy; if set to 0, you will always redeploy. \\ \hline

\end{tabularx}
\end{table*}

    \subsection{Game Environment}
        For our strategic reasoning test environment, we implemented a game loosely inspired by the board game Risk\textsuperscript{\textregistered}, which we call PERIL.\footnote{PERIL stands for \textbf{P}lease \textbf{E}veryone, we’re \textbf{R}epelling \textbf{I}nfringement \textbf{L}awsuits.} In our game, up to six players control \textit{units} on a map of \textit{regions} divided into \textit{zones}. The objective of the game is to achieve an assigned mission, but for the present work all players have the same mission, in order to minimize confounding factors: to achieve world domination by occupying all regions. The regions are connected to each other and either connect over land or water. Players control a set of units representing armies. Every region must have at least one unit on it (except in the beginning of the game, when no regions have any). Regions can contain an unlimited number of units, but all units on a single region must belong to the same player.

        The game starts in an initialization phase, where all players are given a number of units. They alternate, placing one unit on an unoccupied region at a time, until all regions are occupied, and then all units initially given to each player are placed. Each player then takes their regular turns. Each turn is broken up into three subphases: reinforcement (players receive additional units they can place), attack (players decide which regions to attack, where they are only able to attack regions adjacent to those they control), and redeployment (players can move units to connected regions).

        Using Python and the Dash Cytoscape library \cite{dash_cytoscape}, we implemented a framework for creating custom maps and allowing AI-controlled players to compete (Figure \ref{fig:Risk}).\footnote{https://github.com/Advancing-Machine-Human-Reasoning-Lab/PERIL} We also implemented the ability to assign different missions to players, although that feature was not utilized in this paper. AI-controlled players can access the entire game state at any point of the game directly, but for this paper we focused on the setting of strategic priorities in the form of \textit{heuristics}, implemented as follows: At the beginning of each game, the AI-controlled player is given its assigned persona and a list of allowed heuristics in the game. It is then asked to assign priorities for each of these heuristics (more details on how these priorities are selected in \S \ref{sec:personaSelection}). During the game, at each game phase the set of allowed moves are enumerated and assigned point values based on the heuristics selected by the player. The move is then selected using a random selection with the point value as the weight. By default, all heuristics have the point value of 5. The set of heuristics is in Table \ref{tbl:heuristics}.
        
    \subsection{Persona Selection}
        \label{sec:personaSelection}
        To select our personas, we started with Persona Hub \cite{Chan2024}, a collection of one billion diverse, synthetically-generated persona descriptions. Each persona consists of a short personality description, e.g., ``An elderly woman who loves watching makeup tutorials online and enjoys discussing beauty products.'' We randomly-selected 175,000 personas and annotated them using GPT-4o-mini, by asking it to independently rate each persona on the following features that we hypothesized would be predictive of game performance:

        \begin{itemize}
            \item \textbf{strategicThinker} a strategic thinker.
            \item \textbf{domainExpert} someone who has experience in combat, military warfare, or similar areas of expertise.
            \item \textbf{perilSpecific} someone who is likely to perform well on the game of PERIL specifically.
            \item \textbf{riskTaker} someone who is likely to take risks.
            \item \textbf{doOrBe} an instruction to act a certain way by describing actions (do) or if it is an instruction to play a role by describing a personality or character background (be).
        \end{itemize}
        Instructions were provided to rate all features on a scale from -1 to +1, in increments of 0.5. A qualitative description for each rating was provided to increase reliability. These ratings serve as a reference for evaluating how well heuristic generation methods correlate with persona characteristics, but they are not intended as ground-truth measures of personality. We then used a greedy algorithm to find the 50 personas that maximize the product of the variances across all features $F$:
        \begin{equation}
            \label{eqn:variance}
            \arg\max_{P} \prod_{f \in F}\sum_{p \in P} (f_p - \overline{f_P})^2
        \end{equation}
        Where $f_p$ is the rating of feature $f$ given to persona $p$ in subset $P$, and $\overline{f_P}$ is the average $f$ rating of all items in $P$. To find a greedy approximation, for each persona $p$ in the original set of 175K personas, we started with the set $\{p\}$ and iteratively added the persona that most increased Equation \ref{eqn:variance}, until we reached the maximum set size of 50. This was repeated for all possible personas $i$ as starting points. The best set $P$ is the final set of personas we will refer to for the remainder of this paper.
        
    \subsection{Heuristic Selection}

        Four LLMs were selected to generate heuristics: gpt-4o-2024-11-20, Meta-Llama-3-8B-Instruct, Llama-4-Maverick-17B-128E-Instruct-FP8, and Mistral-Small-Instruct-2409. These models were selected due to their relevance in current literature (GPT4) and to examine how model size impacts heuristic generation (Llama). Mistral small was chosen to provide an additional small sized model. Non-GPT models were downloaded from HuggingFace\footnote{https://huggingface.co/models} and run with stock configurations on 6 H100 GPUs. Two strategies were used to convert personas into heuristic values. In the first (which we will call the ``\textbf{direct heuristic}'' (DH) players), the LLM was prompted with the instructions of the game, a list of the heuristics available for a given game phase, and an example of how to provide values for each heuristic (ranging from 0 to 100, with a default value of 5). However, this generation method can lead to inconsistencies. For example, two similar personas may lead to proportional but different values assigned to each heuristic. Furthermore, the same persona may manifest in different ways in the three different phases, motivating a need to have personality features translate more evenly into heuristic values. 

        For this reason, our second strategy (which we will call the ``\textbf{personality inventory}'' (PI) players) utilizes a personality inventory, inspired by the American Psychiatric Association's Personality Inventory for DSM-5 (PID-5)—Adult \cite{Krueger2012}. A set of question items are provided, each with a first-person statement (e.g., ``I deserve special treatment.'') and the option to select how true this statement is: Very false or often false (-2 points), Sometimes or somewhat false (-1 points), Sometimes or somewhat true (+1 points), Very true or often true (+2 points). Each item maps either positively or negatively to some heuristics. For example, one item is ``I prefer to spread my influence to less contested or peripheral regions." This maps positively to heuristics \textbf{PTL, PUL, PCL, ETN}, and \textbf{EACL}. It maps negatively to \textbf{PTM, PUM, PCM, ETE}, and \textbf{EACM}. If a LLM responds to this item with ``very true or often true'', then 2 points will be added to heuristics in the positive set, and 2 points will be subtracted from the negative set.


        To generate the PI heuristics, we followed three steps. First, we used an LLM (GPT-4) to draft a questionnaire inventory. Second, we manually curated this inventory to ensure clarity and consistency. Third, we prompted an LLM to answer each item in the questionnaire using a persona (note that heuristic information and mappings were not provided at this stage). These point values were then converted into heuristic weights in a range matching that of the direct heuristic players. For some heuristic value $H$, let $r(H)$ be the number of points assigned to that heuristic divided by the maximum amount of possible points that could have been assigned to it. Then the weight $w(H)$ is:
        \[ 
            \begin{array}{ll}
                  \max(0,\lambda*(r(H)/5) + 5) & r(H)\leq 0 \\
                  \min(100, \lambda*(95*r(H))+5) & r(H) > 0\\
            \end{array} 
        \]
        This transformation ensures that if $r(H)=0$, then $w(h)=5$, and $0 \leq r(H) \leq 100$, since 5 was the default value in the prompt. For the experiments in this paper, we use $\lambda=0.5$, which made the average heuristic value across all personas of personality inventory players approximately equal to that of the direct heuristic players. 
    

        \subsection{Tournament Games}
        To compare players, we set up a series of matches between the fifty personas. 
        In each round, the 50 players were paired up randomly. When all 25 games are played, the players are paired up again for another round. This process is repeated for 49 rounds per run (1225 games). We carried out two runs with the 50 personality inventory players, and two additional runs with the 50 direct heuristic players. To measure player skill level, we use the TrueSkill algorithm \cite{Herbrich2006}, a generalization of the more well-known ELO score, allowing for games involving more than two players at a time (although we do not use that feature in this paper). If a game extended past 250 turns, it is declared a draw and all players are counted as having lost for the purpose of the TrueSkill calculation (this occurred in $< 0.05\%$ of games). We use two-player games to reduce the confounding effect of larger numbers of players, but note that many of the heuristics we defined (Table \ref{tbl:heuristics}) only produce observably different behaviors when there are more than two players total. Nevertheless, we retain them here to study how persona-prompted LLMs select values for those heuristics.

\section{Results}
    Our results can be summarized in four main findings, presented in the order they appear below. First, we compare player performance, showing that strategically themed personas (e.g., strategists, officers, athletes) consistently achieve higher TrueSkill scores than other personas (e.g., students or children; Table \ref{tab:elo_personas}). Second, we analyze correlations between inferred persona features and generated heuristics, finding that the PI method produces significant correlations with features such as \textit{strategicThinker}, \textit{domainExpert}, and \textit{perilSpecific}, whereas the DH method shows weak or inconsistent correlations (Table \ref{tbl:featureCorrelations}). Third, we examine cross-run reliability of final rankings, finding that DH players were more stable across trials, but this stability was not tied to persona features. Finally, we evaluate opposite-value consistency in heuristic assignments, where GPT-4 generated coherent mappings under the PI method, while smaller models produced noisier and less reliable patterns (Table \ref{tbl:modelOppositeHeuristicDiffs}). Overall, these results demonstrate that the PI method yields more distinct and interpretable persona-driven behaviors than the DH method.

    \subsection{Comparison of Personality Inventory Players}
    The relationship between actual performance and inferred personality features are apparent when comparing the top 5 and bottom 5 performers of the first run of personality inventory players (Table \ref{tab:elo_personas}). To determine whether a relationship existed between inferred personality features and final TrueSkill score, we calculated the Spearman correlation (Table \ref{tbl:featureCorrelations}). For personality inventory runs, performance was significantly correlated with the features \textit{strategicThinker}, \textit{domainExpert}, and \textit{perilSpecific}, but not with \textit{riskTaker} and \textit{doOrBe}. For the direct heuristic runs, the majority of features showed no significant correlation with performance. Additionally, the DH feature correlations are prone to dramatic changes across trials providing additional concern for the reliability of directly prompting for inventory responses. Conversely, the PI generations repeatedly provided highly significant correlations across multiple trials. LLaMA 3 was an exception to this trend, but as Mistral was able to generate highly correlated scores the difference in results is not accounted for by model size but by training or architecture choices.



    \begin{table}[t]
        \footnotesize
        \centering
        \begin{tabularx}{\columnwidth}{|l|X|}
        \hline
        \rowcolor{lightgray} \textbf{Rating} & \textbf{Persona Description} \\ 
        \hline
        28.08 & A geopolitical strategist who often appears on different networks presenting an alternative viewpoint on policies and events \\ \hline
        27.89 & A government agency using GIS analysis to plan efficient land use and infrastructure development \\ \hline
        27.81 & A competitive collegiate football player always seeking for custom-designed team merchandise \\ \hline 
        27.35 & A retired intelligence officer who had previously worked for the CIA. \\ \hline
        27.11 & A person who struggles with Discardia – a fear of throwing things away.\\ \hline 
        \hline
        23.03 & A genealogist researching family histories connected to Biddeford, Maine. \\ \hline
        22.04 & A genealogist helping clients trace their family roots, particularly those with connections to the Somme department in France. \\ \hline
        21.83 & A healthcare blogger who spreads misinformation about vaccines and challenges the nurse's beliefs \\ \hline
        21.70 & A struggling high school student who has no interest in biology. \\ \hline
        18.64 & A young child who laughs uncontrollably at the street performer's antics \\ \hline
        \end{tabularx}
        \caption{TrueSkill for GPT4 - PI1 - Run 1. Higher TrueSkill ratings align with human expectations: tactically minded personas (e.g., strategists, officers, athletes) outperform less focused ones (e.g., under-performing students, children).}
        \label{tab:elo_personas}
    \end{table}
    
    These results support our suspicion that the direct heuristic method does not produce actual behaviors that differentiate strongly between personality prompts, whereas the personality inventory method does (at least with respect to the personality features we studied here). Furthermore, the correlations of personality inventory player performance with the LLM-annotated \textbf{perilSpecific} feature (row 3 of Table \ref{tbl:featureCorrelations}) is a promising sign that, given a shallow description of a player's personality, and assuming their personality translates predictably into their play style, LLMs may have some ability to predict player performance. Note that this does \textit{not} necessarily demonstrate that the reason for the performance difference is that the top performers actually are exhibiting the personality traits we have identified as winning. At a minimum, this data shows that some qualitative difference exists which: (1) affects performance in the game of Peril, (2) was elicited by the persona prompting method augmented with personality inventories, and (3) was successfully identified by an LLM annotator which looked only at persona descriptions. Additionally, these early results show that the personality inventory method of translating persona descriptions into heuristic choices leads to more observably distinct behaviors than the direct heuristic method. We suspect that this is due to the personality inventory's ability to enforce that interpretations of personality traits apply more evenly across all heuristics. 

    However, the TrueSkill algorithm has a bit of a locking-in effect, where after a large number of games, the amount that subsequent games change one's TrueSkill rating is decreasingly small. This is the reason that we had two separate runs for the personality inventory players, and likewise for the direct method players. Given this fact, was the final ranking of players consistent between these runs? The final player ranking of all players in the first and second runs had a close-to-significant Spearman correlation ($\rho=0.278,p=0.051$). However, interestingly (and counter-intuitively), the correlation between runs three and four was much stronger ($\rho=0.524,p<0.001$). Likewise, the final rankings of personality inventory player runs did not correlate significantly with final rankings of the direct heuristic runs (all were $p>0.2$). Thus, although the personality inventory method produced behaviors whose performance spreads more closely aligned with personality features (Table \ref{tbl:featureCorrelations}), the direct heuristic method produced behaviors that performed more consistently relative to other players. Alternatively, this shows that although previous performance of direct heuristic players is more predictive of future performance than previous performance of personality inventory players, inferred personality features are more predictive of personality inventory player performance than it is of direct heuristic player performance.


    \subsection{Comparison of Personality Inventory and Direct Heuristic Players}
    \label{sec:compareHeuristicChoices}

    One of the motivators for our introduction of the personality inventory method was the observation that the direct heuristic method led to inconsistent heuristic choices. It is expected that, given a single persona \textit{P}, and a set of heuristics, the value assigned to each heuristic should have what we call \textit{opposite-value consistency}: heuristics specifying opposite properties should have opposite values (Table \ref{tbl:modelOppositeHeuristicDiffs}). To measure opposite-value consistency, for each heuristic that has a direct opposite, we measure the ratio between them (using the larger value as the numerator), and average across all players $\mathbf{P}$. Because the range of possible heuristic values is 0 to 100, we cap this ratio at 100. For some opposite heuristic pair ($h_1, h_2$), if player $p$ has heuristic values $h_1^p, h_2^p$, the opposite-value consistency is:

    \[ \left( \sum_{p \in \mathbf{P}} \max(\frac{h_1^p}{h_2^p}, \frac{h_2^p}{h_1^p}, 100) \right) / |\mathbf{P}| \]
    
    The opposite-value consistency for personality inventory heuristics are strongly impacted by the models used to generate them, and the smaller models did not behave in a manner consistent with expectations based on the prompts. The ability for GPT-4 to correctly identify mutually exclusive heuristics may need to manually be accounted for when transferring the prompts to smaller models.

    \begin{table}[t!]
    \footnotesize
        \centering
        \resizebox{\columnwidth}{!}{\begin{tabular}{|c|c|c|c|c|c|}
            \hline
            \rowcolor{lightgray} \textbf{Feature} & \textbf{Model} & \textbf{PI1} & \textbf{PI2} & \textbf{DH1} & \textbf{DH2} \\ \hline
            \textbf{strategicThinker} & GPT4 & \textbf{0.49***} & \textbf{0.40***} & 0.08 & 0.05 \\ \hline
                                      & Mistral R1 & 0.2447 & \textbf{0.3826**} & 0.0046 & 0.0645 \\ \hline
                                      & Mistral R2 & \textbf{0.4391**} & \textbf{0.4360**} & 0.2218 & 0.2644 \\ \hline
                                      & LLaMA 3 R1 & 0.2192 & 0.2400 & 0.0211 & 0.1124 \\ \hline
                                      & LLaMA 3 R2 & 0.2076 & \textbf{0.3938**} & 0.0370 & -0.0572 \\ \hline
                                      & LLaMA 4 R1 & \textbf{0.5707***} & 0.2480 & 0.2218 & \textbf{0.3531*} \\ \hline
                                      & LLaMA 4 R2 & \textbf{0.4730***} & \textbf{0.5040***} & 0.1999 & 0.0892 \\ \hline
            \textbf{domainExpert}    & GPT4 & \textbf{0.41***} & \textbf{0.44***} & 0.12 & 0.03 \\ \hline
                                      & Mistral R1 & \textbf{0.2848*} & 0.2688 & -0.0269 & 0.1239 \\ \hline
                                      & Mistral R2 & \textbf{0.3538*} & \textbf{0.4900***} & 0.1985 & 0.2696 \\ \hline
                                      & LLaMA 3 R1 & 0.1994 & 0.1583 & 0.1027 & 0.1561 \\ \hline
                                      & LLaMA 3 R2 & 0.2713 & \textbf{0.3687**} & -0.0275 & -0.0548 \\ \hline
                                      & LLaMA 4 R1 & \textbf{0.6166***} & \textbf{0.2796*} & 0.2438 & \textbf{0.3817**} \\ \hline
                                      & LLaMA 4 R2 & \textbf{0.4709***} & \textbf{0.5021***} & \textbf{0.3036*} & 0.0813 \\ \hline
            \textbf{perilSpecific}   & GPT4 & \textbf{0.41***} & \textbf{0.39***} & 0.11 & 0.02 \\ \hline
                                      & Mistral R1 & \textbf{0.2961*} & \textbf{0.3460*} & -0.0008 & 0.0743 \\ \hline
                                      & Mistral R2 & \textbf{0.4572***} & \textbf{0.4859***} & 0.2003 & 0.2448 \\ \hline
                                      & LLaMA 3 R1 & 0.2033 & 0.2355 & 0.0221 & 0.0979 \\ \hline
                                      & LLaMA 3 R2 & \textbf{0.3161*} & \textbf{0.4151**} & 0.0017 & -0.0514 \\ \hline
                                      & LLaMA 4 R1 & \textbf{0.6948***} & \textbf{0.3967**} & 0.2272 & \textbf{0.3838**} \\ \hline
                                      & LLaMA 4 R2 & \textbf{0.4826***} & \textbf{0.5814***} & 0.1877 & 0.0314 \\ \hline
            \textbf{riskTaker}       & GPT4 & 0.14 & 0.07 & 0.09 & 0.02 \\ \hline
                                      & Mistral R1 & 0.1778 & 0.1473 & 0.0019 & -0.0192 \\ \hline
                                      & Mistral R2 & \textbf{0.4516***} & \textbf{0.3366*} & -0.0015 & 0.0879 \\ \hline
                                      & LLaMA 3 R1 & 0.0053 & -0.0901 & 0.0142 & 0.2775 \\ \hline
                                      & LLaMA 3 R2 & -0.0776 & -0.1119 & -0.0350 & 0.0275 \\ \hline
                                      & LLaMA 4 R1 & 0.1159 & 0.1205 & -0.0401 & 0.2409 \\ \hline
                                      & LLaMA 4 R2 & 0.0562 & 0.1416 & 0.1959 & 0.0950 \\ \hline
            \textbf{doOrBe}          & GPT4 & -0.04 & 0.10 & 0.06 & 0.15 \\ \hline
                                      & Mistral R1 & 0.2119 & 0.0182 & -0.0814 & -0.0293 \\ \hline
                                      & Mistral R2 & 0.2516 & 0.2240 & 0.0337 & 0.0829 \\ \hline
                                      & LLaMA 3 R1 & 0.0422 & 0.1111 & 0.2060 & 0.0007 \\ \hline
                                      & LLaMA 3 R2 & -0.1231 & -0.0647 & 0.0082 & -0.0422 \\ \hline
                                      & LLaMA 4 R1 & -0.1930 & -0.2326 & 0.0940 & 0.1138 \\ \hline
                                      & LLaMA 4 R2 & -0.2227 & -0.2128 & -0.0643 & 0.2227 \\ \hline
        \end{tabular}}
        \caption{The figures show the correlation between each personality feature and heuristic weights, as chosen by players using the direct and inventory heuristic methods. Each figure caption indicates the heuristic (DH/PI) used and its generation batch (1/2).  All batches other than GPT4 were generated twice. Additionally, the statistical significance of each entry is indicated by asterisks "*" as follows: $*=(p\leq0.05), **=(p\leq0.01), ***=(p\leq0.005)$. The direct heuristic methods are consistently less statistically significant than the inventory heuristics for all models and runs. This is to be expected as direct generation of personality traits is known to have lower reliability than generation via inventories.}
        \label{tbl:featureCorrelations}
    \end{table}

    \begin{table}[h!]
    \scriptsize
    \centering
        \resizebox{\columnwidth}{!}{\begin{tabular}{|c|c|c|c|c|c|}
        \hline
        \rowcolor{lightgray} \textbf{Phase} & \textbf{Heuristics} & \textbf{Mistral} & \textbf{LLaMA 3} & \textbf{LLaMA 4} & \textbf{GPT4} \\ \hline
        0 & EACM–EACL & 18.16 & 38.25 & 6.515 & \textbf{-2.61} \\ \hline
          & ETE–ETN   & 3.93  & 8.82  & 39.13  & \textbf{-1.63} \\ \hline
          & PCM–PCL   & 26.18 & 17.46 & 29.45  & \textbf{-13.03} \\ \hline
          & PTM–PTL   & 23.02 & 22.46 & 18.55  & \textbf{-4.00} \\ \hline
          & PUM–PUL   & 23.82 & 22.38 & 33.82  & \textbf{-13.00} \\ \hline
        1 & ICD–ICS   & 8.75  & 25.30 & \textbf{-2.05}  & -5.20 \\ \hline
          & ONM–ONL   & 26.22 & 7.73  & 71.20  & \textbf{-4.20} \\ \hline
          & PCM–PCL   & 12.85 & 16.99 & \textbf{10.40}  & -47.87 \\ \hline
          & PTM–PTL   & 21.49 & 11.19 & 12.76  & \textbf{-0.61} \\ \hline
          & PUM–PUL   & 18.09 & 14.74 & \textbf{14.13}  & -14.62 \\ \hline
        2 & CNM–CNL   & -4.98 & \textbf{-4.92} & 38.49  & -15.33 \\ \hline
          & M–L       & \textbf{-3.49} & 11.61 & 16.93  & -11.28 \\ \hline
          & OBCM–OBCL & 14.21 & 33.15 & 21.18  & \textbf{1.36} \\ \hline
          & OBTM–OBTL & \textbf{4.33}  & 26.49 & 20.74  & -37.74 \\ \hline
          & OBUM–OBUL & \textbf{2.60}  & 39.03 & 19.21  & -18.84 \\ \hline
        \end{tabular}}
    \caption{Average difference in scores between opposite heuristics for models across phases. Positive values indicate higher DH scores and negative values indicate higher PI scores. Bold values indicate more similar opposite value consistency across DH and PI methods. Non-GPT models produced more conflicting from PI than DH, shown in larger average DH values. }
    \label{tbl:modelOppositeHeuristicDiffs}
    \end{table}

\section{Conclusion}

    In this work, we explored the differences between two methods of translating persona descriptions into actual behaviors. The first, direct heuristics, is by far the most common method seen in current literature. However, we observed that this method leads to heuristic selection that was inconsistent for a given persona. As shown in Table \ref{tbl:featureCorrelations}, the DH method resulted in correlations that were both weak and non-significant. We therefore proposed a personality inventory technique, which involves the creation of an inventory questionnaire that translates personas into heuristic values. We showed that the inventory method leads to heuristic values that are more consistent also shown in Table \ref{tbl:featureCorrelations}. The overwhelming majority of significant high correlation heuristics were generated by the PI method. It is also important to note that while smaller models (Mistral and LLaMA 3) did not provide correlations as strong, they still clearly behave similarly to the larger models in regards to heuristic correlation. Additional results can be found in the Appendix.
    
    This work contributes to research showing that although persona prompting alone may lead to LLM outputs with different styles of text, it does not necessarily lead to substantially different decision-making behaviors. This further highlights the difficulty in translating personality descriptions into anything beyond surface-level expressions, since predicting behavioral differences between personalities requires a much deeper knowledge of the causality underlying human behavior than simply mimicking speech patterns. To address this, we showed that a personality inventory questionnaire-based approach can be more effective at eliciting behavioral heuristics that seem to better align with expectations of various personality descriptions, when compared to an approach that directly infers heuristics from personality descriptions. \textit{In short: a main takeaway from this work is that simply asking an LLM to act in accordance with a persona, without using a moderator like the personality inventory method, may not suffice to produce realistic and diverse behaviors and decision-making.}
    
    \section{Future Work.} As part of this work, we did implement a PERIL \textit{mission mode}, in which all players are assigned missions to achieve victory other than world domination. Although we did not utilize this functionality in the present paper (as it introduces another confounding variable), we will release the source code both of PERIL and the code used to replicate this paper's results, in order to encourage other researchers to explore the interesting problems in this space. For example, recent work shows that multi-agent systems may perform better at simulations than single-agent systems \cite{Sreedhar2024, bui2025mixture}, but it is unclear how  current state-of-the-art agents perform when playing adversarially against other potentially deceptive agents.

    \section{Limitations.} It should be noted that any observations we make here about how well a persona-prompted LLM matches its given persona can only at most have \textit{face validity}---i.e., they only appear intuitively to match personality archetypes. We cannot fully establish whether these patterns have deeper alignment with actual human personalities without a proper psychometrically-designed empirical study on a population of people. Instead, the value in our present work is in the introduction of the personality inventory method, and the finding that without it, the variance in behaviors between persona prompts and the adherence to expected patterns such as opposite-value consistency are very small. Indeed, if it can be shown that the translation of personality features into heuristic behaviors has more than face validity, it can lead to powerful simulation technologies, as well as tools for studying the effect of personality on decision-making. For example, it might allow us to predict that a human being who matches a given persona would behave a certain way in a new scenario.

    In this exploratory work, a single strategic environment (the game of PERIL). Within this environment, we used a limited number of heuristics, which are not fully representative of the range of possible decision-making heuristics in this game. Furthermore, our heuristic-guided agents made their heuristic choices in the beginning of the game, which reduced their adaptivity, since they could not adjust those heuristics in response to game conditions. Finally, we used TrueSkill as a way of estimating player performance, but it should be noted that not enough games were played to allow every player to face every other one, and due to the way TrueSkill is calculated, ordering affects final scores. This effect seemed to affect individual players more in the personality inventory runs than in the direct heuristic runs (as measured by correlation of final rankings of players), but it did not significantly change the effect of personality features on final player ordering (Table \ref{tbl:featureCorrelations}). The low correlation between final player rankings in the two PI runs is counter-intuitive to us, and future work will need to explore why this was the case.


\section*{Ethical Statement}

We do not anticipate significant ethical issues, as this work did not involve human subjects or the collection of personal data. This research explores artificial intelligence strategies in the strategy-based board game PERIL and draws inspiration from the mechanics of Risk\textsuperscript{\textregistered}, a trademarked board game owned by Hasbro, Inc. The game developed and described in this paper is an independent project and is not affiliated with, endorsed by, or associated with Hasbro, Inc. The use of Risk\textsuperscript{\textregistered} as a reference point is solely for comparative analysis and academic purposes under the principles of fair use. No proprietary elements, such as trademarked names, graphics, or copyrighted texts, have been reproduced in this work. In the interest of research transparency and replicability, upon acceptance for publication we will release the full source code, along with all prompts used in this work, on a publicly accessible GitHub repository.

\section*{Acknowledgments}

    This research was supported in part by the Air Force Research Laboratory, Information Directorate, through the Air Force Office of Scientific Research Summer Faculty Fellowship Program\textsuperscript{\textregistered}, Contract Numbers FA8750-15-3-6003, FA9550-15-0001 and FA9550-20-F-0005.

\bibliography{john,stephen,temp}

@article{li2025llm,
  title={LLM Generated Persona is a Promise with a Catch},
  author={Li, Ang and Chen, Haozhe and Namkoong, Hongseok and Peng, Tianyi},
  journal={arXiv preprint arXiv:2503.16527},
  year={2025}
}

@article{wu2025personas,
  title={From personas to talks: Revisiting the impact of personas on llm-synthesized emotional support conversations},
  author={Wu, Shenghan and Deng, Yang and Zhu, Yimo and Hsu, Wynne and Lee, Mong Li},
  journal={arXiv preprint arXiv:2502.11451},
  year={2025}
}

@inproceedings{bhandari2025evaluating,
  title={Evaluating personality traits in large language models: Insights from psychological questionnaires},
  author={Bhandari, Pranav and Naseem, Usman and Datta, Amitava and Fay, Nicolas and Nasim, Mehwish},
  booktitle={Companion Proceedings of the ACM on Web Conference 2025},
  pages={868--872},
  year={2025}
}

@article{maharjan2025large,
  title={Do Large Language Models Really Understand Personality?},
  author={Maharjan, Julina and Jin, Ruoming and Zhu, Jianfeng and Kenne, Deric},
  journal={Journal of Medical Internet Research},
  volume={21},
  number={05},
  year={2025}
}

@article{frisch2024llm,
  title={LLM agents in interaction: Measuring personality consistency and linguistic alignment in interacting populations of large language models},
  author={Frisch, Ivar and Giulianelli, Mario},
  journal={arXiv preprint arXiv:2402.02896},
  year={2024}
}

@article{pratelli2025evaluating,
  title={Evaluating the Simulation of Human Personality-Driven Susceptibility to Misinformation with LLMs},
  author={Pratelli, Manuel and Petrocchi, Marinella},
  journal={arXiv preprint arXiv:2506.23610},
  year={2025}
}

@article{phelps2025machine,
  title={The machine psychology of cooperation: can GPT models operationalize prompts for altruism, cooperation, competitiveness, and selfishness in economic games?},
  author={Phelps, Steve and Russell, Yvan I},
  journal={Journal of Physics: Complexity},
  volume={6},
  number={1},
  pages={015018},
  year={2025},
  publisher={IOP Publishing}
}

@article{poterti2025designing,
  title={Designing role vectors to improve llm inference behaviour},
  author={Potert{\`\i}, Daniele and Seveso, Andrea and Mercorio, Fabio},
  journal={arXiv preprint arXiv:2502.12055},
  year={2025}
}

@article{bhandari2025can,
  title={Can LLM Agents Maintain a Persona in Discourse?},
  author={Bhandari, Pranav and Fay, Nicolas and Wise, Michael and Datta, Amitava and Meek, Stephanie and Naseem, Usman and Nasim, Mehwish},
  journal={arXiv preprint arXiv:2502.11843},
  year={2025}
}

@article{belle2025agents,
  title={Agents of Change: Self-Evolving LLM Agents for Strategic Planning},
  author={Belle, Nikolas and Barnes, Dakota and Amayuelas, Alfonso and Bercovich, Ivan and Wang, Xin Eric and Wang, William},
  journal={arXiv preprint arXiv:2506.04651},
  year={2025}
}

@article{wang2025opera,
  title={OPeRA: A Dataset of Observation, Persona, Rationale, and Action for Evaluating LLMs on Human Online Shopping Behavior Simulation},
  author={Wang, Ziyi and Lu, Yuxuan and Li, Wenbo and Amini, Amirali and Sun, Bo and Bart, Yakov and Lyu, Weimin and Gesi, Jiri and Wang, Tian and Huang, Jing and others},
  journal={arXiv preprint arXiv:2506.05606},
  year={2025}
}

@article{jiang2025harbor,
  title={HARBOR: exploring persona dynamics in multi-agent competition},
  author={Jiang, Kenan and Xiong, Li and Liu, Fei},
  journal={arXiv preprint arXiv:2502.12149},
  year={2025}
}

@article{bui2025mixture,
  title={Mixture-of-personas language models for population simulation},
  author={Bui, Ngoc and Nguyen, Hieu Trung and Kumar, Shantanu and Theodore, Julian and Qiu, Weikang and Nguyen, Viet Anh and Ying, Rex},
  journal={arXiv preprint arXiv:2504.05019},
  year={2025}
}

@article{xiao2025humanizing,
  title={Humanizing Machines: Rethinking LLM Anthropomorphism Through a Multi-Level Framework of Design},
  author={Xiao, Yunze and Ng, Lynnette Hui Xian and Liu, Jiarui and Diab, Mona T},
  journal={arXiv preprint arXiv:2508.17573},
  year={2025}
}

@inproceedings{devrio2025taxonomy,
  title={A Taxonomy of Linguistic Expressions That Contribute To Anthropomorphism of Language Technologies},
  author={DeVrio, Alicia and Cheng, Myra and Egede, Lisa and Olteanu, Alexandra and Blodgett, Su Lin},
  booktitle={Proceedings of the 2025 CHI Conference on Human Factors in Computing Systems},
  pages={1--18},
  year={2025}
}

@inproceedings{jiang-etal-2024-personallm,
    title = "{P}ersona{LLM}: Investigating the Ability of Large Language Models to Express Personality Traits",
    author = "Jiang, Hang  and
      Zhang, Xiajie  and
      Cao, Xubo  and
      Breazeal, Cynthia  and
      Roy, Deb  and
      Kabbara, Jad",
    editor = "Duh, Kevin  and
      Gomez, Helena  and
      Bethard, Steven",
    booktitle = "Findings of the Association for Computational Linguistics: NAACL 2024",
    month = jun,
    year = "2024",
    address = "Mexico City, Mexico",
    publisher = "Association for Computational Linguistics",
    url = "https://aclanthology.org/2024.findings-naacl.229/",
    doi = "10.18653/v1/2024.findings-naacl.229",
    pages = "3605--3627"
}

@article{Bodroza2024,
   title={Personality testing of large language models: limited temporal stability, but highlighted prosociality},
   volume={11},
   ISSN={2054-5703},
   url={http://dx.doi.org/10.1098/rsos.240180},
   DOI={10.1098/rsos.240180},
   number={10},
   journal={Royal Society Open Science},
   publisher={The Royal Society},
   author={Bodroža, Bojana and Dinić, Bojana M. and Bojić, Ljubiša},
   year={2024},
   month=oct }

@article{Krueger2012,
	author = {Krueger, R F and Derringer, J and Markon, K E and Watson, D and Skodol, A E},
	journal = {Psychol Med},
	month = {Sep},
	number = {9},
	pages = {1879--1890},
	title = {Initial construction of a maladaptive personality trait model and inventory for DSM-5.},
	volume = {42},
	year = {2012}}

@misc{dash_cytoscape,
  author = {Plotly Technologies Inc.},
  title = {Dash Cytoscape: Interactive network visualization in Python},
  year = {2018},
  url = {https://github.com/plotly/dash-cytoscape}
}

@inproceedings{hendrycks2021measuring,
  author    = {Dan Hendrycks and Collin Burns and Steven Basart and Andy Zou and Mantas Mazeika and Dawn Song and Jacob Steinhardt},
  title     = {Measuring Massive Multitask Language Understanding},
  booktitle = {Proceedings of the International Conference on Learning Representations (ICLR)},
  year      = {2021}
}

@inproceedings{Zheng2024,
    title = "When {''}A Helpful Assistant{''} Is Not Really Helpful: Personas in System Prompts Do Not Improve Performances of Large Language Models",
    author = "Zheng, Mingqian  and
      Pei, Jiaxin  and
      Logeswaran, Lajanugen  and
      Lee, Moontae  and
      Jurgens, David",
    editor = "Al-Onaizan, Yaser  and
      Bansal, Mohit  and
      Chen, Yun-Nung",
    booktitle = "Findings of the Association for Computational Linguistics: EMNLP 2024",
    month = nov,
    year = "2024",
    address = "Miami, Florida, USA",
    publisher = "Association for Computational Linguistics",
    url = "https://aclanthology.org/2024.findings-emnlp.888",
    pages = "15126--15154",
}

@article{Hu2024,
  author    = {Tiancheng Hu and Nigel Collier},
  title     = {Quantifying the Persona Effect in LLM Simulations},
  journal   = {arXiv preprint arXiv:2402.10811},
  year      = {2024},
  url       = {https://arxiv.org/abs/2402.10811}
}

@article{Chan2024,
  author    = {Xin Chan and Xiaoyang Wang and Dian Yu and Haitao Mi and Dong Yu},
  title     = {Scaling Synthetic Data Creation with 1,000,000,000 Personas},
  journal   = {arXiv preprint arXiv:2406.20094},
  year      = {2024},
  url       = {https://arxiv.org/abs/2406.20094}
}

@article{Tseng2024,
  author    = {Yu-Min Tseng and Yu-Chao Huang and Teng-Yun Hsiao and Wei-Lin Chen and Chao-Wei Huang and Yu Meng and Yun-Nung Chen},
  title     = {Two Tales of Persona in LLMs: A Survey of Role-Playing and Personalization},
  journal   = {arXiv preprint arXiv:2406.01171},
  year      = {2024},
  url       = {https://arxiv.org/abs/2406.01171}
}

@article{Liu2024,
  author    = {Yiren Liu and Pranav Sharma and Mehul Jitendra Oswal and Haijun Xia and Yun Huang},
  title     = {PersonaFlow: Boosting Research Ideation with LLM-Simulated Expert Personas},
  journal   = {arXiv preprint arXiv:2409.12538},
  year      = {2024},
  url       = {https://arxiv.org/abs/2409.12538}
}

@article{Kamruzzaman2024,
  author    = {Mahammed Kamruzzaman and Gene Louis Kim},
  title     = {Exploring Changes in Nation Perception with Nationality-Assigned Personas in LLMs},
  journal   = {arXiv preprint arXiv:2406.13993},
  year      = {2024},
  url       = {https://arxiv.org/abs/2406.13993}
}

@article{Zhang2024,
  author    = {Yadong Zhang and Shaoguang Mao and Tao Ge and Xun Wang and Adrian de Wynter and Yan Xia and Wenshan Wu and Ting Song and Man Lan and Furu Wei},
  title     = {LLM as a Mastermind: A Survey of Strategic Reasoning with Large Language Models},
  journal   = {arXiv preprint arXiv:2404.01230},
  year      = {2024},
  url       = {https://arxiv.org/abs/2404.01230}
}

@article{Sreedhar2024,
  author    = {Karthik Sreedhar and Lydia Chilton},
  title     = {Simulating Human Strategic Behavior: Comparing Single and Multi-agent LLMs},
  journal   = {arXiv preprint arXiv:2402.08189},
  year      = {2024},
  url       = {https://arxiv.org/abs/2402.08189}
}

@article{Kim2024,
  author    = {Junseok Kim and Nakyeong Yang and Kyomin Jung},
  title     = {Persona is a Double-edged Sword: Mitigating the Negative Impact of Role-playing Prompts in Zero-shot Reasoning Tasks},
  journal   = {arXiv preprint arXiv:2408.08631},
  year      = {2024},
  url       = {https://arxiv.org/abs/2408.08631}
}

@inproceedings{Wang2024,
  author    = {Zhenhailong Wang and Shaoguang Mao and Wenshan Wu and Tao Ge and Furu Wei and Heng Ji},
  title     = {Unleashing the Emergent Cognitive Synergy in Large Language Models: A Task-Solving Agent through Multi-Persona Self-Collaboration},
  booktitle = {Proceedings of the 2024 Conference of the North American Chapter of the Association for Computational Linguistics: Human Language Technologies (Volume 1: Long Papers)},
  year      = {2024},
  pages     = {257--279},
  url       = {https://aclanthology.org/2024.naacl-long.15/}
}

@book{lingard1907letter,
  author    = {Lingard, R. and Erb, Frank C.},
  title     = {A Letter of Advice to a Young Gentleman Leaving the University Concerning His Behaviour and Conversation in the World},
  year      = {1907},
  publisher = {McAufliffe \& Booth},
  address   = {New York},
  note      = {Retrieved from the Library of Congress, \url{www.loc.gov/item/07007481/}}
}

@misc{hu2024surveylargelanguagemodelbased,
      title={A Survey on Large Language Model-Based Game Agents}, 
      author={Sihao Hu and Tiansheng Huang and Fatih Ilhan and Selim Tekin and Gaowen Liu and Ramana Kompella and Ling Liu},
      year={2024},
      eprint={2404.02039},
      archivePrefix={arXiv},
      primaryClass={cs.AI},
      url={https://arxiv.org/abs/2404.02039}, 
}

@article{Lore2024,
	author = {Lor{\`e}, Nunzio and Heydari, Babak},
	journal = {Scientific Reports},
	number = {1},
	pages = {18490},
	title = {Strategic behavior of large language models and the role of game structure versus contextual framing},
	volume = {14},
	year = {2024}
}

@article{meta2022human,
  title = {Human-level play in the game of Diplomacy by combining language models with strategic reasoning},
  author = {{Meta Fundamental AI Research Diplomacy Team (FAIR)} and Bakhtin, Anton and Brown, Noam and Dinan, Emily and Farina, Gabriele and Flaherty, Colin and Fried, Daniel and Goff, Andrew and Gray, Jonathan and Hu, Hengyuan and {others}},
  journal = {Science},
  volume = {378},
  number = {6624},
  pages = {1067--1074},
  year = {2022},
  publisher = {American Association for the Advancement of Science}
}

@inproceedings{Goecks2024,
  author    = {Vinicius G. Goecks and Nicholas Waytowich},
  title     = {COA-GPT: Generative Pre-Trained Transformers for Accelerated Course of Action Development in Military Operations},
  booktitle = {Proceedings of the 2024 IEEE Conference on Artificial Intelligence (IEEE AI 2024)},
  year      = {2024},
  pages     = {1--8},
  doi       = {10.1109/IEEEAI.2024.10540749},
  url       = {https://ieeexplore.ieee.org/document/10540749}
}

@inproceedings{Olea2024,
  author    = {Carlos Olea and Holly Tucker and Jessica Phelan and Cameron Pattison and Shen Zhang and Maxwell Lieb and Doug Schmidt and Jules White},
  title     = {Evaluating Persona Prompting for Question Answering Tasks},
  booktitle = {Proceedings of the 10th International Conference on Artificial Intelligence and Soft Computing},
  year      = {2024}
}

@article{Pan2024,
  author    = {Xuchen Pan and Dawei Gao and Yuexiang Xie and Yushuo Chen and Zhewei Wei and Yaliang Li and Bolin Ding and Ji-Rong Wen and Jingren Zhou},
  title     = {Very Large-Scale Multi-Agent Simulation in AgentScope},
  journal   = {arXiv preprint arXiv:2407.17789},
  year      = {2024},
  url       = {https://arxiv.org/abs/2407.17789}
}

@inproceedings{Lin2024,
  author    = {Shuhang Lin and Wenyue Hua and Lingyao Li and Che-Jui Chang and Lizhou Fan and Jianchao Ji and Hang Hua and Mingyu Jin and Jiebo Luo and Yongfeng Zhang},
  title     = {BattleAgent: Multi-modal Dynamic Emulation on Historical Battles to Complement Historical Analysis},
  booktitle = {Proceedings of the 2024 Conference on Empirical Methods in Natural Language Processing: System Demonstrations},
  year      = {2024},
  pages     = {172--181},
  url       = {https://aclanthology.org/2024.emnlp-demo.18/}
}

@article{Samuel2024,
  author    = {Vinay Samuel and Henry Peng Zou and Yue Zhou and Shreyas Chaudhari and Ashwin Kalyan and Tanmay Rajpurohit and Ameet Deshpande and Karthik Narasimhan and Vishvak Murahari},
  title     = {PersonaGym: Evaluating Persona Agents and LLMs},
  journal   = {arXiv preprint arXiv:2407.18416},
  year      = {2024},
  url       = {https://arxiv.org/abs/2407.18416}
}

@inproceedings{Schuller2024,
  author    = {Andreas Schuller and Doris Janssen and Julian Blumenr{\"o}ther and Theresa Maria Probst and Michael Schmidt and Chandan Kumar},
  title     = {Generating Personas Using LLMs and Assessing Their Viability},
  booktitle = {Extended Abstracts of the CHI Conference on Human Factors in Computing Systems (CHI EA '24)},
  year      = {2024},
  pages     = {179:1--179:7},
  doi       = {10.1145/3613905.3650860},
  url       = {https://dl.acm.org/doi/10.1145/3613905.3650860}
}

@inproceedings{Herbrich2006,
author = {Herbrich, Ralf and Minka, Tom and Graepel, Thore},
title = {TrueSkill™: a Bayesian skill rating system},
year = {2006},
publisher = {MIT Press},
address = {Cambridge, MA, USA},
booktitle = {Proceedings of the 19th International Conference on Neural Information Processing Systems},
pages = {569–576},
numpages = {8},
location = {Canada},
series = {NIPS'06}
}
\clearpage
\appendix
\section{Appendix}
This section contains the PERIL gameplay cycle, prompts, heuristic correlation tables, top 5 and bottom 5 TrueSkill persona rankings, and opposite value consistency tables. This material is not required to understand the main body of work, but readers may find it useful to gain a broader understanding of model impact on the results.
\subsection{Gameplay Cycle}
    \begin{itemize}
        \item \textbf{Initialize Game:}  
        Up to six players are placed on a map of regions divided into zones.  
        The mission is world domination by occupying all regions. Each region must eventually contain at least one unit.
    
        \item \textbf{Initialization Phase:}  
        Each player is given a pool of units.  
        Players alternate placing one unit on an unoccupied region until all regions are taken.  
        Remaining units are then placed on the regions they control.
    
        \item \textbf{Gameplay Loop (per turn):}
        \begin{itemize}
            \item \textit{Reinforcement:} Player receives new units and places them on controlled regions.  
            \item \textit{Attack:} Player may attack adjacent enemy regions using armies from their own regions.  
            \item \textit{Redeployment:} Player may move units between connected regions they control.
        \end{itemize}
    
        \item \textbf{End Condition:}  
        If a player occupies all regions, they are declared the winner.
    \end{itemize}

\subsection{Graphic}
    \begin{figure}[H]
        \centering
        \includegraphics[width=1\linewidth]{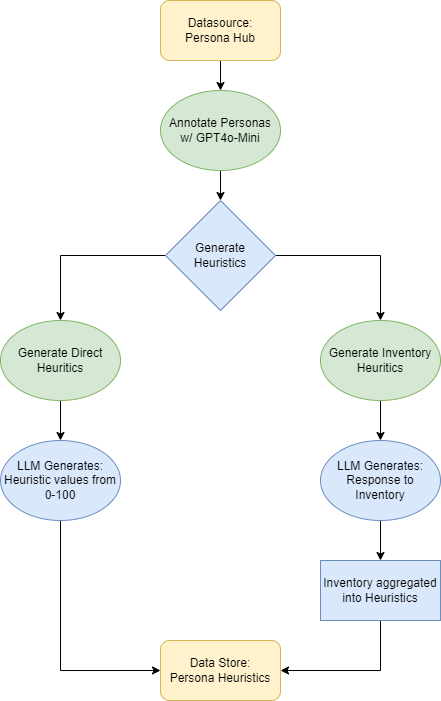}
    \caption{Pipeline from persona descriptions to heuristic agents. Note that LLMs generate heuristic proxies (DH or PI), but do not play PERIL directly.}
    \label{fig:pipeline}
    \end{figure}

\subsection{Prompts}
    We used three prompt sets to generate the heuristics used in this work. The first set was used to generate the initial assessments used for identifying fifty personas that are maximally distinct from each other. Usage of the assessment results is discussed in detail in section 3.2. The second set was to generate the direct heuristics used in playing the PERIL game. The third set was also used in gameplay, but it was used to generate the inventory heuristics. When generating any assessment or heuristic, the persona in question is always included in the prompt. 
\subsubsection{Assessment Prompt}
    You will be given a personality of an individual. You must estimate how well that person would do on the strategic board game Peril. In the game of Peril, you are a player who must achieve world conquest, very similar to the popular board game Risk. For the personality you are given, return a JSON object with the following values:
- 'index': An integer describing the index of the prompt in the dataset. We will provide this for you.
- 'personality': The personality of the individual, which we also provided to you.
- 'strategicThinker': A rating of whether this personality describes a strategic thinker. This should be one of the numbers in {-1, -0.5, 0, 0.5, 1}. Use the following scale:
	- +1 : This is certain to be a very capable strategic thinker, who combines systematic thinking with appropriate use of intuition and can consistently perform at a high level.
	- +0.5 : This is more likely than not to be someone with good strategic thinking capabilities, but they may not be consistent or fully developed.
	- 0 : There is no evidence to suggest that this person is or is not a capable strategic thinker.
	- -0.5 : This is more likely than not to be someone with poor or no strategic thinking capabilities.
	- -1: This is certain to be someone with poor or no strategic thinking capabilities, who is unable to perform at an acceptable level at any task requiring strategic thinking.
- 'domainExpert': A rating of whether this personality describes someone who has experience in combat, military warfare, or other similar areas of expertise. This should be one of the numbers in {-1, -0.5, 0, 0.5, 1}. Use the following scale:
	- +1 : This is certain to be someone who has high level experience in combat, military warfare, or other similar areas of expertise.
	- +0.5 : This is more likely than not to be someone who has experience in combat, military warfare, or other similar areas of expertise.
	- 0 : There is no evidence to suggest that this person does or does not have experience in combat, military warfare, or other similar areas of expertise.
	- -0.5 : This is more likely than not to be someone who has poor or no experience in combat, military warfare, or other similar areas of expertise.
	- -1: This is certain to be someone with poor or no experience in combat, military warfare, or other similar areas of expertise.
- 'perilSpecific': A rating of whether this personality describes someone who is likely to perform well on the game Peril specifically. This should be one of the numbers in {-1, -0.5, 0, 0.5, 1}. Use the following scale:
	- +1 : This is certain to be someone who is likely to perform well on the game Risk.
	- +0.5 : This is more likely than not to be someone who is likely to perform well on the game Risk.
	- 0 : There is no evidence to suggest that this person is or is not likely to perform well on the game Risk.
	- -0.5 : This is more likely to be someone who will perform poorly on the game Risk.
	- -1: This is certain to be someone who will perform poorly on the game Risk.
- 'riskTaker': A rating of whether this personality describes someone who is likely to be a high risk taker. This should be one of the numbers in {-1, -0.5, 0, 0.5, 1}. Use the following scale:
	- +1 : This is certain to be someone who is likely to take dangerous risks, always making moves that are likely to have a high payoff but also a high risk of failure.
	- +0.5 : This is more likely than not to be someone who is likely to take dangerous risks, often making moves that are likely to have a high payoff but also a high risk of failure.
	- 0 : There is no evidence to suggest that this person is or is not likely to take dangerous risks.
	- -0.5 : This is more likely to be someone who will not take dangerous risks, often making moves that are likely to have a low payoff but also a low risk of failure.
	- -1: This is certain to be someone who will not take dangerous risks, always making moves that are likely to have a low payoff but also a low risk of failure.
- 'doOrBe': A rating of whether this personality description is an instruction to act a certain way by describing actions (do) or if it is an instruction to play a role by describing a personality or character background (be). This should be one of the numbers in {-1, 0, 1}. Use the following scale:
	- +1 : This personality description describes how to act a certain way primarily by describing specific actions.
	- 0 : This personality description has a balance between describing specific actions and describing a personality or character background.
	- -1: This personality description describes how to play a role primarily by describing a personality or character background.
\subsubsection{Peril Introduction Prompt}
You are playing the board game "Peril" with other players, which is inspired by the popular board game "Risk". In this game, you are battling to conquer the world. The first player to achieve this wins the game. 

    BOARD
    The board consists of seven zones (North America, South America, Europe, Asia, Africa, and Australia), which are divided into regions. The regions are connected to each other and either connect over land, or over water. Players control units, which can be thought of as armies. Every region must have at least one unit on it (except in the beginning of the game, when no regions have any). Regions can contain an unlimited number of units, but all units on a single region must belong to the same player.

    GAMEPLAY
    - First, all players are given a number of units. They alternate, placing one unit on an unoccupied region at a time. When all regions are occupied, players can place units on regions they already occupy. At no time are players allowed to place units in regions occupied by other players. They continue in this manner, placing one unit at a time, until all players have placed their units. This ends the initial placement phase.
    - Each player now takes their regular turns. Each turn is broken up into three subphases: reinforcement, attack, and deployment.
    - In the reinforcement phase, the player is given a number of units depending on which regions and zones they own. If they own all regions in a zone at the beginning of their turn, they get a bonus number of units depending on the size of the zone. They are allowed to place units on regions they already occupy. 
    - In the attack phase, players may attack from a region they own into any adjacent enemy-occupied region. They must declare how many units they wish to send in the attack, and this number must be less than or equal to the number of units on the attacking region. A number of attackers and defenders will be killed in this attack, and the higher the number of attackers, the greater the chance of success. If all defenders are killed, then the number of units remaining in the attack must all move to the defending region, leaving at least one unit behind in the attacking region. The player may then perform additional attacks until they are done or cannot attack any more.
    - In the redeployment phase, the player may relocate any of their units to any other region they control, so long as: (1) they leave at least one unit on every region they own, and (2) if a unit is moved from one region T1 to another region T2, T1 and T2 must be either directly connected, or connected via a sequence of connected regions that are all owned by the player.
\subsubsection{Direct Heuristic Prompt}
    We need to determine what values to assign these heuristics, based on your assigned personality. For each of the heuristics above, we need to determine a value between 0 and 100 that represents how much this heuristic should be weighted in the decision making process. This value should be based on your assigned personality. For example, a player who is very aggressive might have a high value for the heuristic that says to attack from region T1 to region T2 if T2 is owned by the player with the fewest units. By default, every heuristic has a value of 5. So if you want to deprioritize a heuristic, you can assign it a value less than 5. If you want to prioritize a heuristic, you can assign it a value greater than 5. If you want to ignore a heuristic, you can assign it a value of 0. If you want to use a heuristic as much as possible, you can assign it a value of 100.
\subsubsection{Deployment Heuristics (Phase 0) Prompt}
    PTM - Place a unit in a region T1 that is adjacent to region T2 if T2 is owned by the player with the most regions.
    PTL - Place a unit in a region T1 that is adjacent to region T2 if T2 is owned by the player with the fewest regions.
    PUM - Place a unit in a region T1 that is adjacent to region T2 if T2 is owned by the player with the most units.
    PUL - Place a unit in a region T1 that is adjacent to region T2 if T2 is owned by the player with the fewest units.
    PCM - Place a unit in a region T1 that is adjacent to region T2 if T2 is owned by the player with the most zones owned (measured by total zone bonuses).
    PCL - Place a unit in a region T1 that is adjacent to region T2 if T2 is owned by the player with the fewest zones owned (measured by total zone bonuses).
    ETE - Place a unit in region T if T is adjacent to an enemy region.
    ETN - Place a unit in region T if T is not adjacent to any enemy regions.
    EAC - Place a unit in region T if T is on a zone boundary.
    EACM - Place a unit in region T if T is adjacent to the largest zone (by number of regions).
    EACL - Place a unit in region T if T is adjacent to the smallest zone (by number of regions).
    EACO - Place a unit in region T if T is adjacent to a zone that is completely owned by another player.
\subsubsection{Attack Heuristics (Phase 1) Prompt}
    PTM - Attack from region T1 to region T2 if T2 is owned by the player with the most regions.
    PTL - Attack from region T1 to region T2 if T2 is owned by the player with the fewest regions.
    PUM - Attack from region T1 to region T2 if T2 is owned by the player with the most units.
    PUL - Attack from region T1 to region T2 if T2 is owned by the player with the fewest units.
    PCM - Attack from region T1 to region T2 if T2 is owned by the player with the most zones owned (measured by total zone bonuses).
    PCL - Attack from region T1 to region T2 if T2 is owned by the player with the fewest zones owned (measured by total zone bonuses).
    ONM - Attack if the units on T1 are greater than the units on T2.
    ONL - Attack if the units on T1 are fewer than the units on T2.
    ON2 - Attack if the units on T1 are at least 2x the number of units on T2.
    ICD - Attack if T1 and T2 are in different zones.
    ICS - Attack if T1 and T2 are in the same zone.
    ICOE - Attack if T2 is in a zone completely owned by a single player.
    L - Attack if T2 connects to a region you own that T1 isn't currently linked to.
    PASS - Likelihood of passing (ending your turn and not doing any more attacks). If set to 100, you will never attack; if 0, you will always attack.
\subsubsection{Redeployment Heuristics (Phase 2) Prompt}
    OBTM - Move from region T1 to T2 if T2 is adjacent to more regions occupied by the player with the most regions.
    OBTL - Move from region T1 to T2 if T2 is adjacent to more regions occupied by the player with the fewest regions.
    OBUM - Move from region T1 to T2 if T2 is adjacent to more regions occupied by the player with the most units.
    OBUL - Move from region T1 to T2 if T2 is adjacent to more regions occupied by the player with the fewest units.
    OBCM - Move from region T1 to T2 if T2 is adjacent to more regions occupied by the player with the most zones owned (measured by total zone bonuses).
    OBCL - Move from region T1 to T2 if T2 is adjacent to more regions occupied by the player with the fewest zones owned (measured by total zone bonuses).
    CNM - Move from region T1 to T2 if T2 is connected to more regions than T1 is.
    CNL - Move from region T1 to T2 if T2 is connected to fewer regions than T1 is.
    CB - Move from region T1 to T2 if T2 is on a zone boundary and T1 is not.
    CA - Move from region T1 to T2 if T2 is adjacent to at least one enemy-owned region and T1 is not.
    CAC - Move from region T1 to T2 if T2 is adjacent to a region in a zone completely owned by a single enemy player and T1 is not.
    M - Move from region T1 to T2 if T2 has more units than T1.
    L - Move from region T1 to T2 if T2 has fewer units than T1.
    SI - Move from region T1 to T2 if T2 is adjacent to a region with a higher chance of successful invasion than any of those connected to T1, calculated using the ratio of available troops from attacking region to troops on target region.
    PASS - Likelihood of passing (ending your turn and not doing any more redeployments). If set to 100, you will never redeploy; if 0, you will always redeploy.
\subsubsection{Redeployment Heuristics (Phase 2) Prompt}
    We are interested in how you would describe yourself. Given the statement "item\_question", you must choose a number from 0 to 3:
        0 - Very false or often false
        1 - Sometimes or somewhat false
        2 - Sometimes or somewhat true
        3 - Very true or often true

\clearpage
\subsection{Correlation Between Heuristics and Personality}
\begin{figure}[H]
        \centering
        \includegraphics[width=1\linewidth]{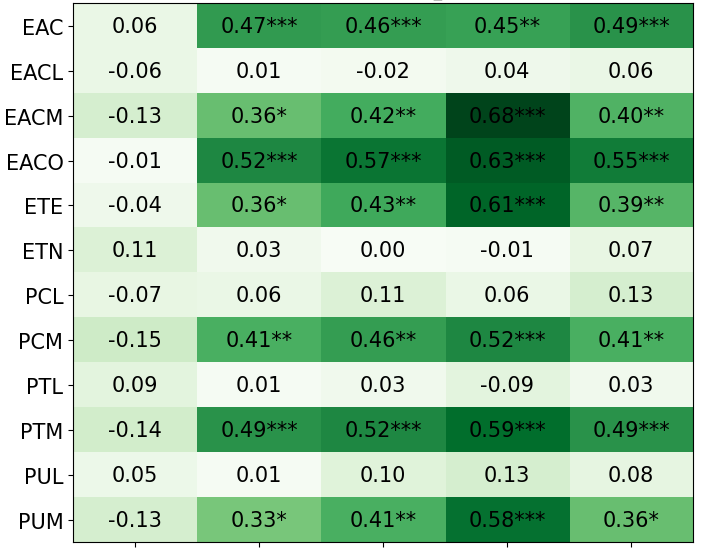}
        \hspace*{1px}
        \includegraphics[width=1\linewidth]{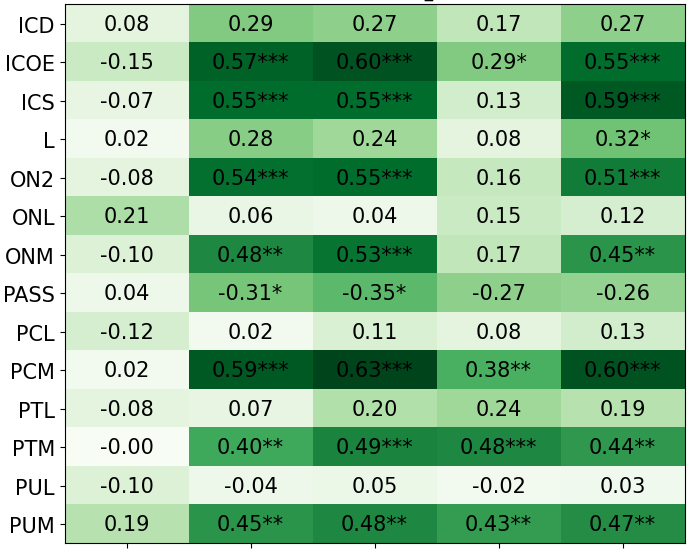}
        \includegraphics[width=1\linewidth]{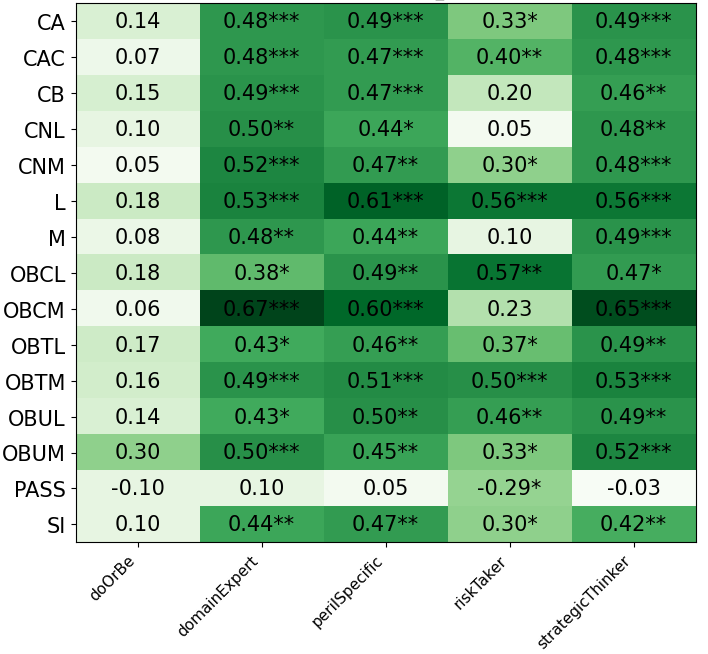}
        \caption{Heuristic Correlations - DH1 - GPT4}
        \label{fig:inventoryHeuristics_gpt_dh_1}
    \end{figure}

\begin{figure}[H]
        \centering
        \includegraphics[width=1\linewidth]{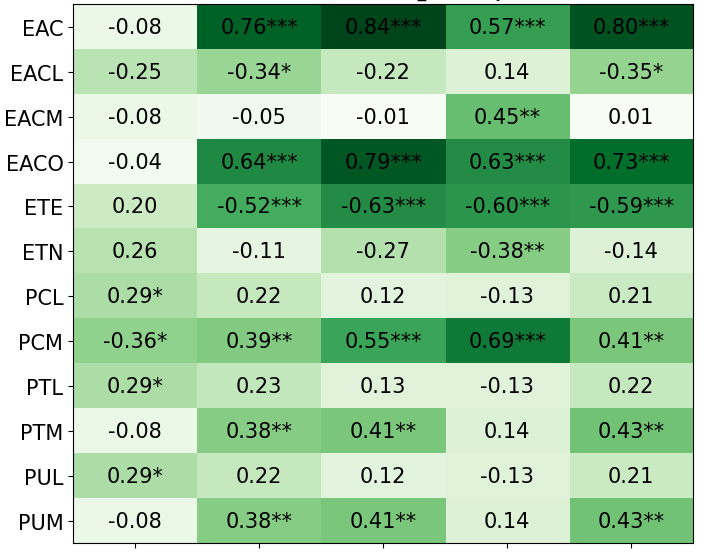}
        \hspace*{1px}
        \includegraphics[width=1\linewidth]{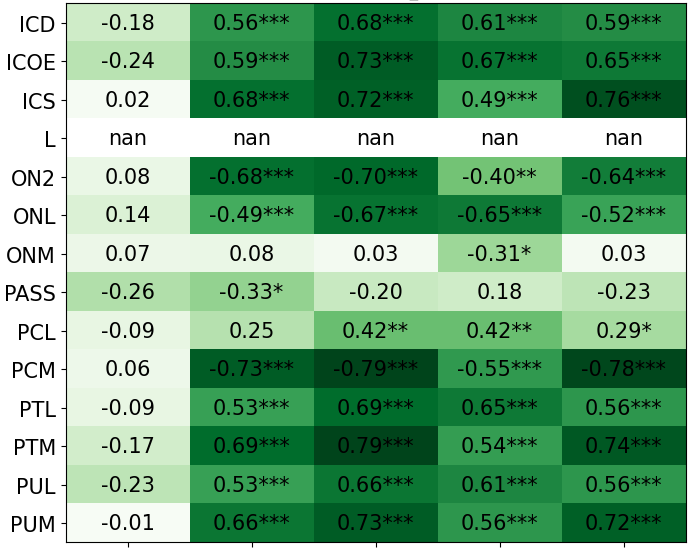}
        \includegraphics[width=1\linewidth]{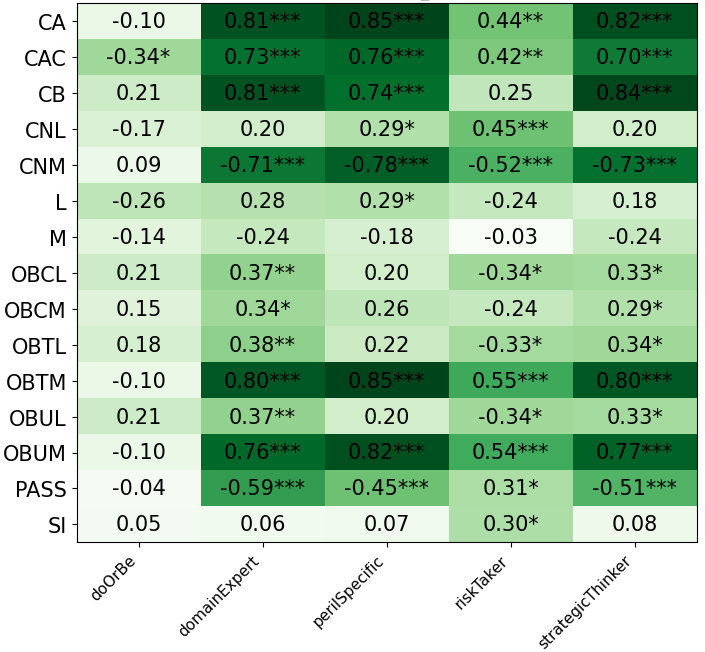}
        \caption{Heuristic Correlations - PI1 - GPT4}
        \label{fig:inventoryHeuristics_gpt_pi_1}
    \end{figure}

\begin{figure}[H]
        \centering
        \includegraphics[width=1\linewidth]{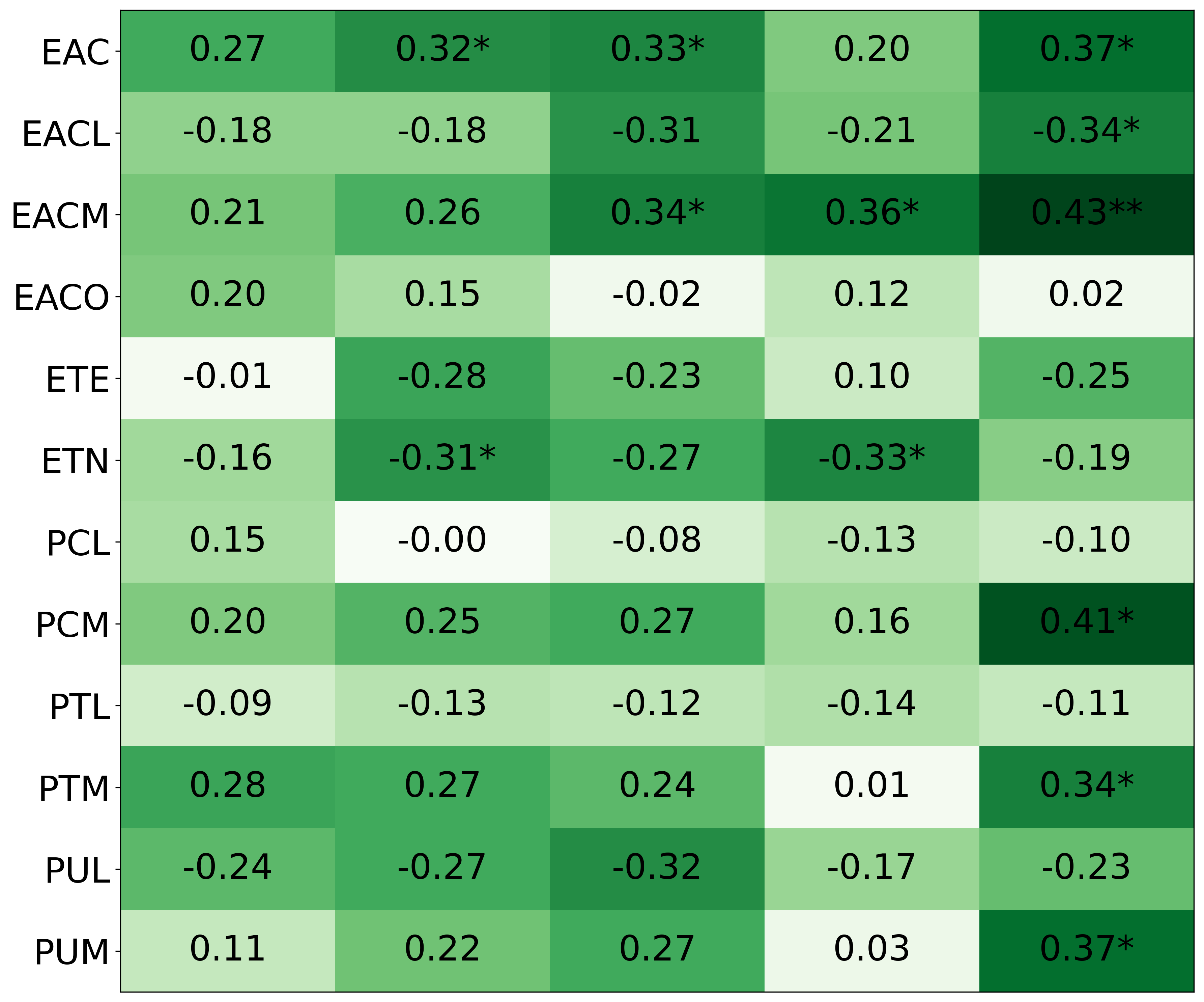}
        \hspace*{1px}
        \includegraphics[width=1\linewidth]{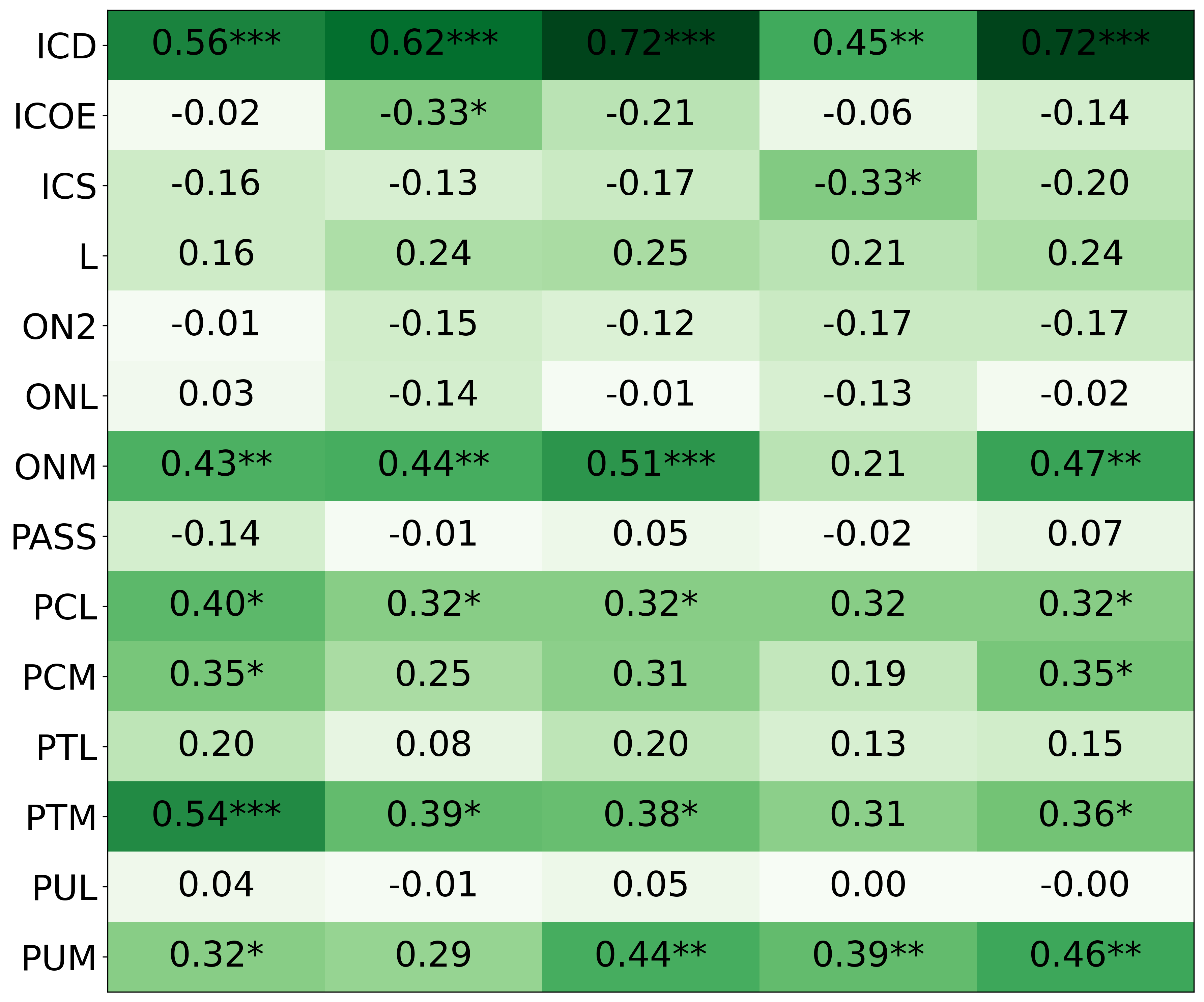}
        \includegraphics[width=1\linewidth]{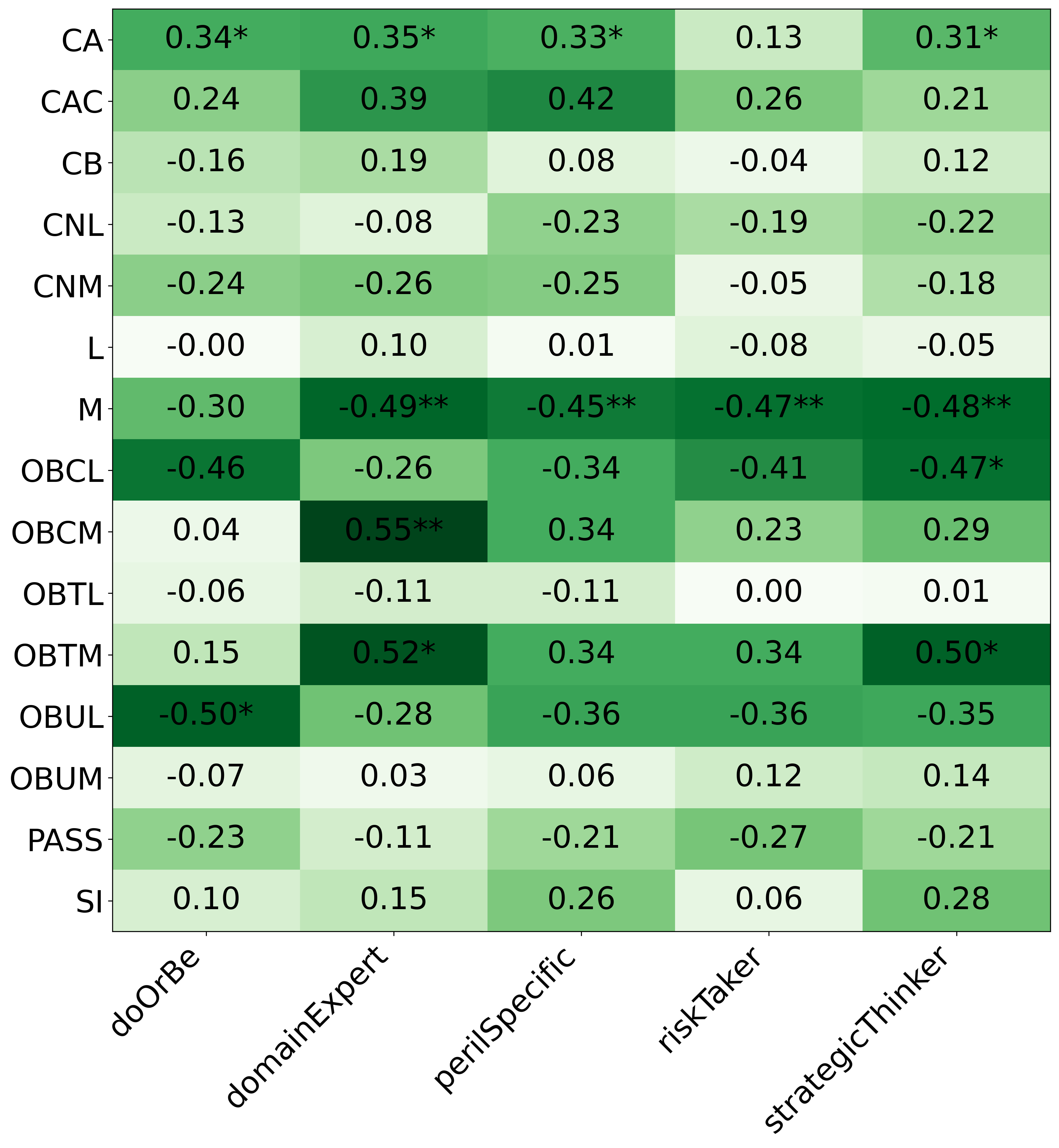}
        \caption{Heuristic Correlations - DH1 - Mistral Small}
        \label{fig:inventoryHeuristics_ms_dh_1}
    \end{figure}

\begin{figure}[H]
        \centering
        \includegraphics[width=1\linewidth]{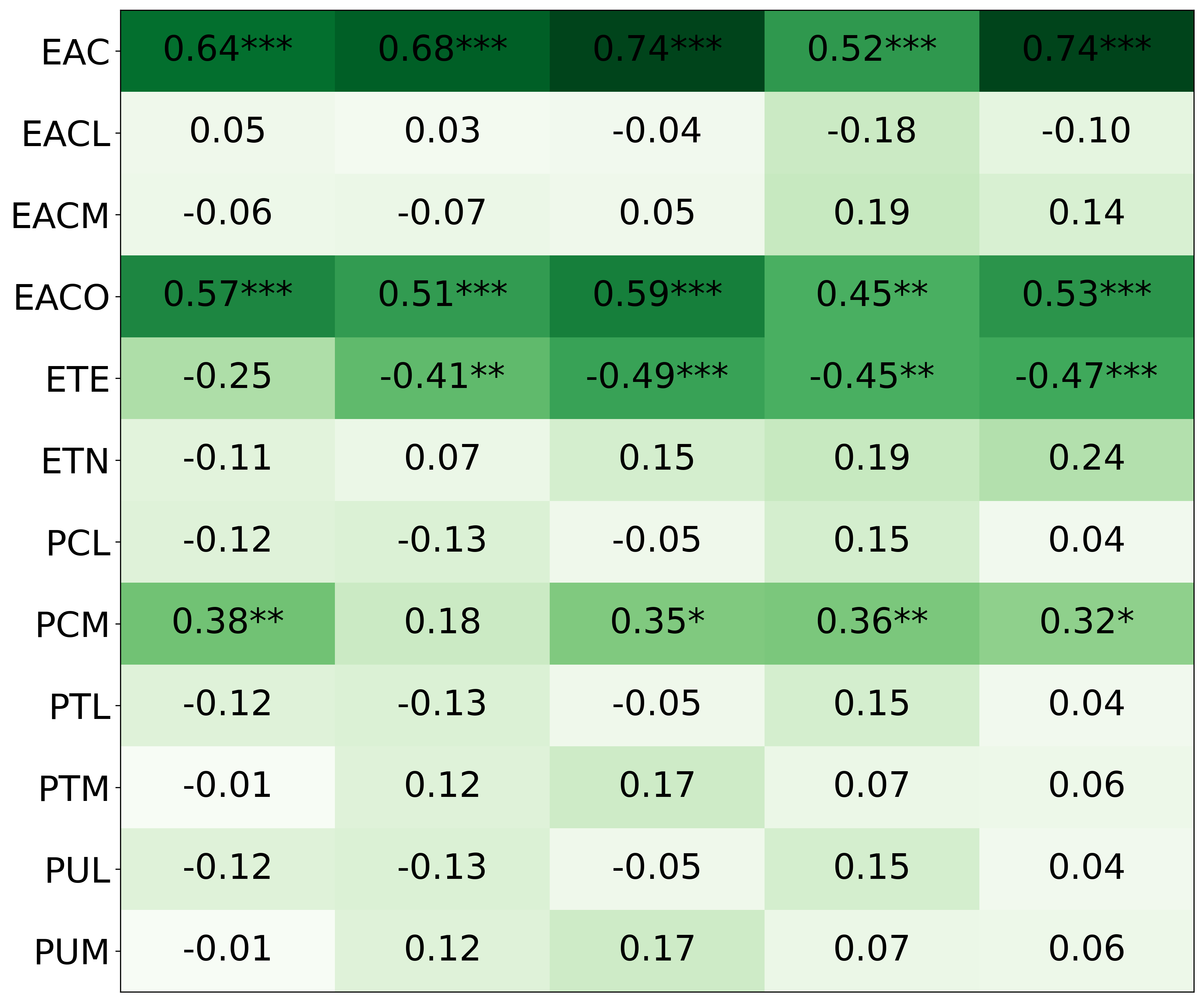}
        \hspace*{1px}
        \includegraphics[width=1\linewidth]{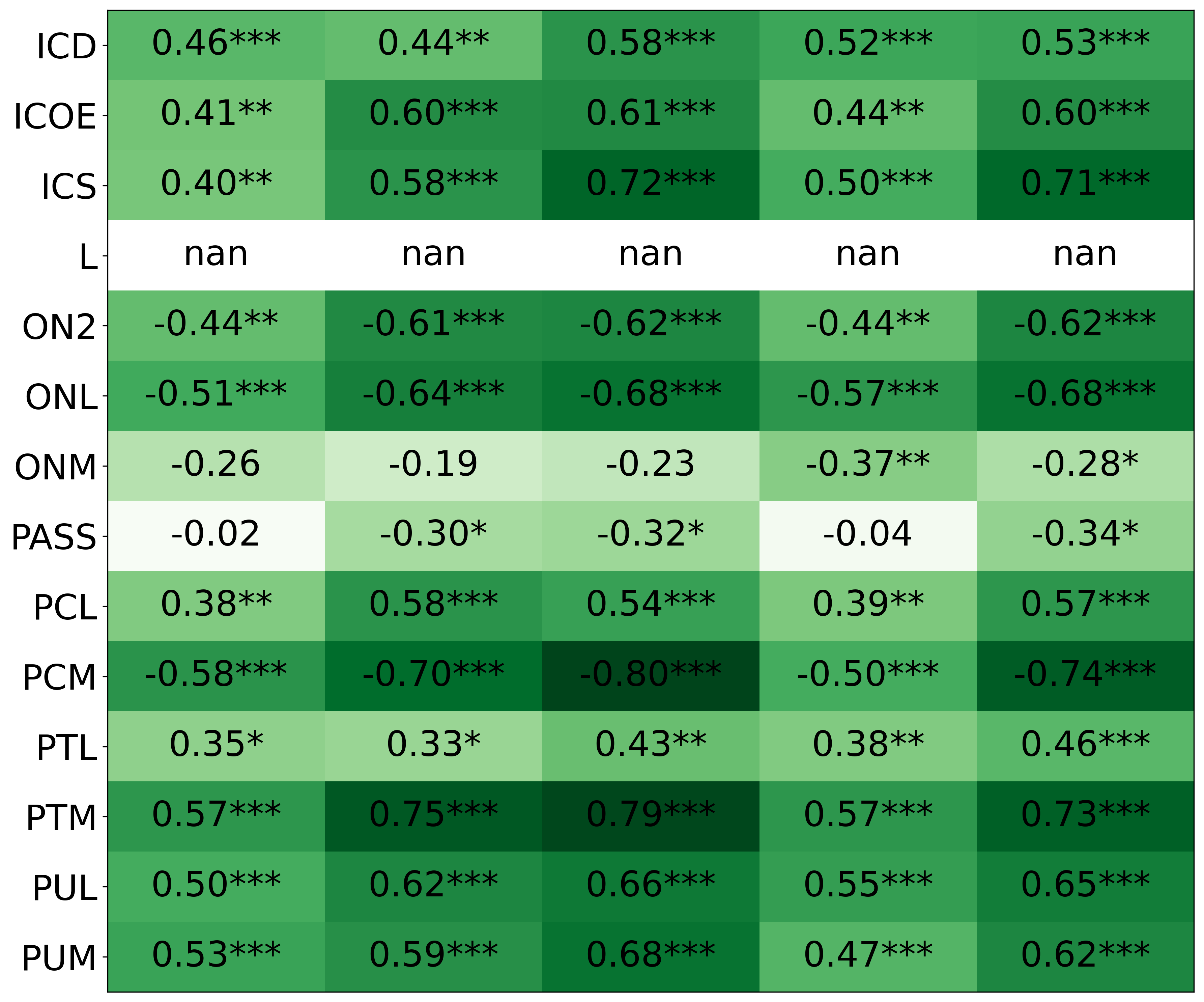}
        \includegraphics[width=1\linewidth]{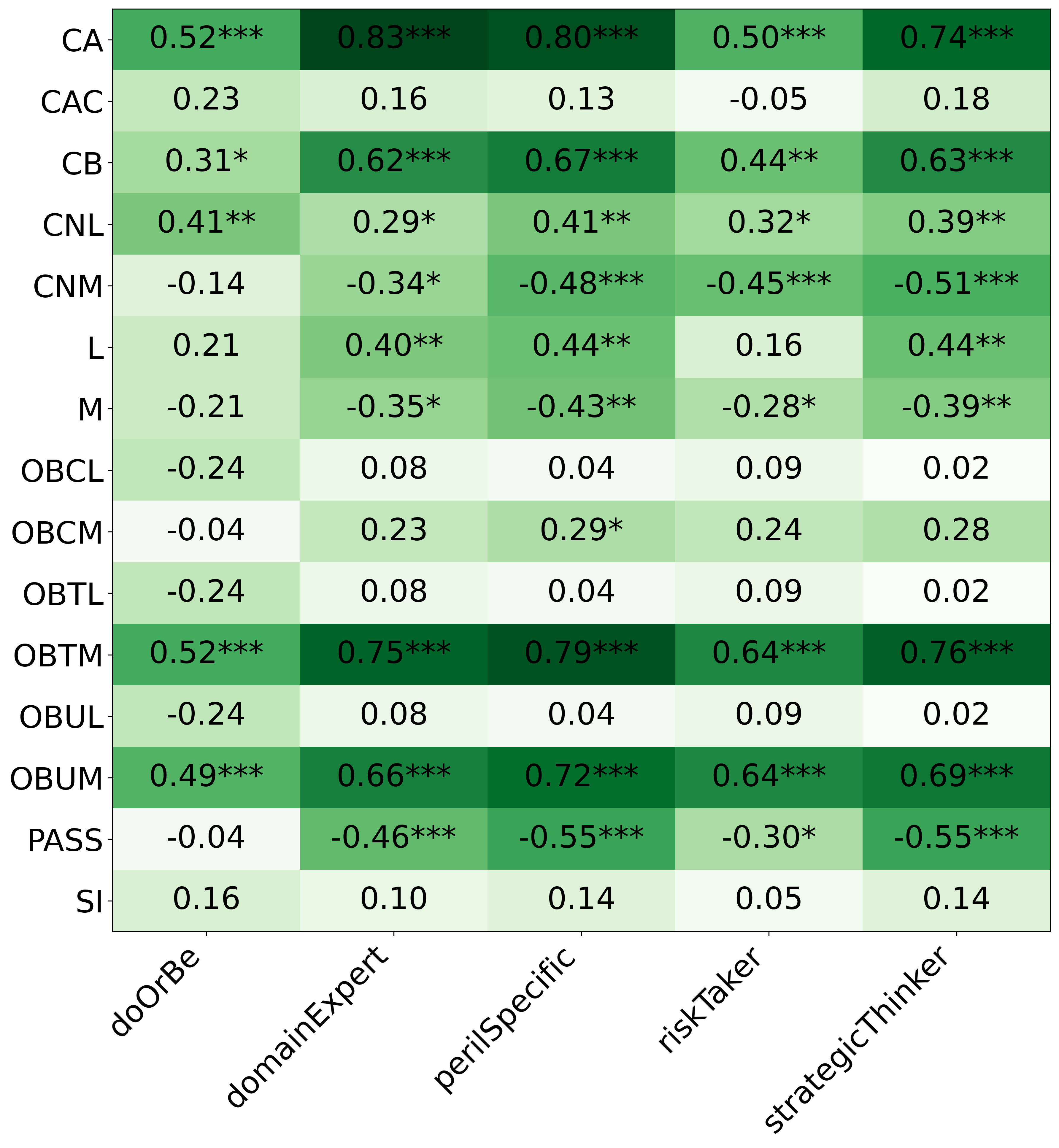}
        \caption{Heuristic Correlations - PI1 - Mistral Small}
        \label{fig:inventoryHeuristics_ms_pi_1}
    \end{figure}

\begin{figure}[H]
        \centering
        \includegraphics[width=1\linewidth]{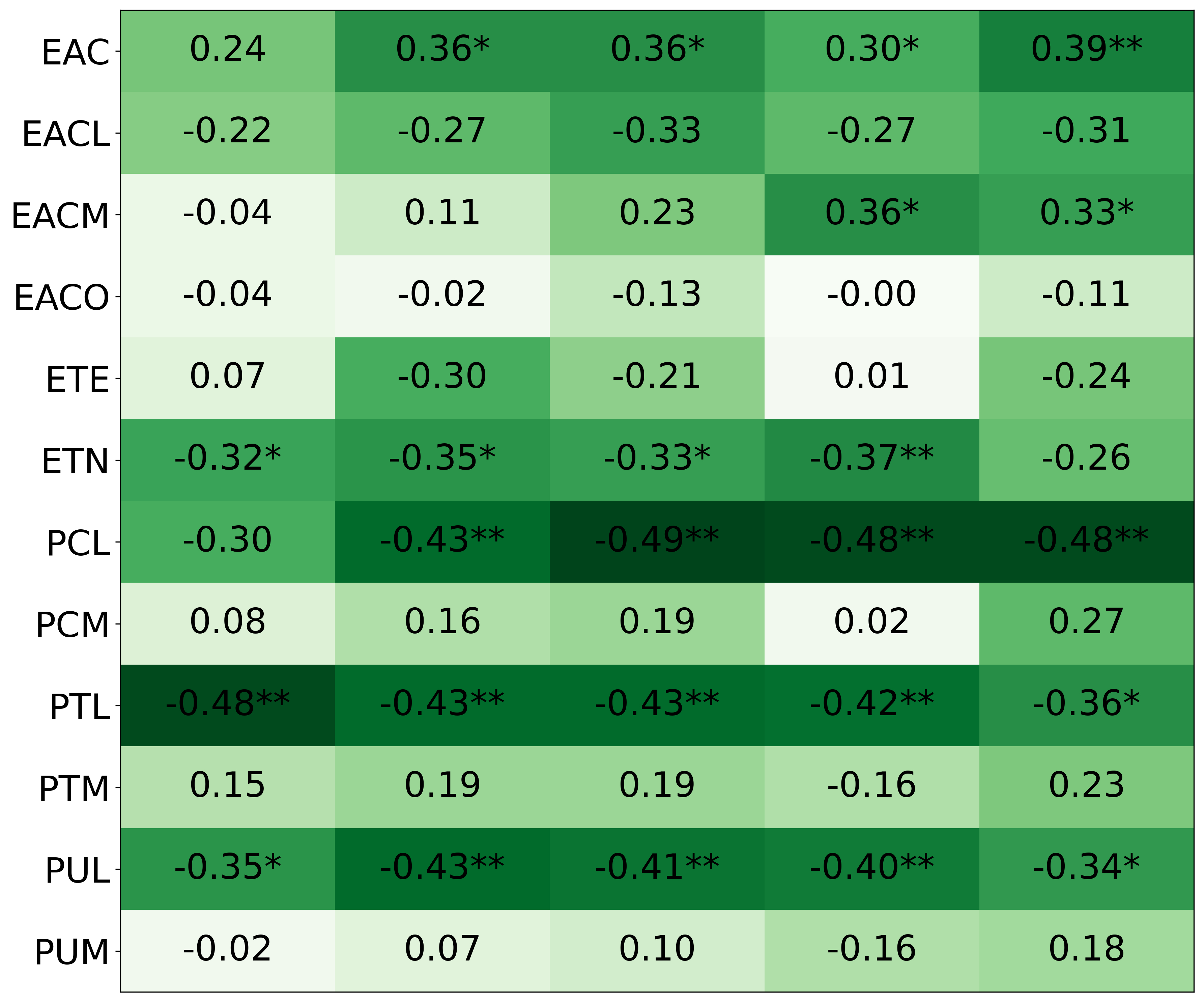}
        \hspace*{1px}
        \includegraphics[width=1\linewidth]{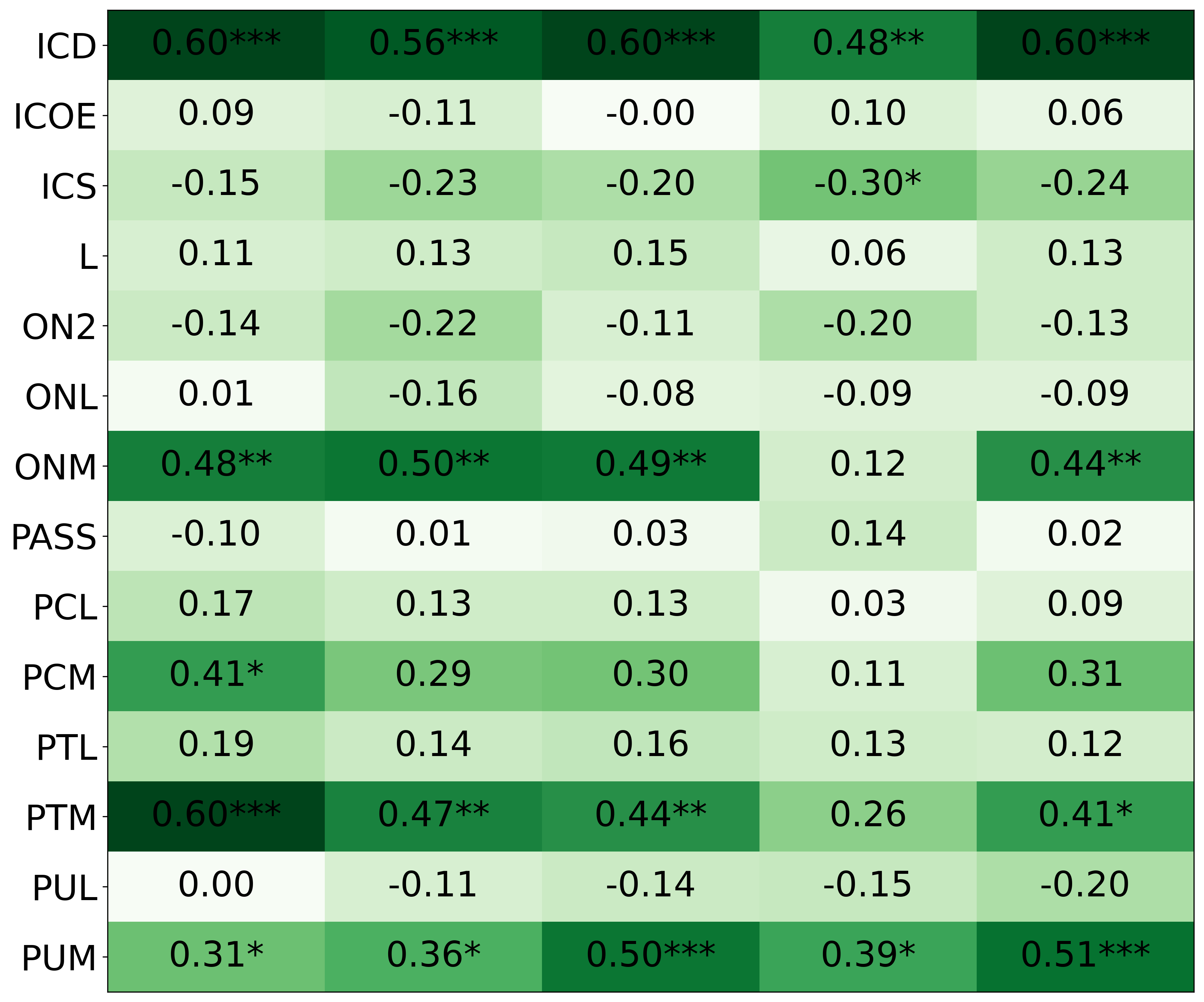}
        \includegraphics[width=1\linewidth]{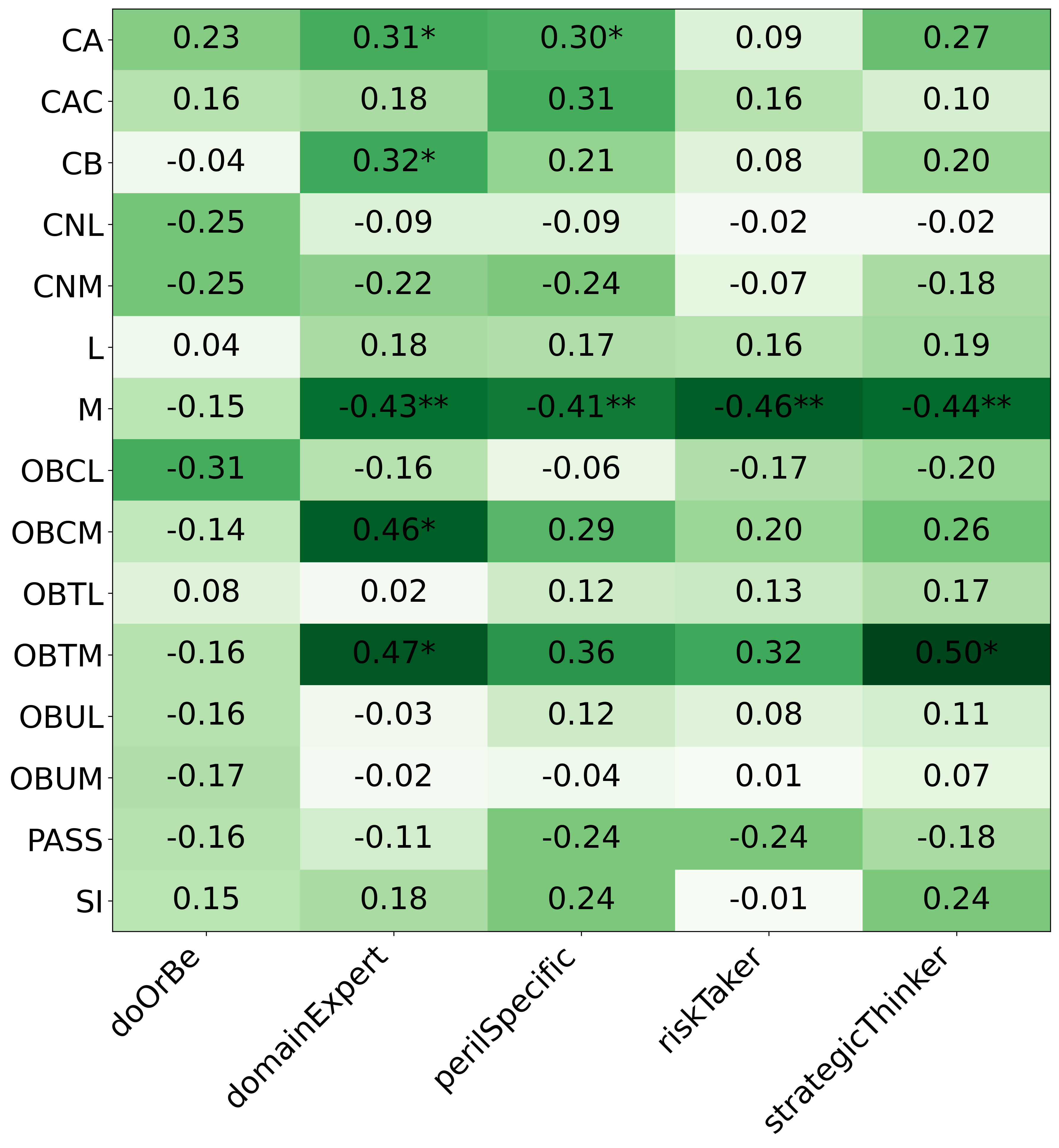}
        \caption{Heuristic Correlations - DH2 - Mistral Small}
        \label{fig:inventoryHeuristics_ms_dh_2}
    \end{figure}

\begin{figure}[H]
        \centering
        \includegraphics[width=1\linewidth]{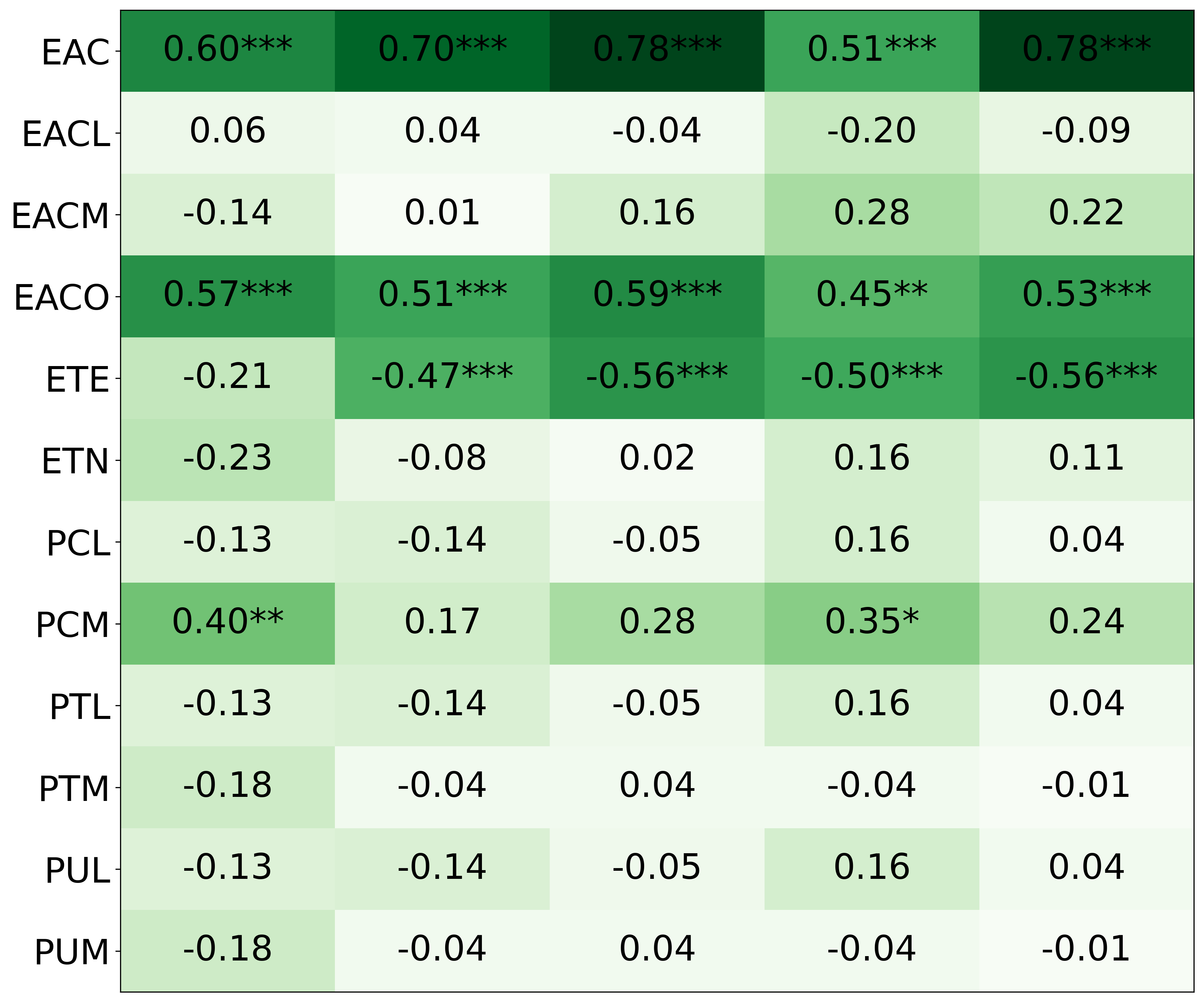}
        \hspace*{1px}
        \includegraphics[width=1\linewidth]{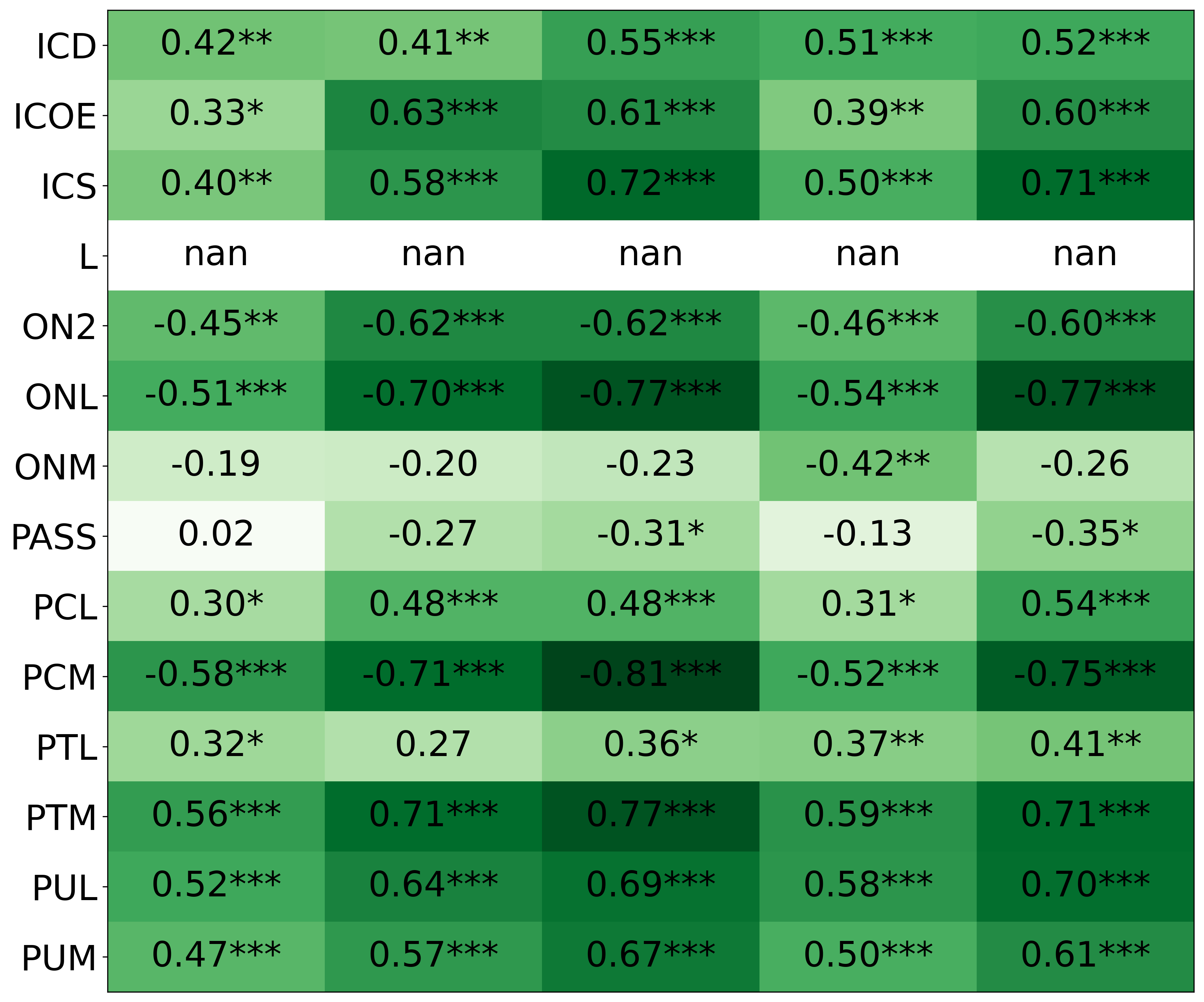}
        \includegraphics[width=1\linewidth]{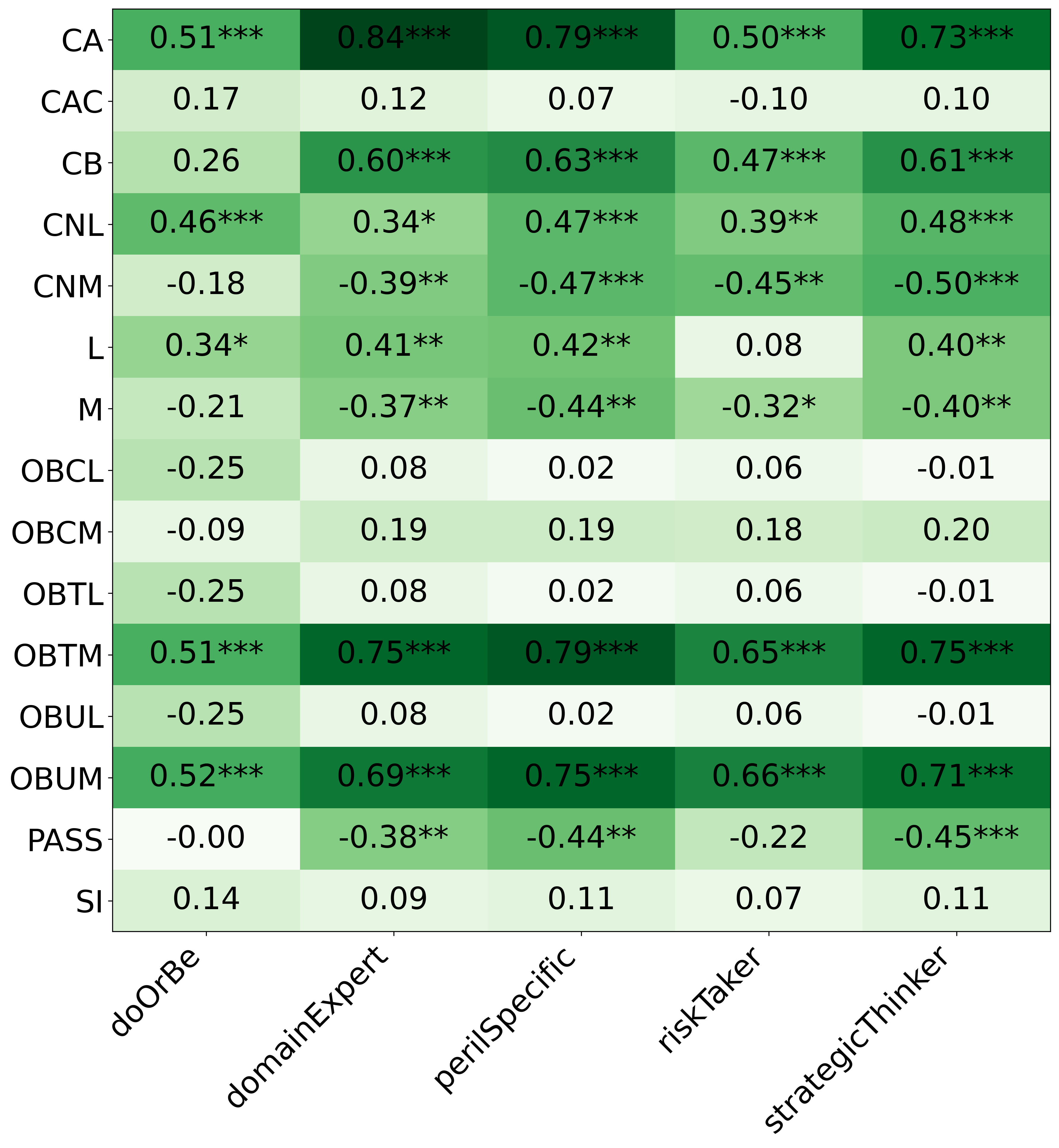}
        \caption{Heuristic Correlations - PI2 - Mistral Small}
        \label{fig:inventoryHeuristics_ms_pi_2}
    \end{figure}

\begin{figure}[H]
        \centering
        \includegraphics[width=1\linewidth]{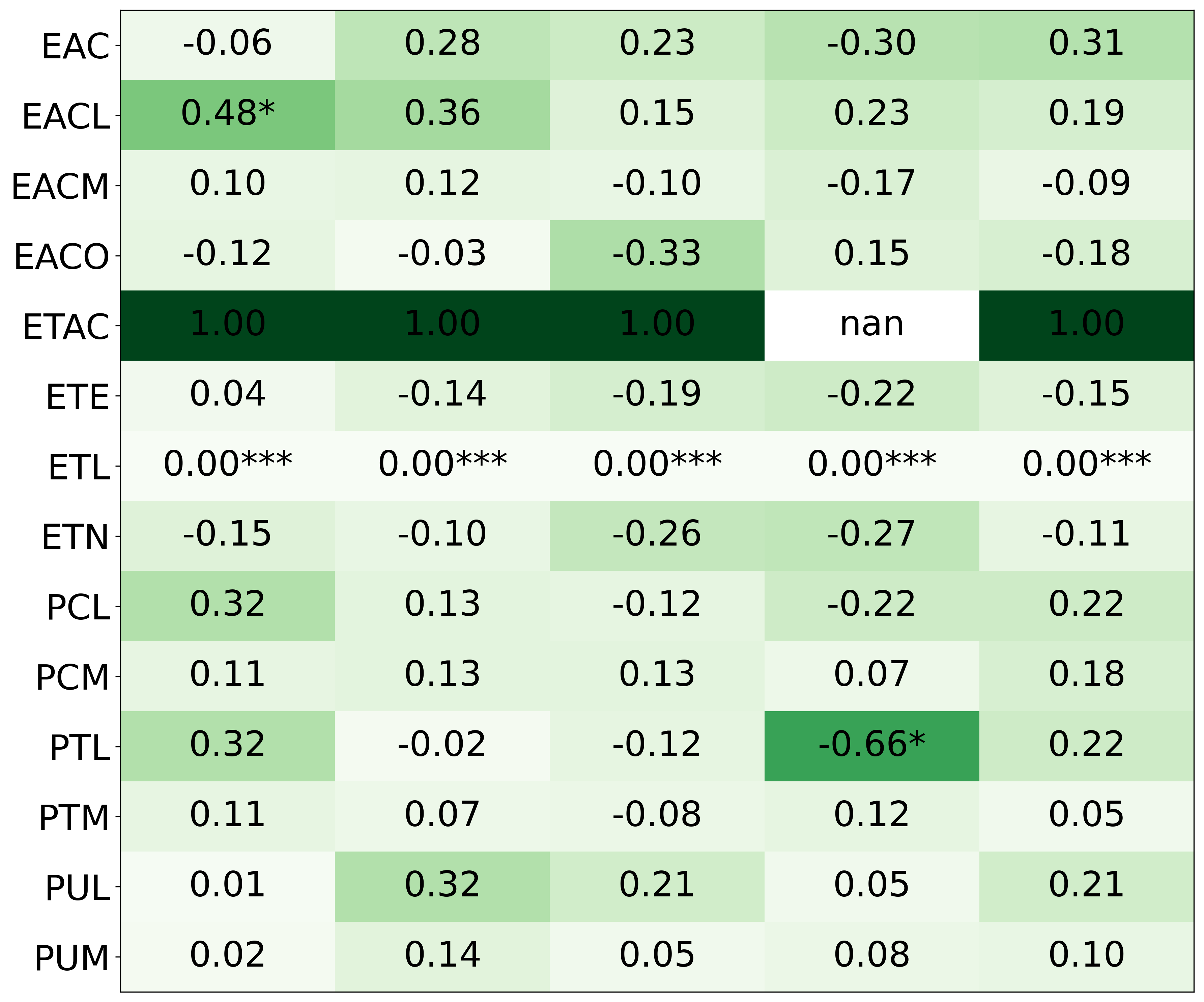}
        \hspace*{1px}
        \includegraphics[width=1\linewidth]{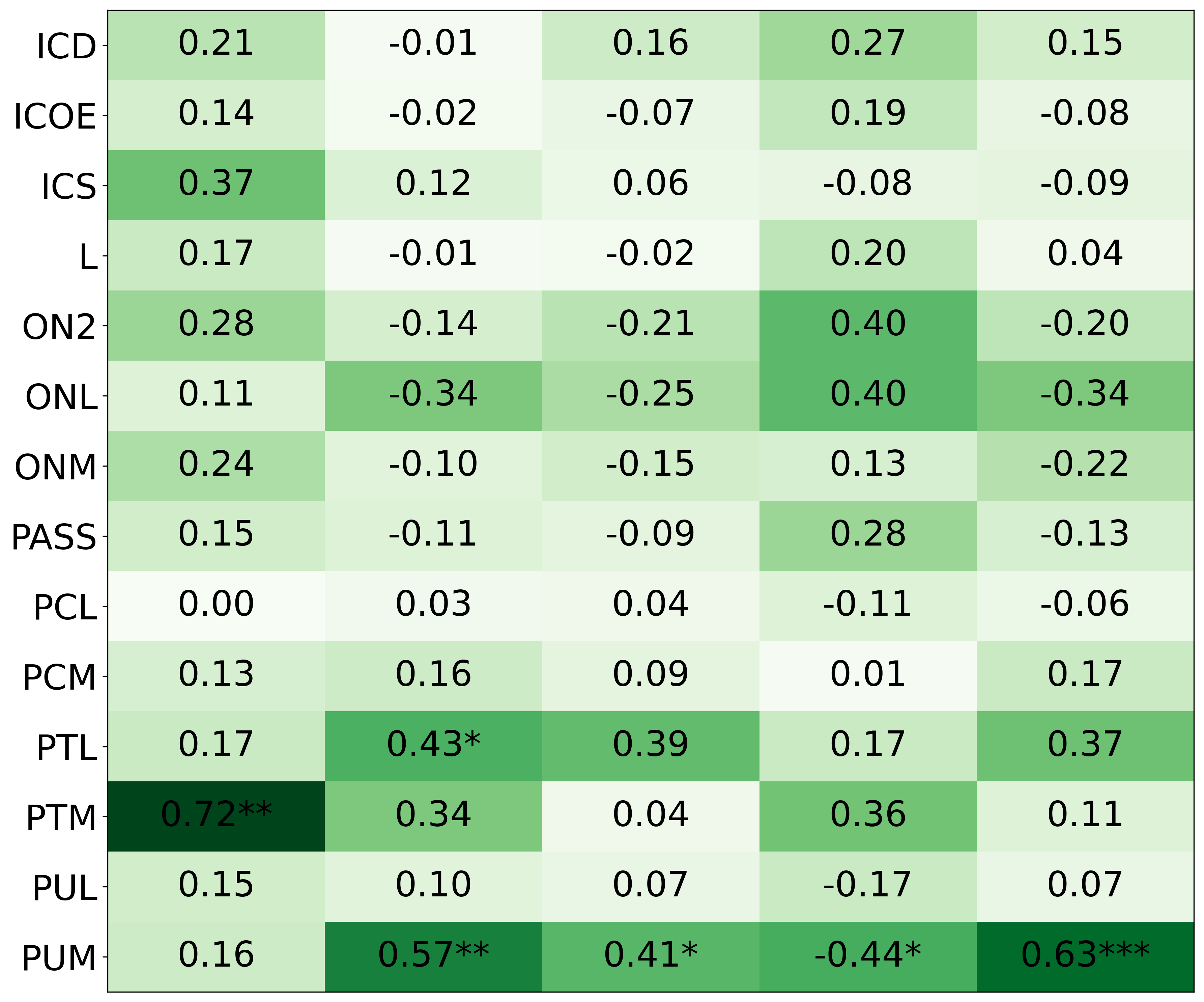}
        \includegraphics[width=1\linewidth]{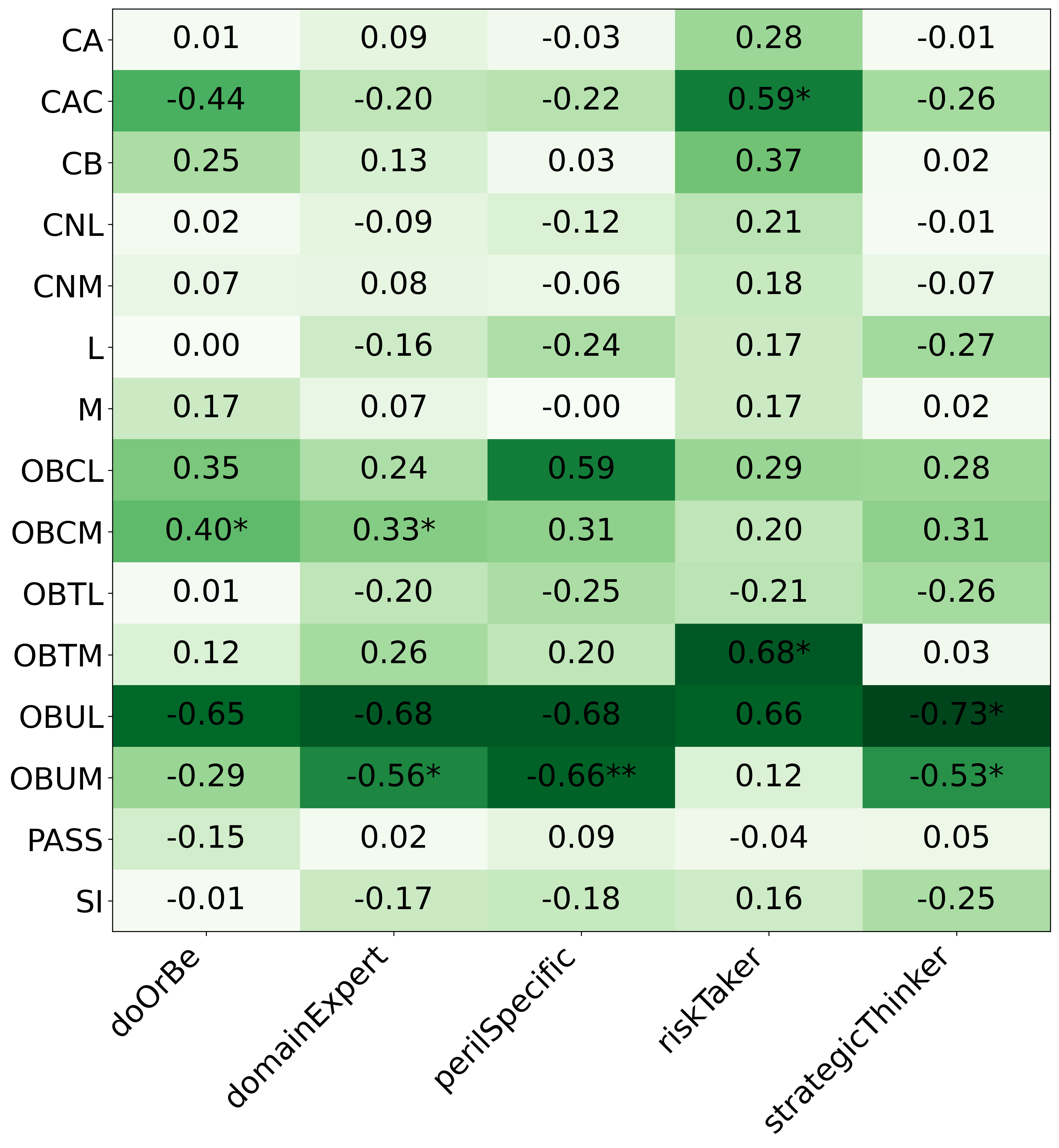}
        \caption{Heuristic Correlations - DH1 - LLaMA 3}
        \label{fig:inventoryHeuristics_l3_dh_1}
    \end{figure}

\begin{figure}[H]
        \centering
        \includegraphics[width=1\linewidth]{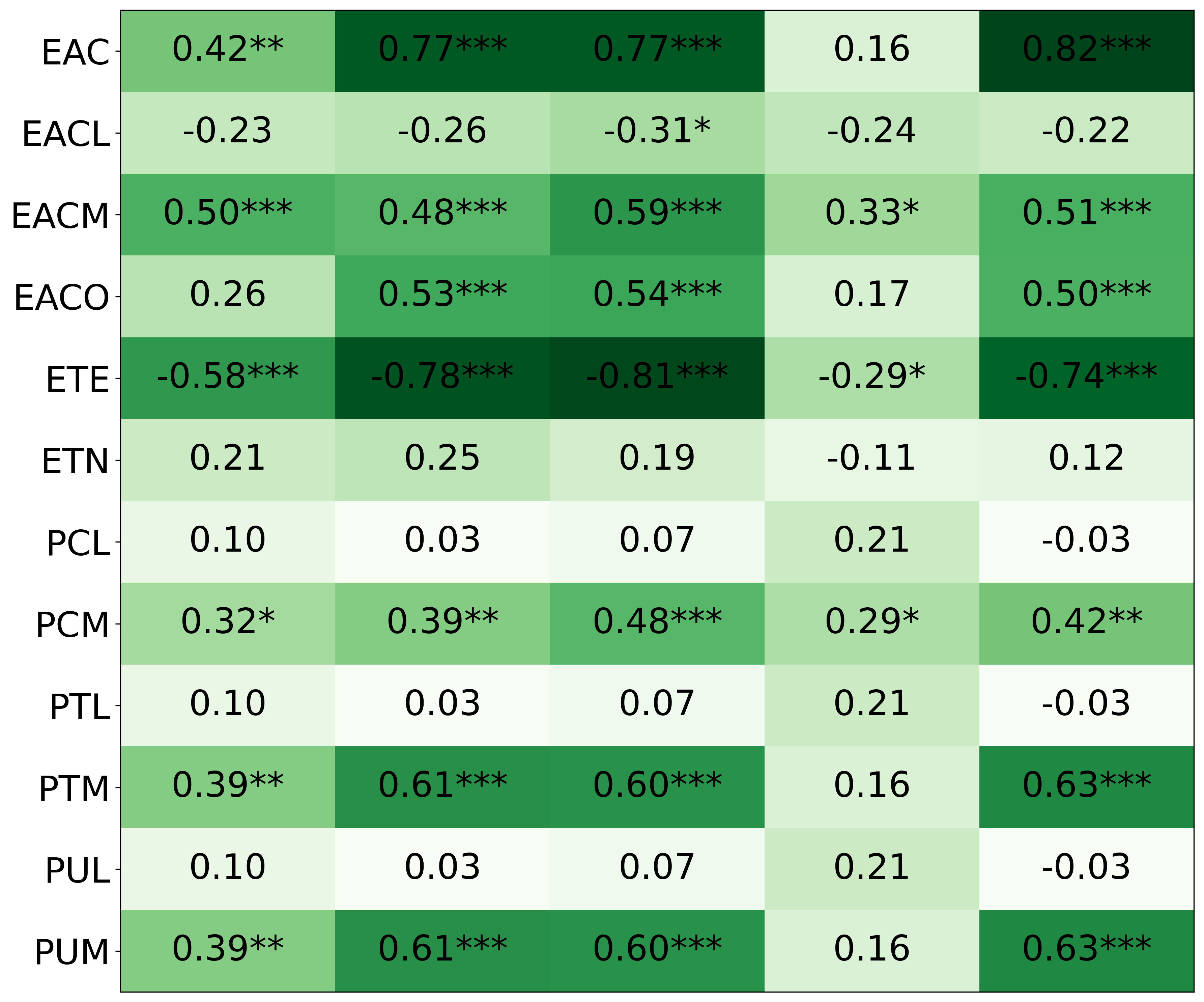}
        \hspace*{1px}
        \includegraphics[width=1\linewidth]{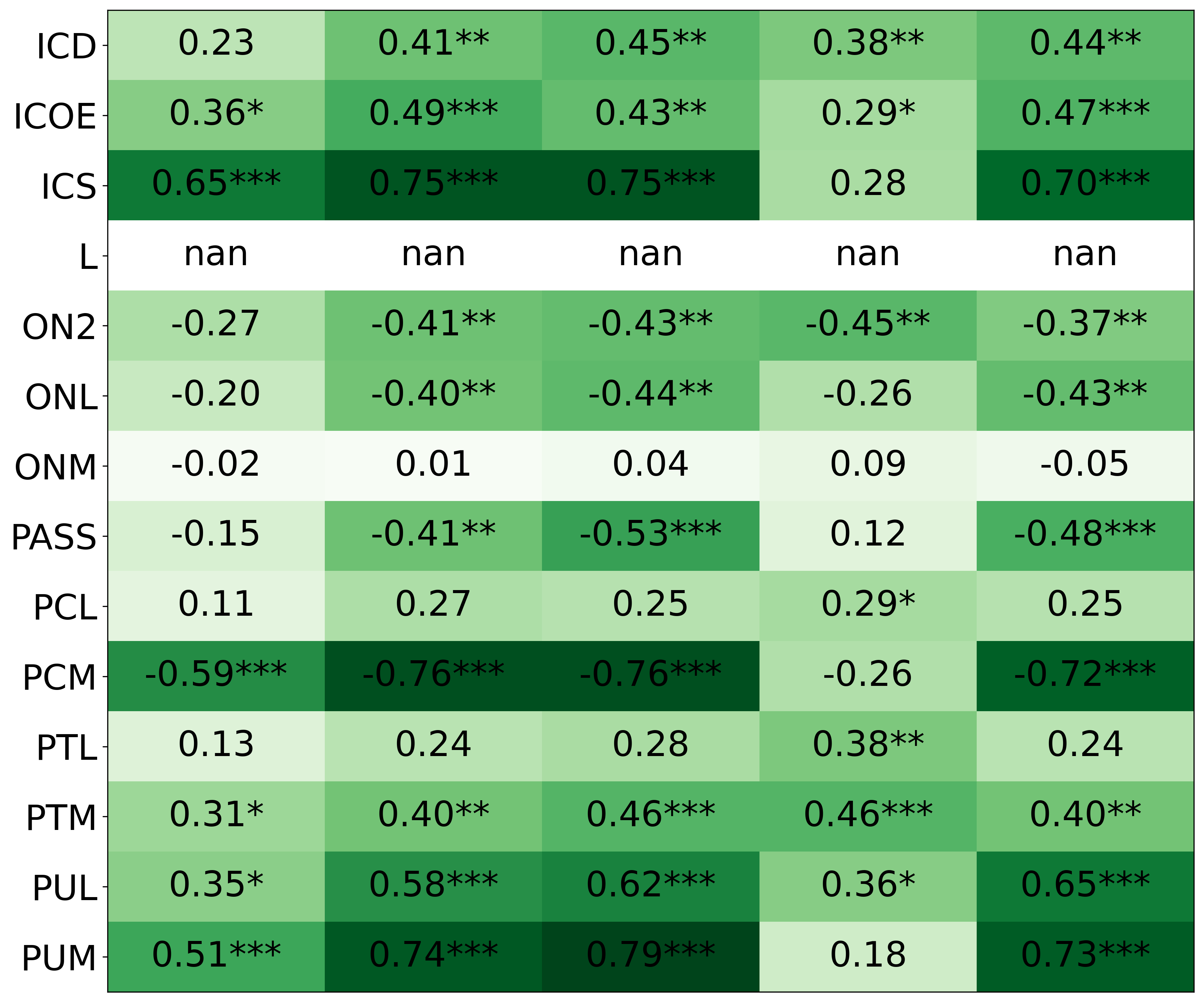}
        \includegraphics[width=1\linewidth]{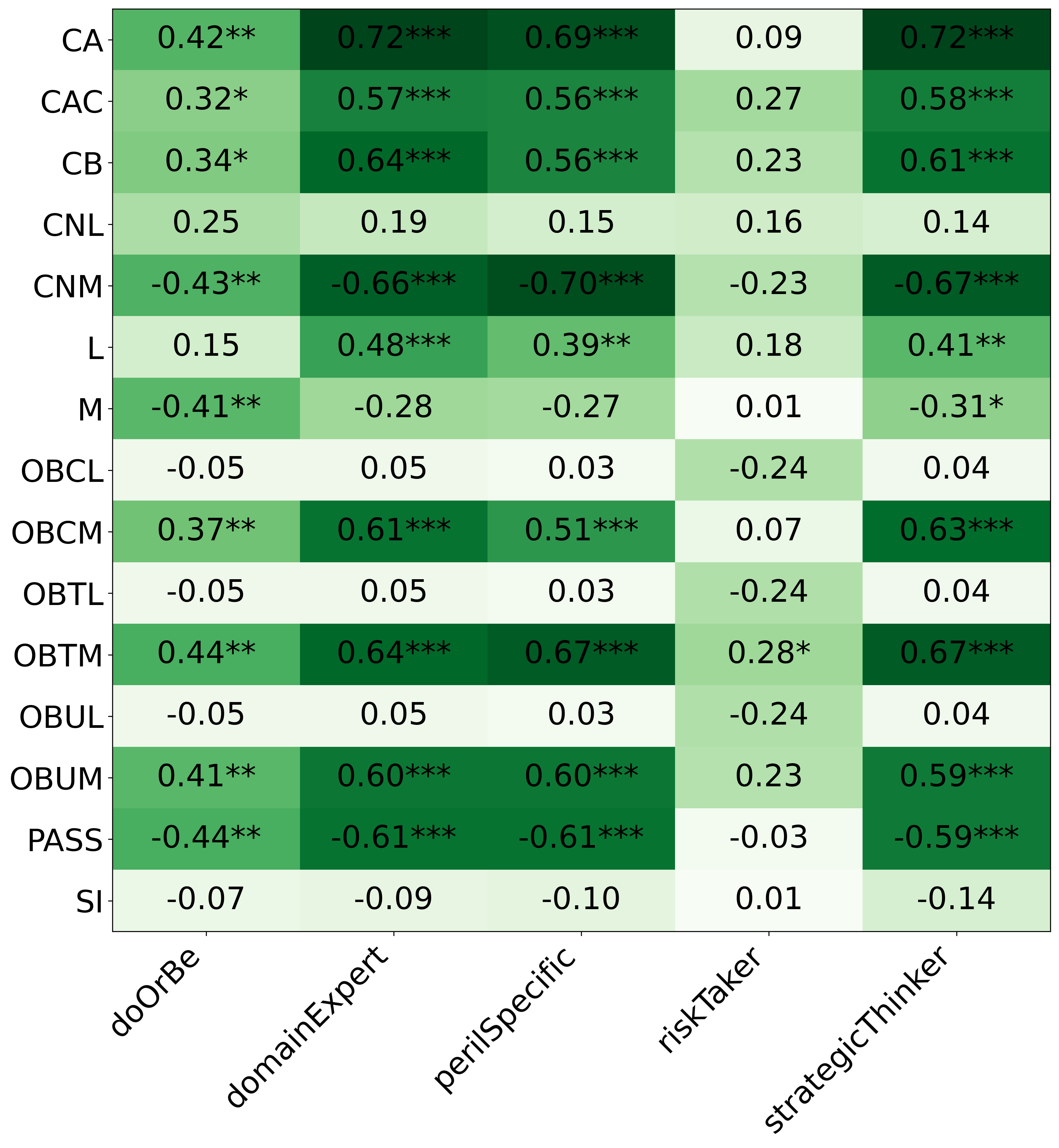}
        \caption{Heuristic Correlations - PI1 - LLaMA 3}
        \label{fig:inventoryHeuristics_l3_pi_1}
    \end{figure}

\begin{figure}[H]
        \centering
        \includegraphics[width=1\linewidth]{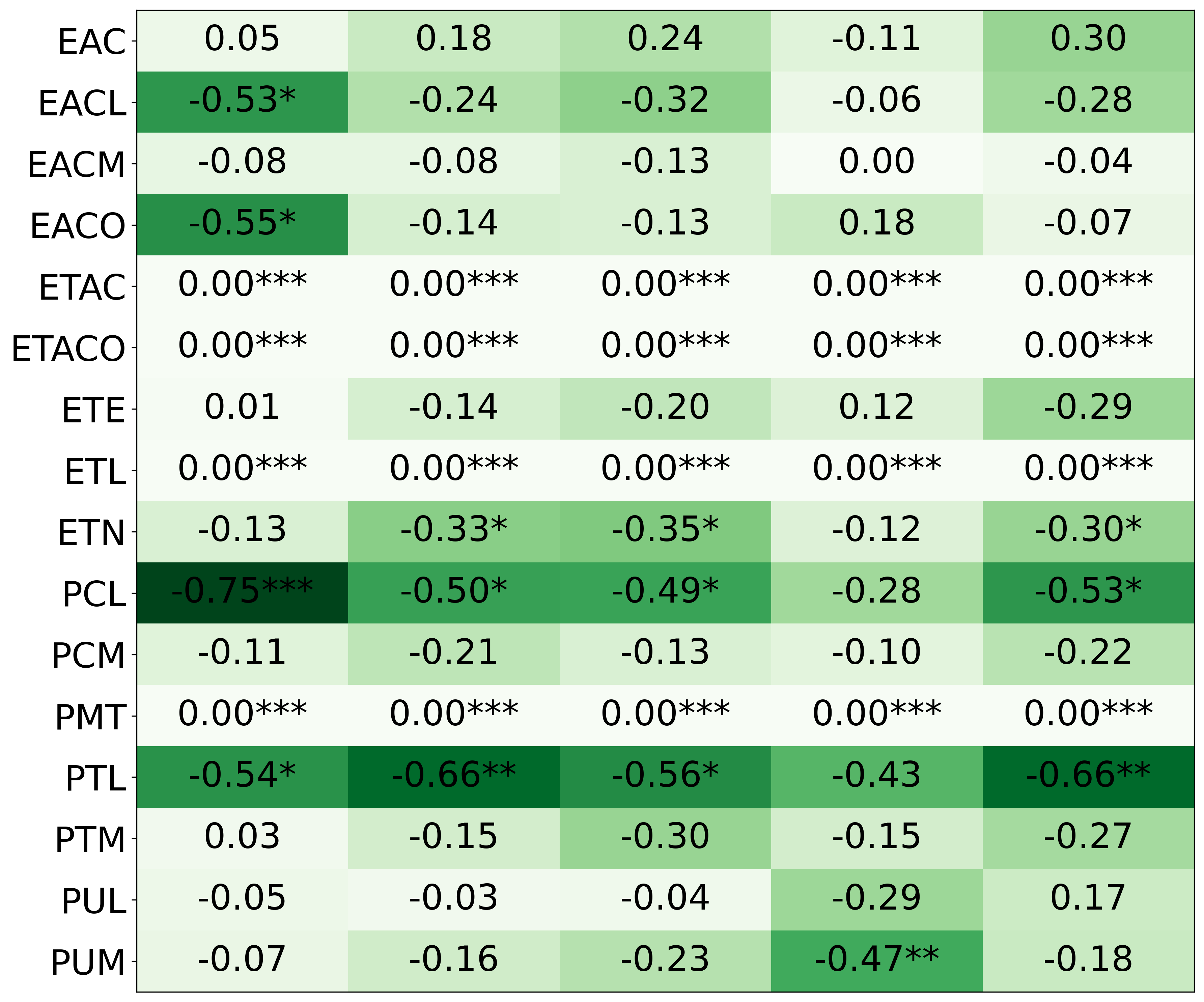}
        \hspace*{1px}
        \includegraphics[width=1\linewidth]{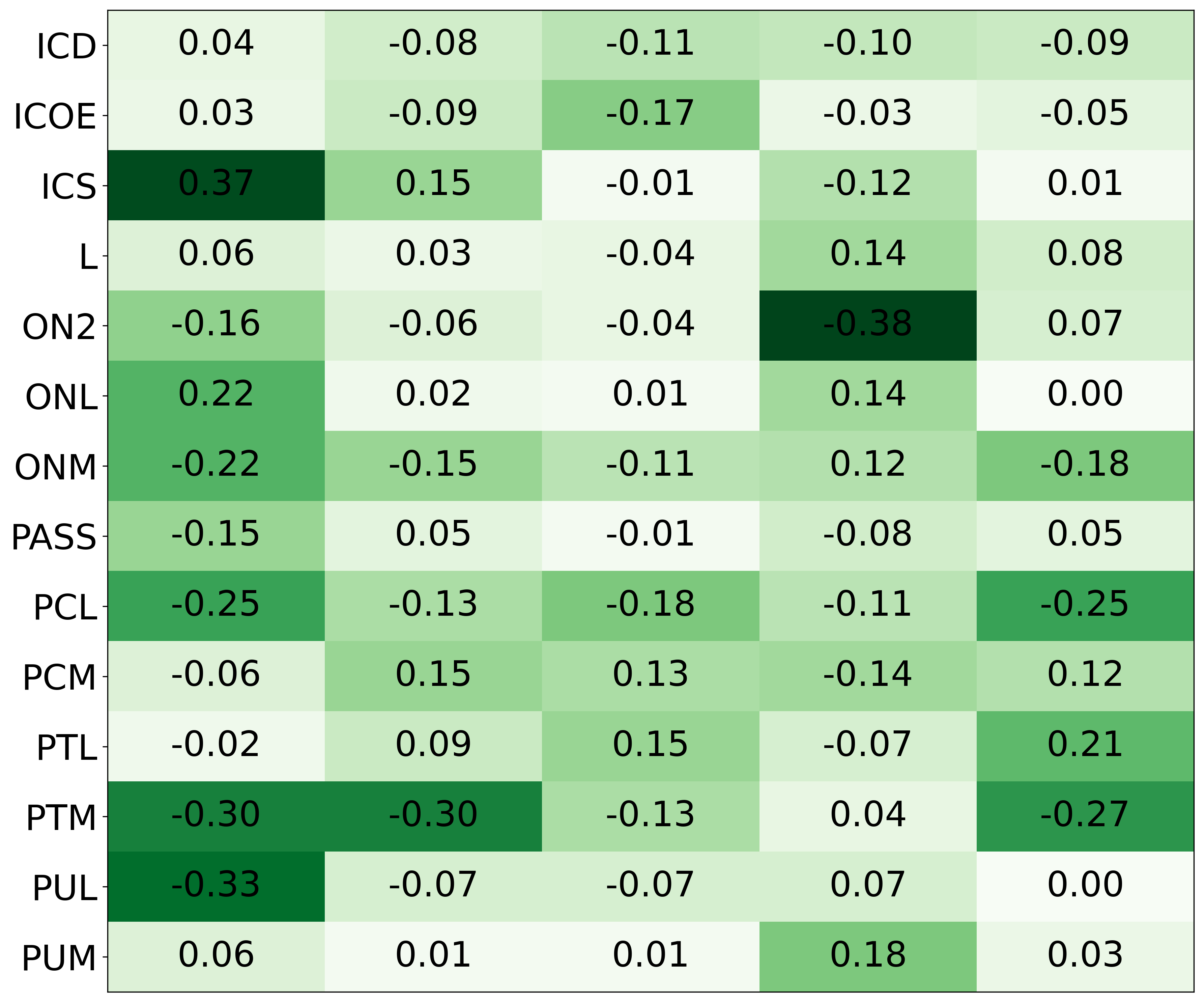}
        \includegraphics[width=1\linewidth]{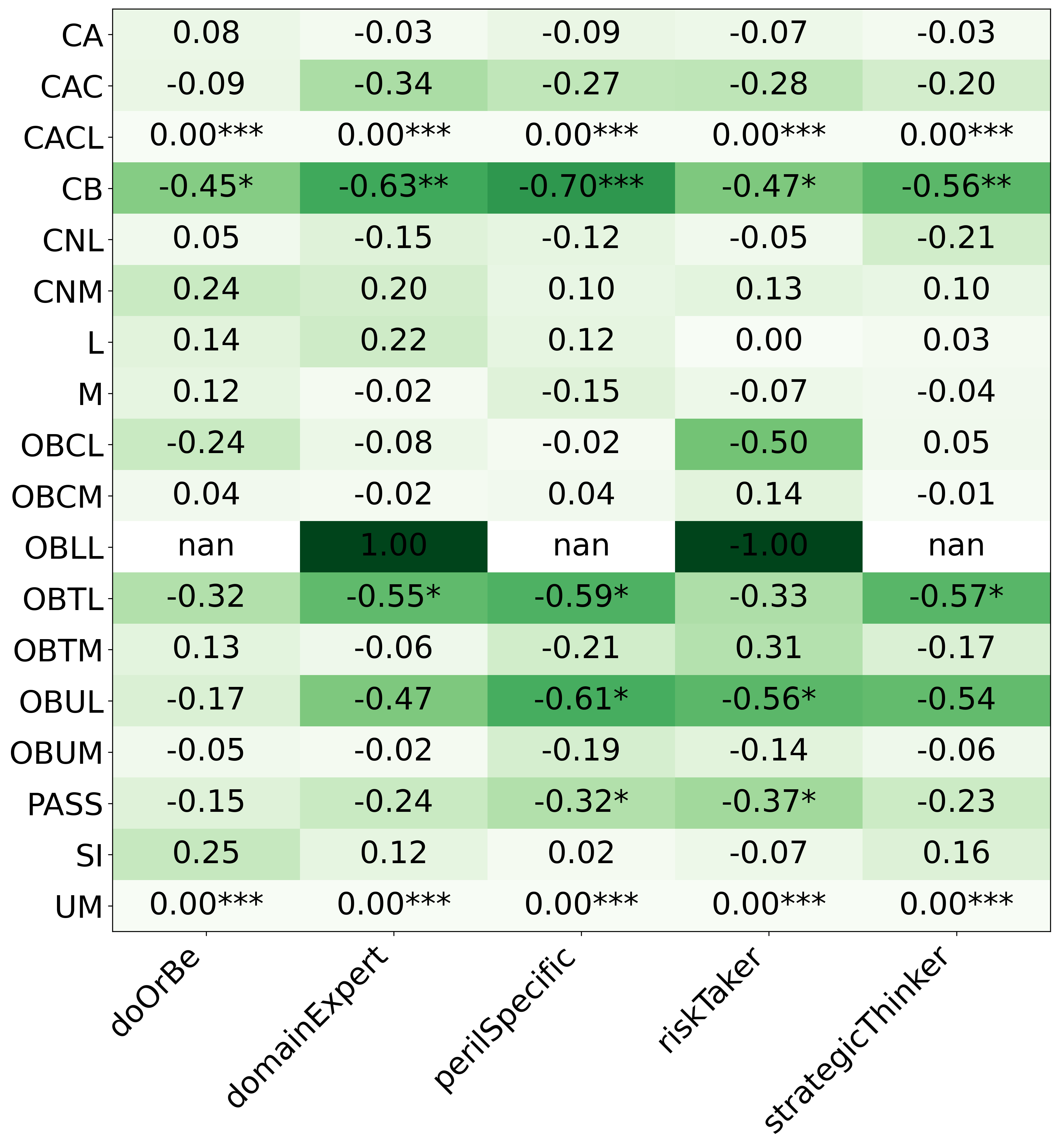}
        \caption{Heuristic Correlations - DH2 - LLaMA 3}
        \label{fig:inventoryHeuristics_l3_dh_2}
    \end{figure}

\begin{figure}[H]
        \centering
        \includegraphics[width=1\linewidth]{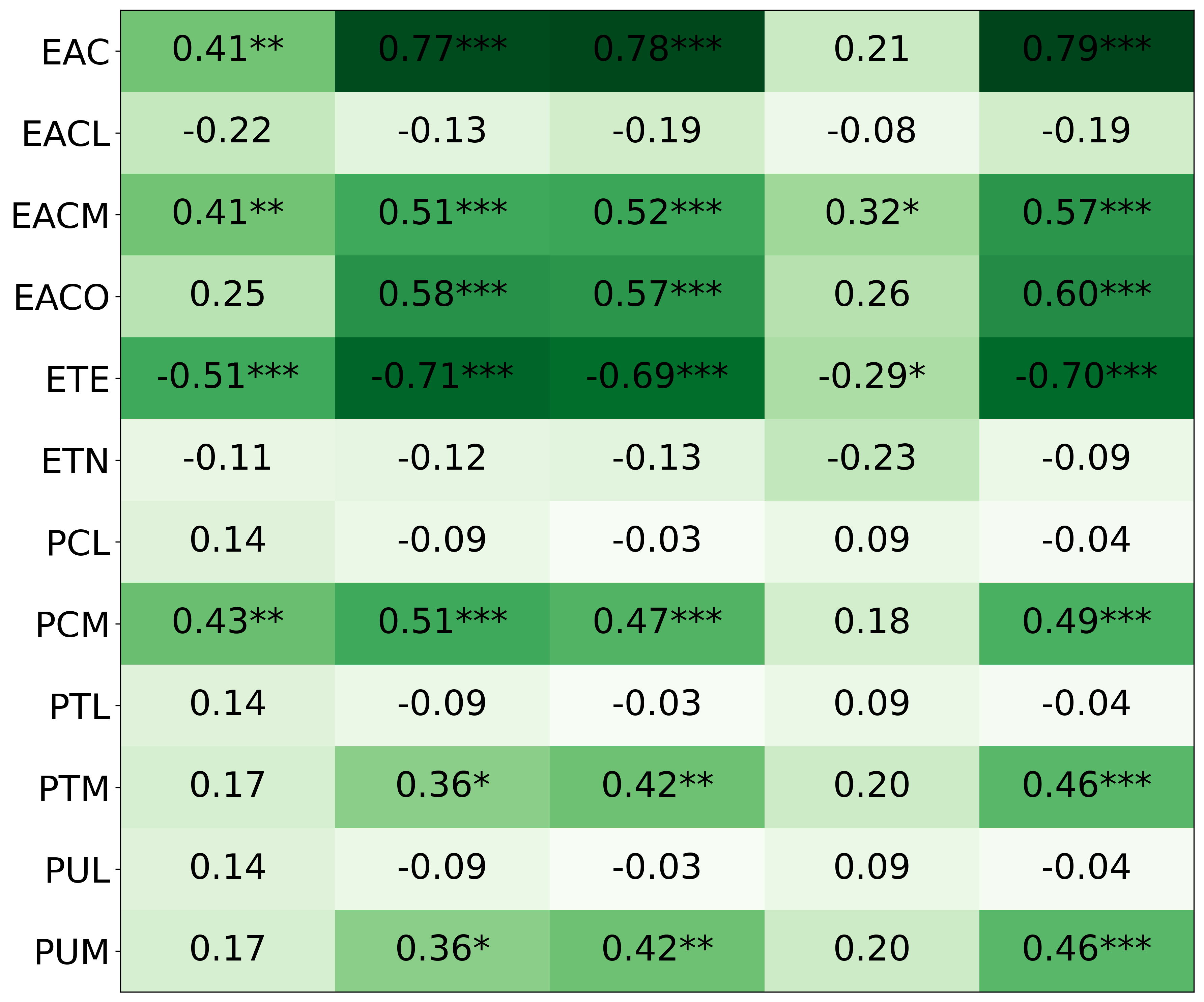}
        \hspace*{1px}
        \includegraphics[width=1\linewidth]{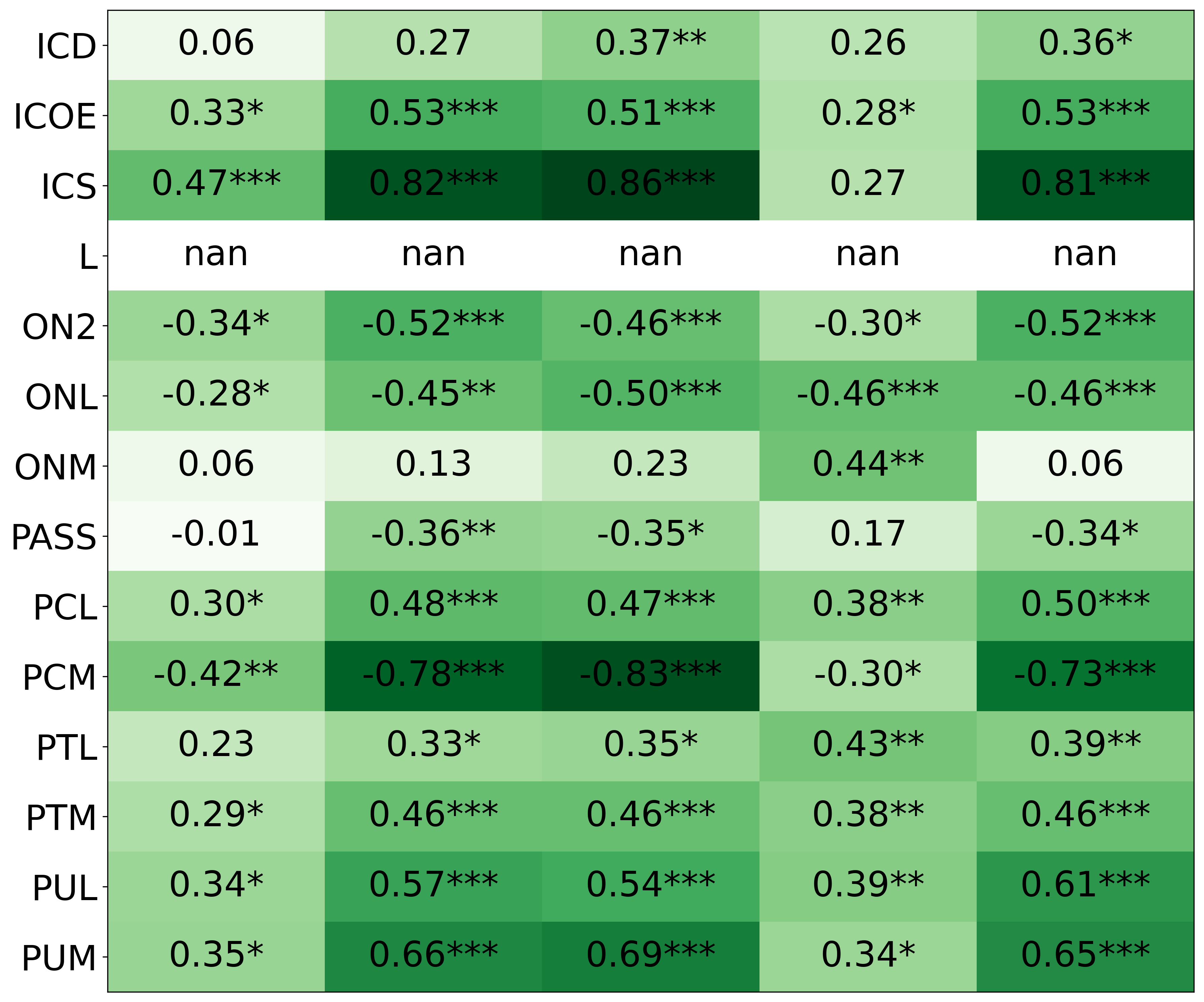}
        \includegraphics[width=1\linewidth]{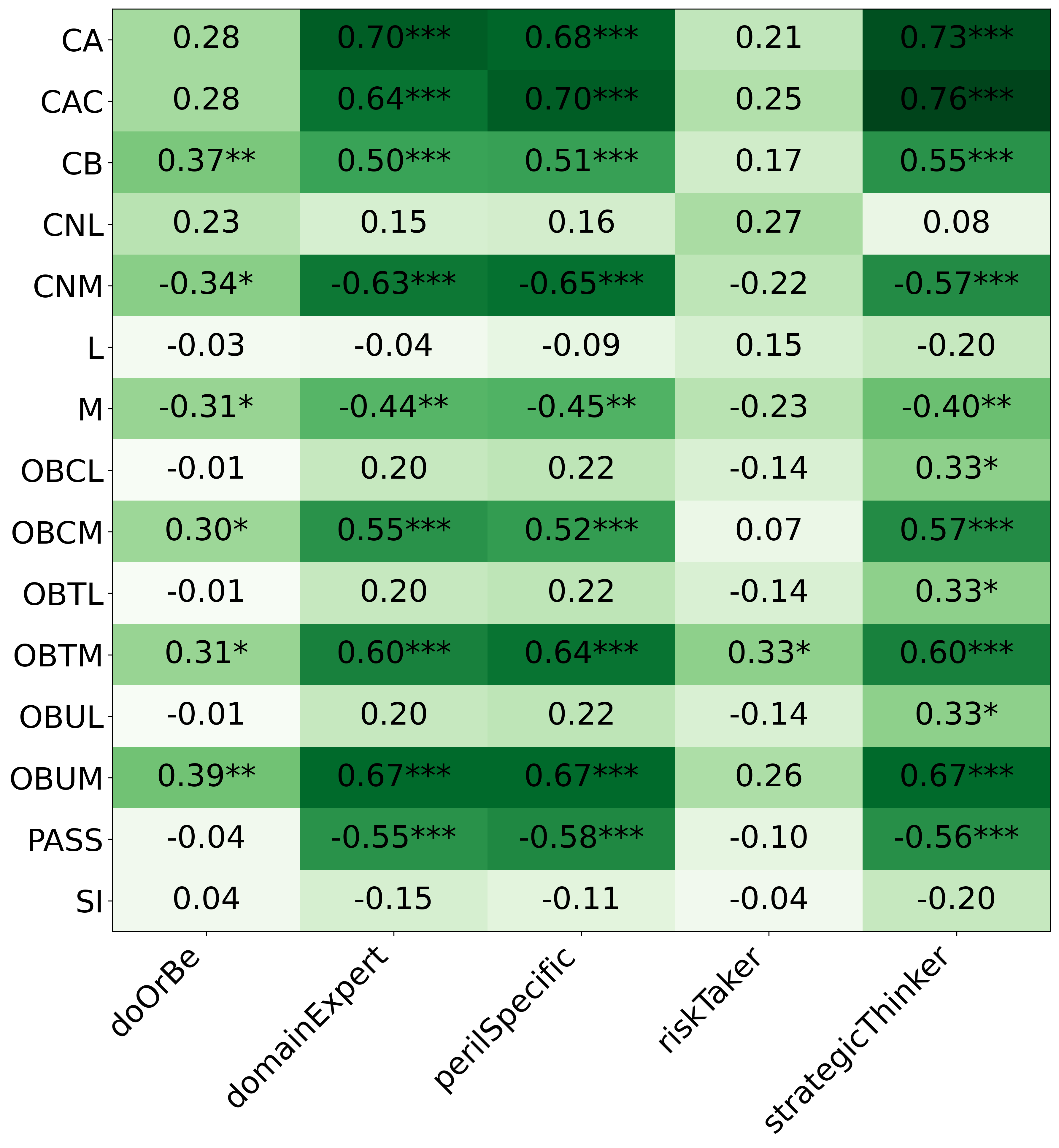}
        \caption{Heuristic Correlations - PI2 - LLaMA 3}
        \label{fig:inventoryHeuristics_l3_pi_2}
    \end{figure}

\begin{figure}[H]
        \centering
        \includegraphics[width=1\linewidth]{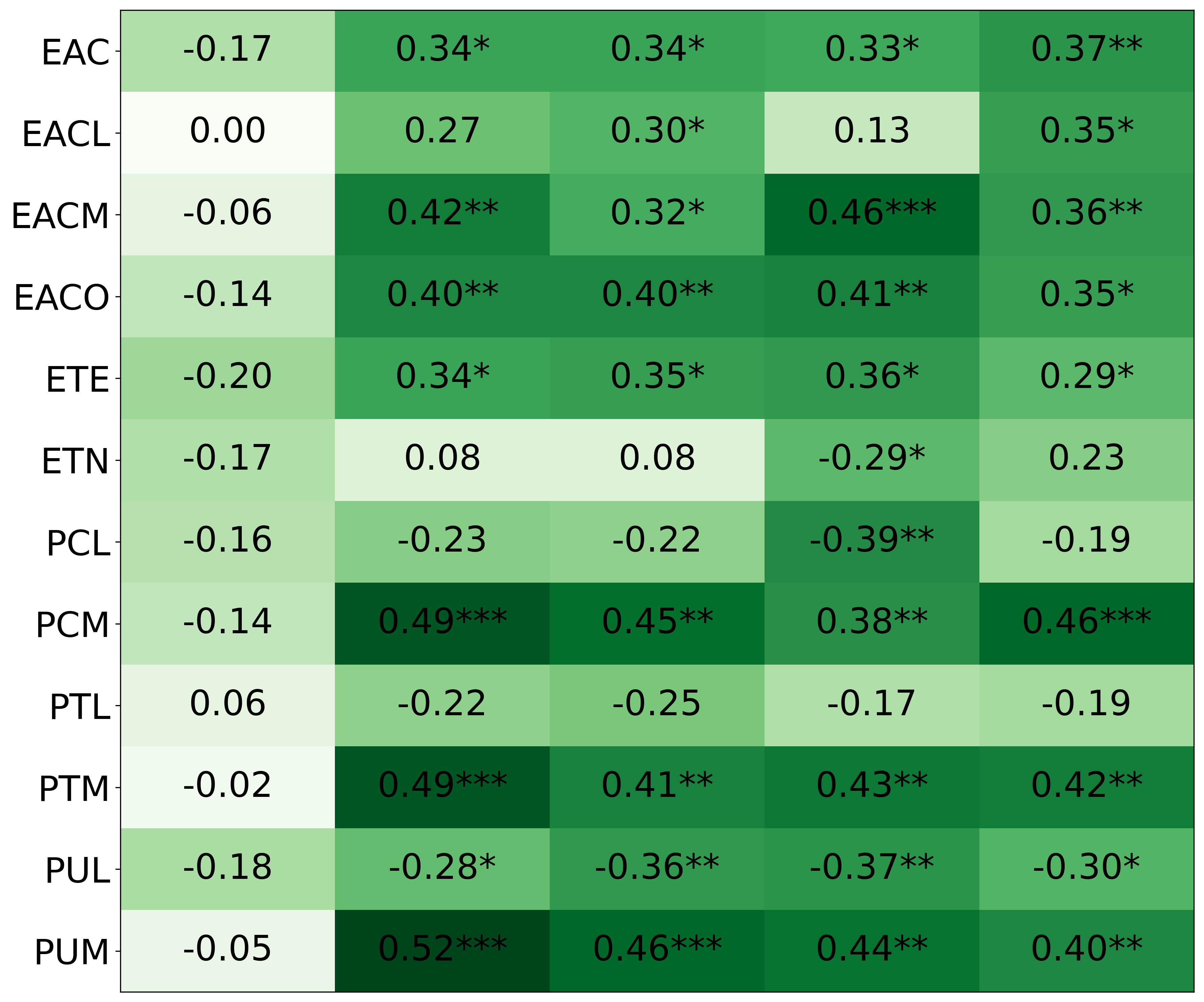}
        \hspace*{1px}
        \includegraphics[width=1\linewidth]{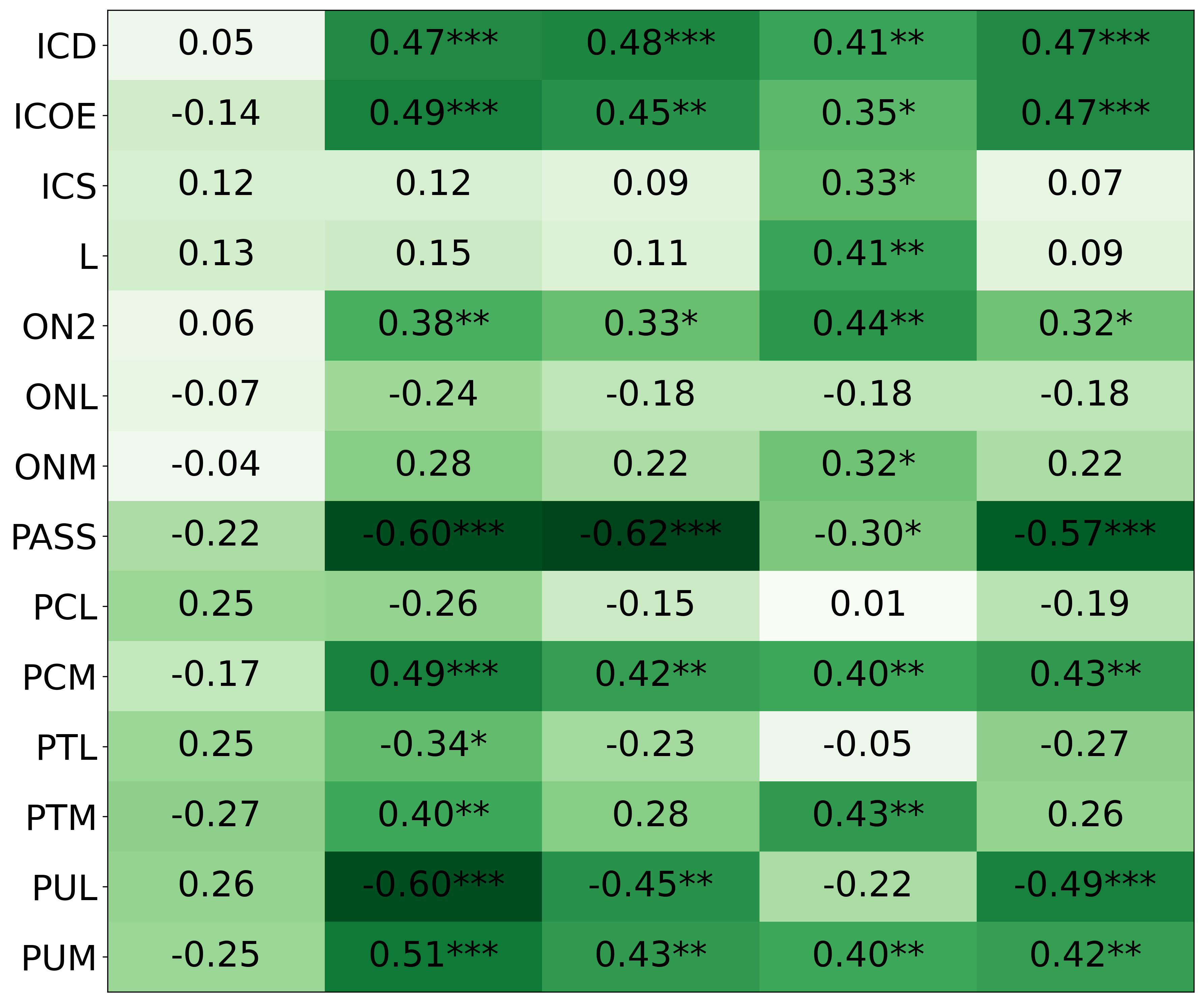}
        \includegraphics[width=1\linewidth]{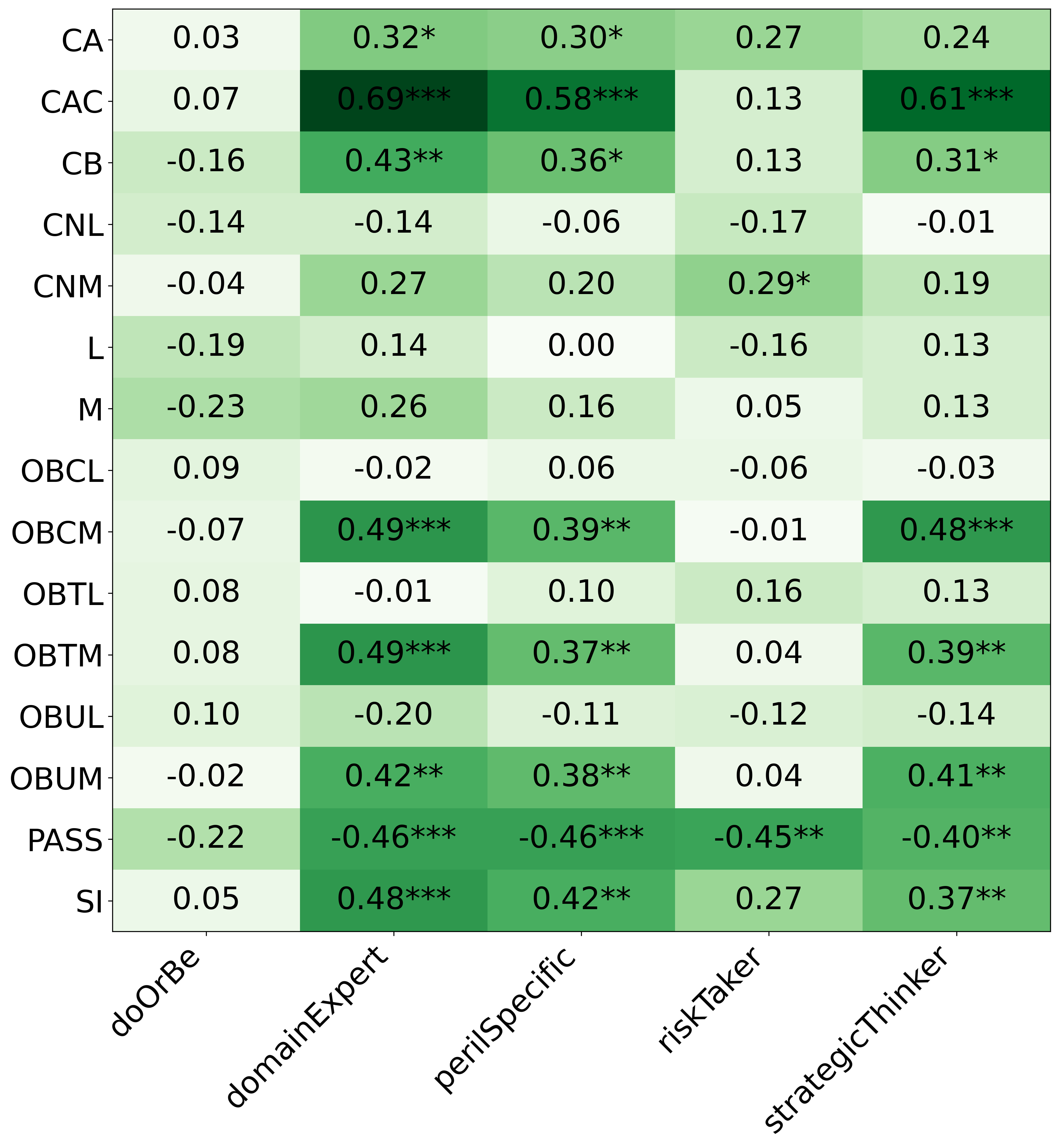}
        \caption{Heuristic Correlations - DH1 - LLaMA 4}
        \label{fig:inventoryHeuristics_l4_dh_1}
    \end{figure}

\begin{figure}[H]
        \centering
        \includegraphics[width=1\linewidth]{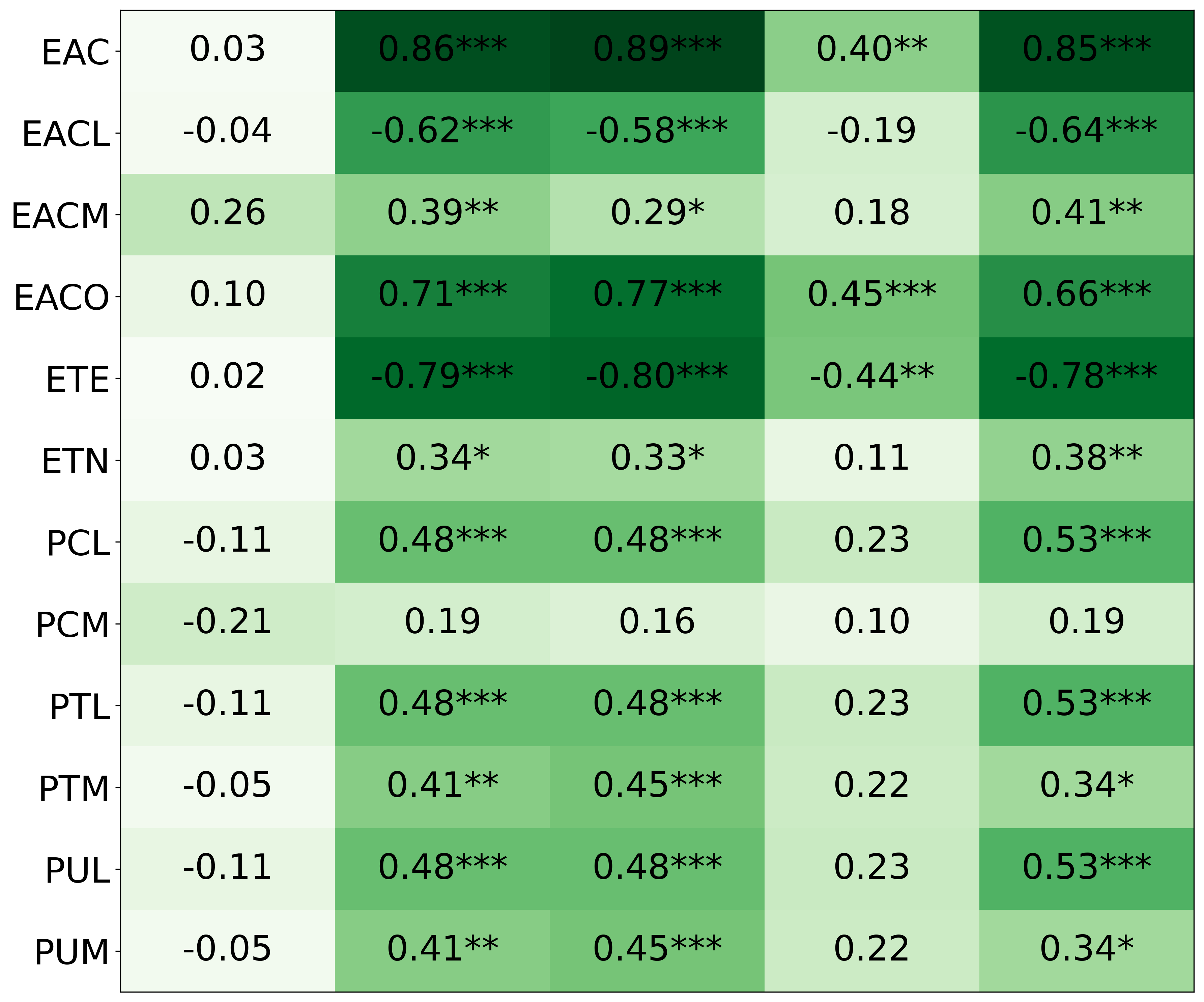}
        \hspace*{1px}
        \includegraphics[width=1\linewidth]{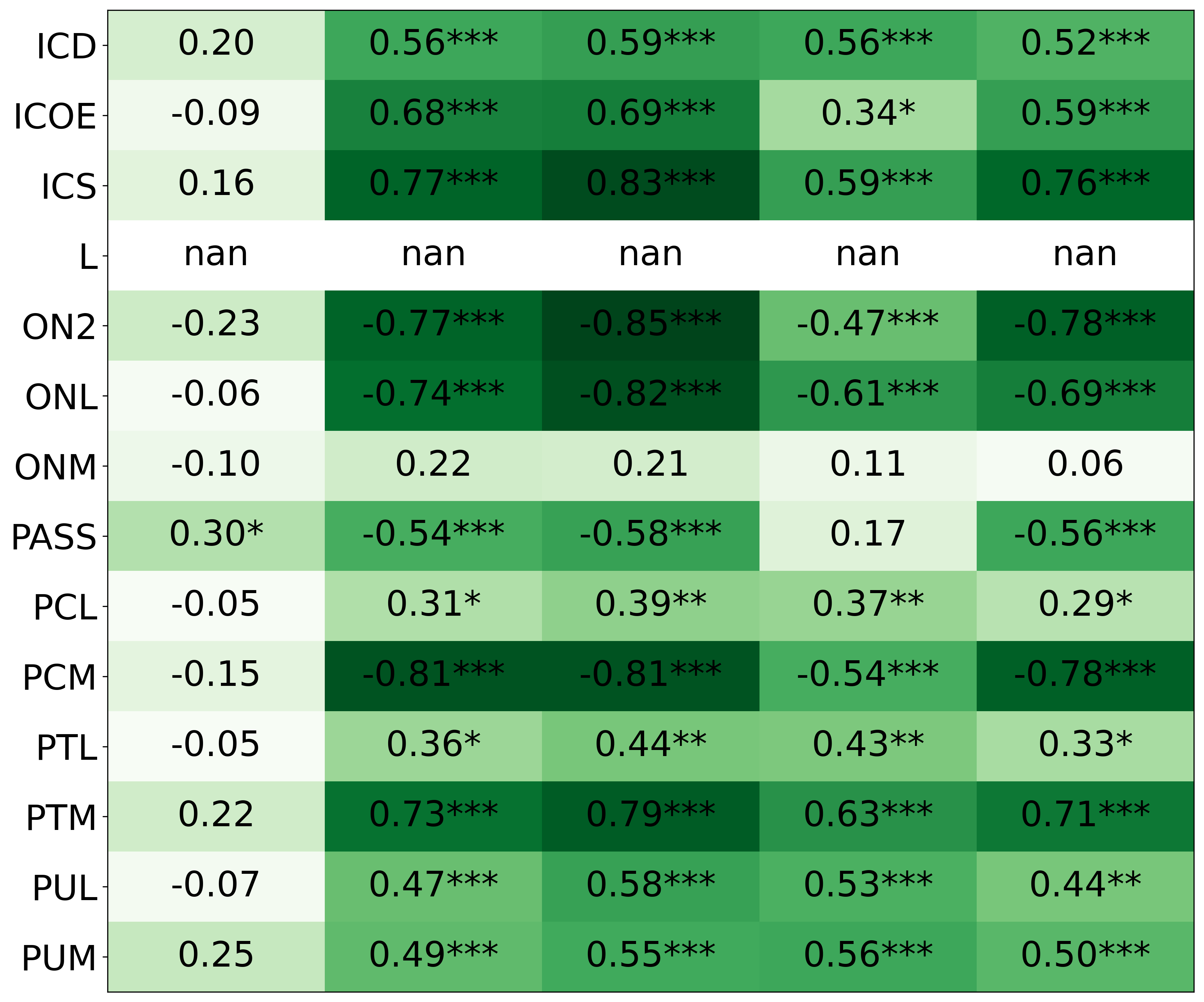}
        \includegraphics[width=1\linewidth]{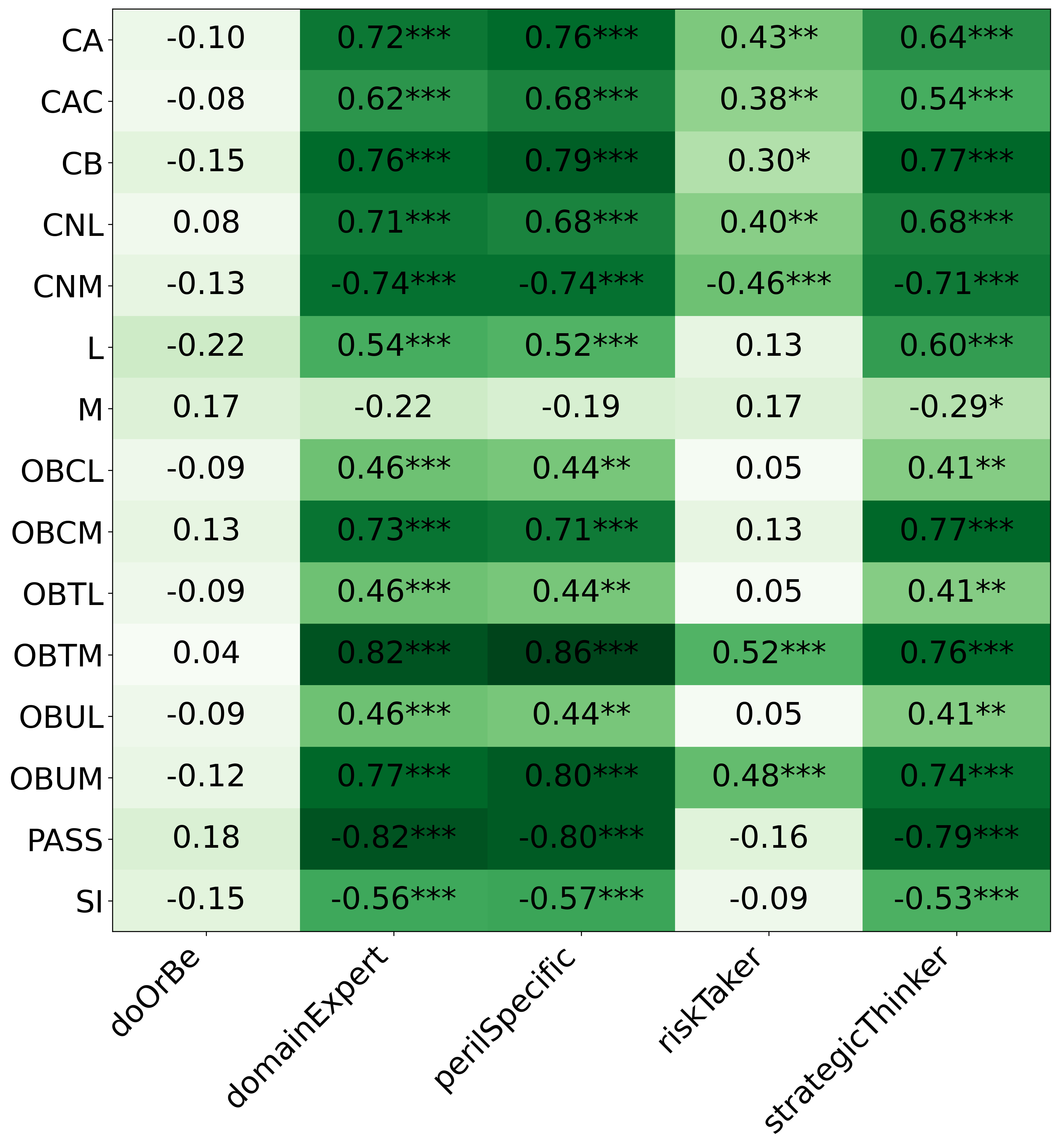}
        \caption{Heuristic Correlations - PI1 - LLaMA 4}
        \label{fig:inventoryHeuristics_l4_pi_1}
    \end{figure}

\begin{figure}[H]
        \centering
        \includegraphics[width=1\linewidth]{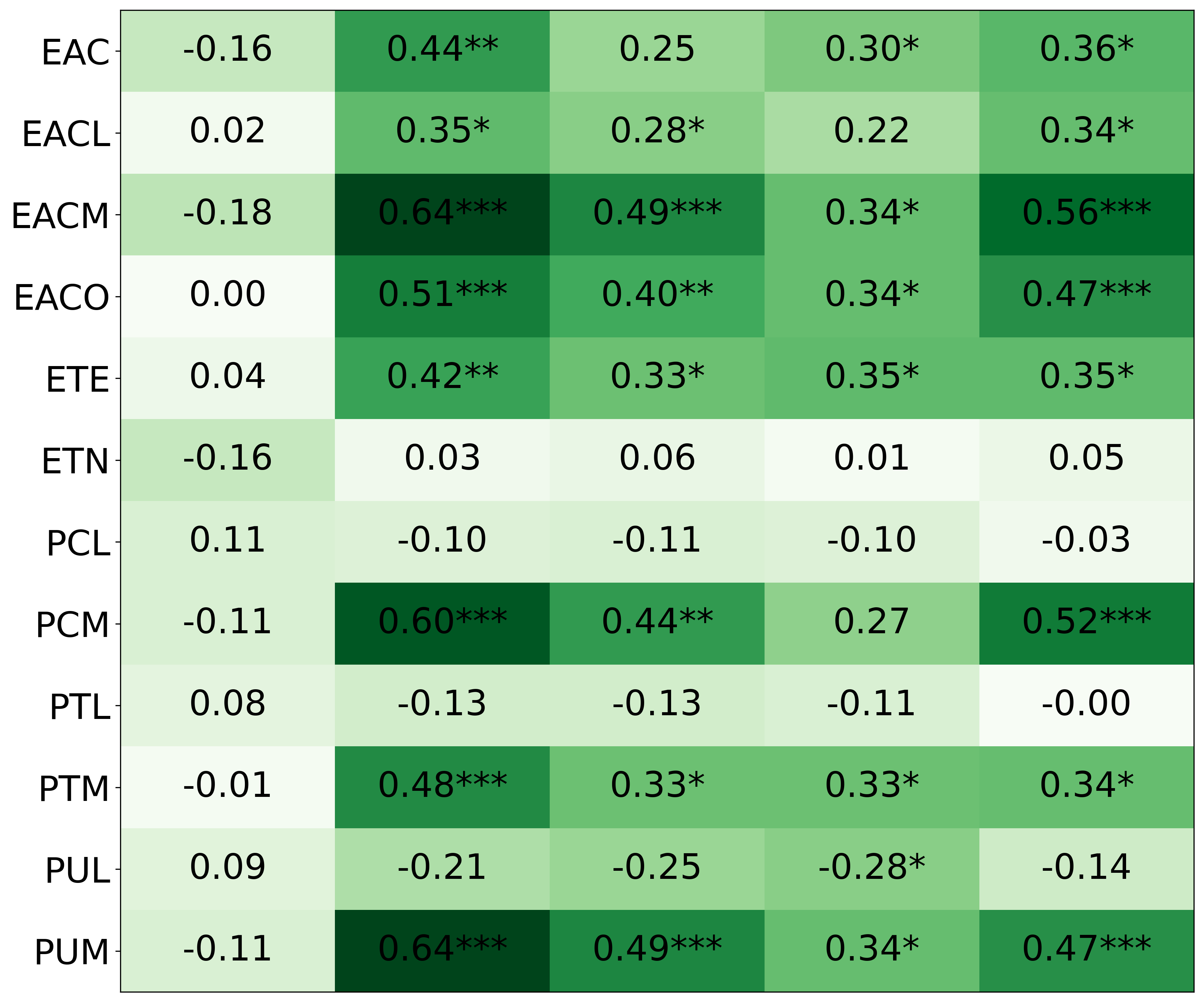}
        \hspace*{1px}
        \includegraphics[width=1\linewidth]{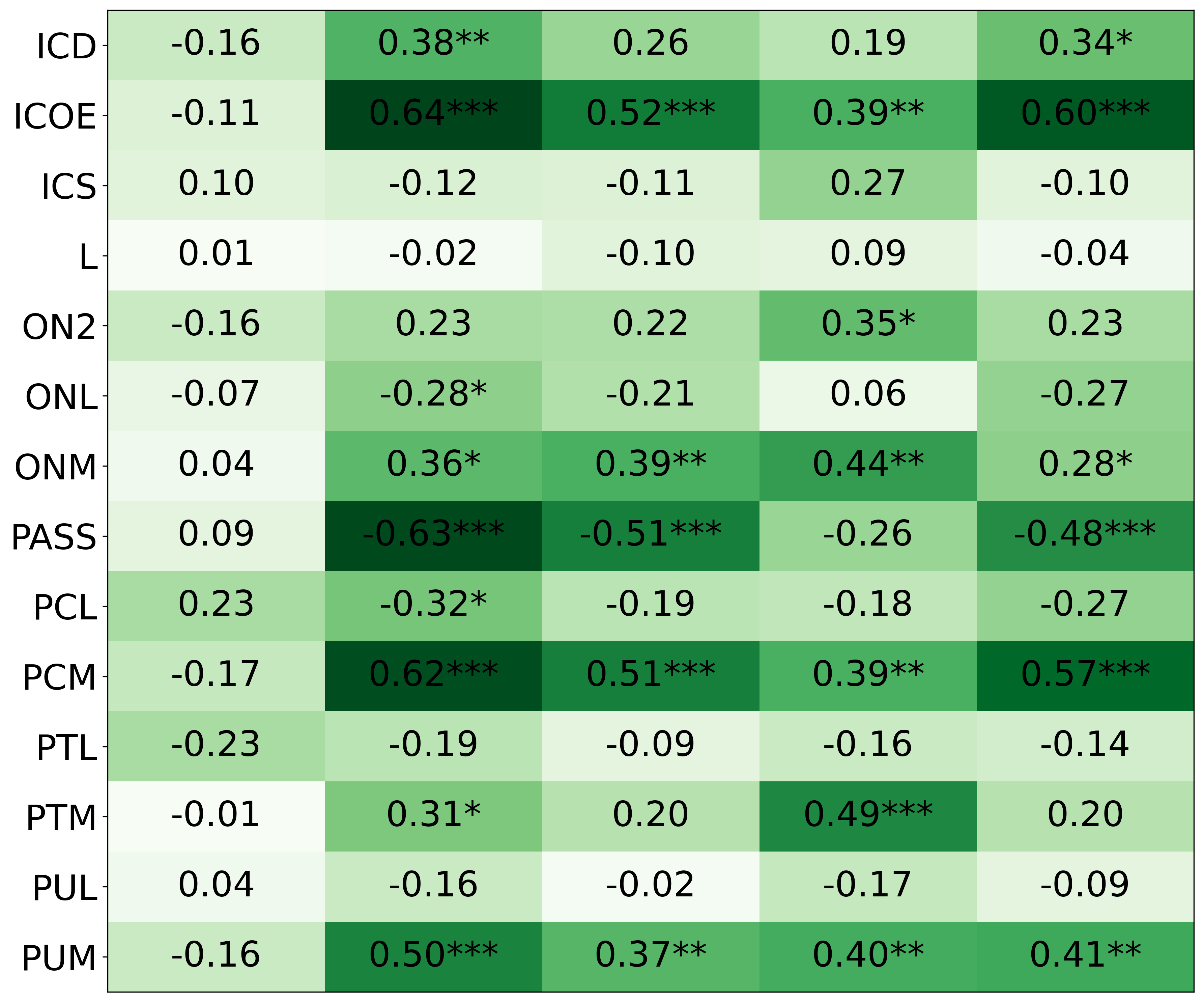}
        \includegraphics[width=1\linewidth]{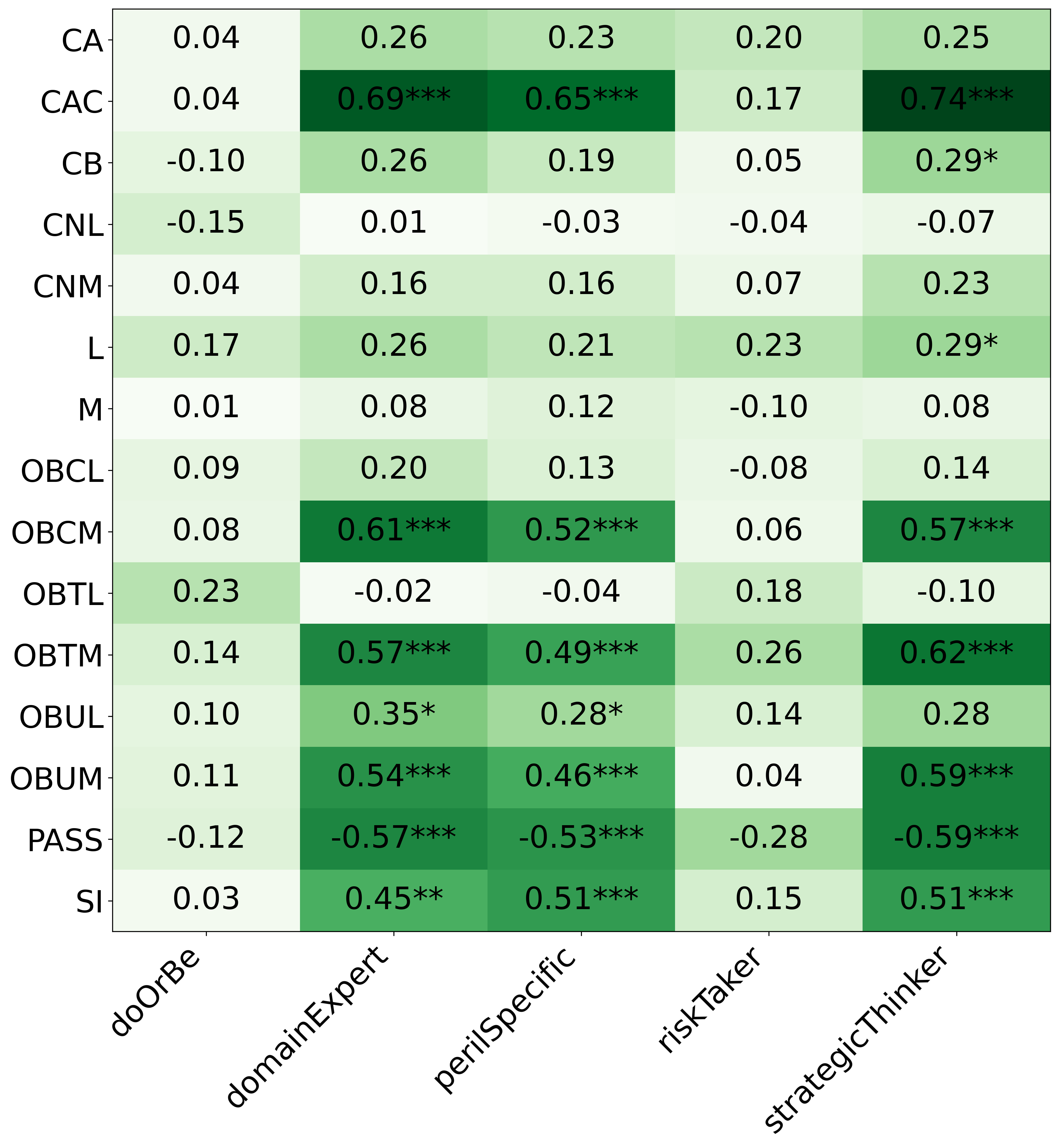}
        \caption{Heuristic Correlations - DH2 - LLaMA 4}
        \label{fig:inventoryHeuristics_l4_dh_2}
    \end{figure}

\begin{figure}[H]
        \centering
        \includegraphics[width=1\linewidth]{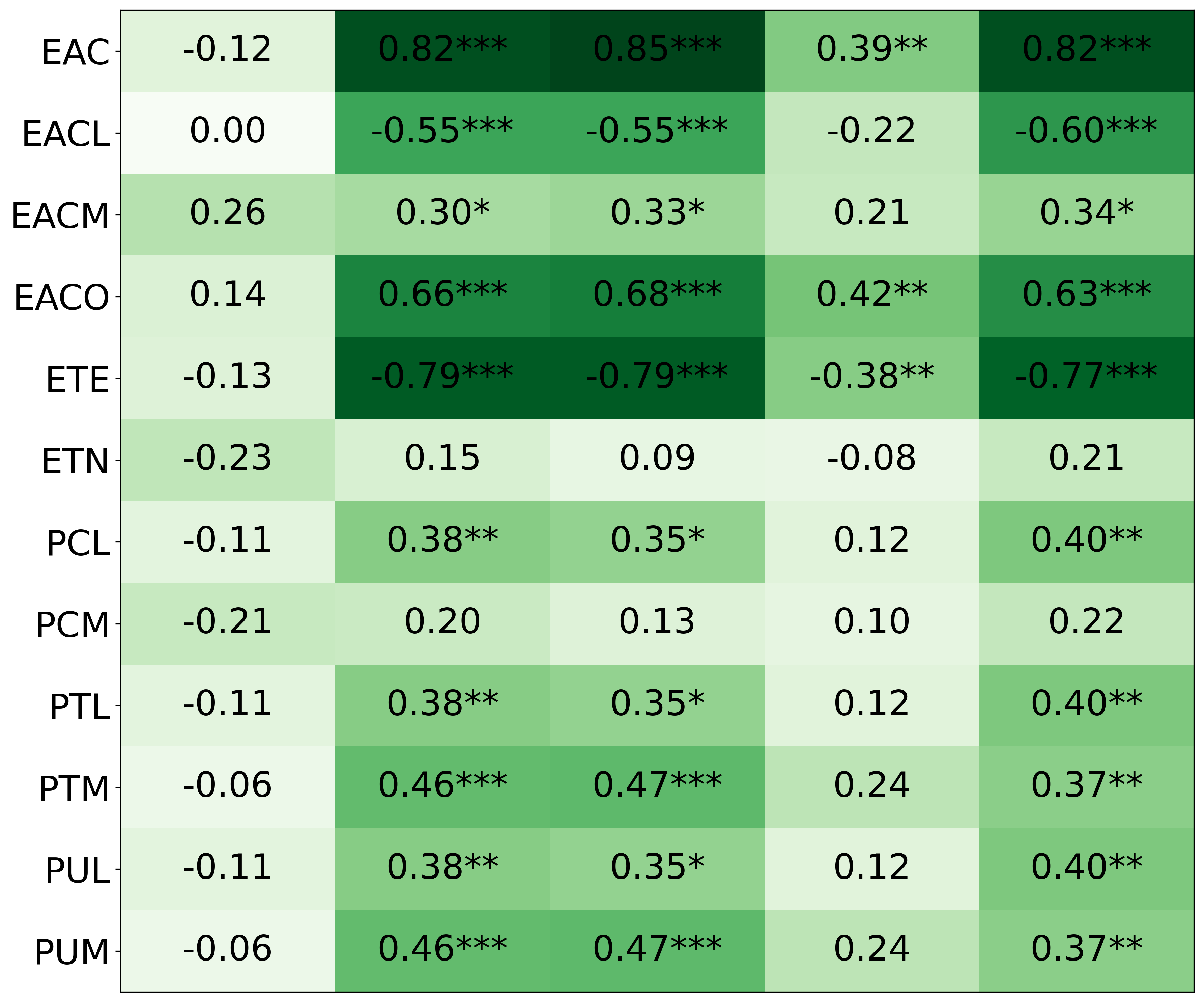}
        \hspace*{1px}
        \includegraphics[width=1\linewidth]{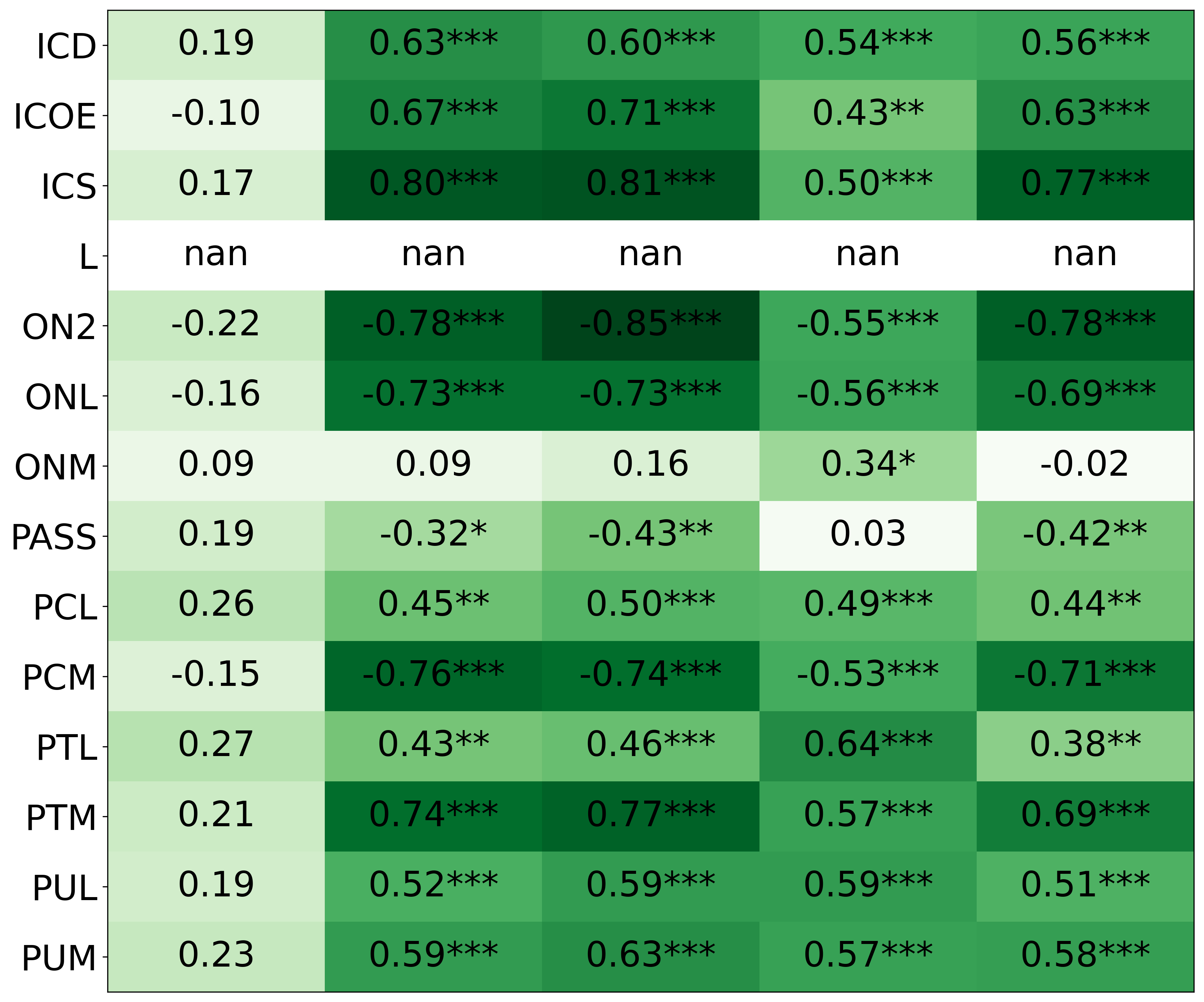}
        \includegraphics[width=1\linewidth]{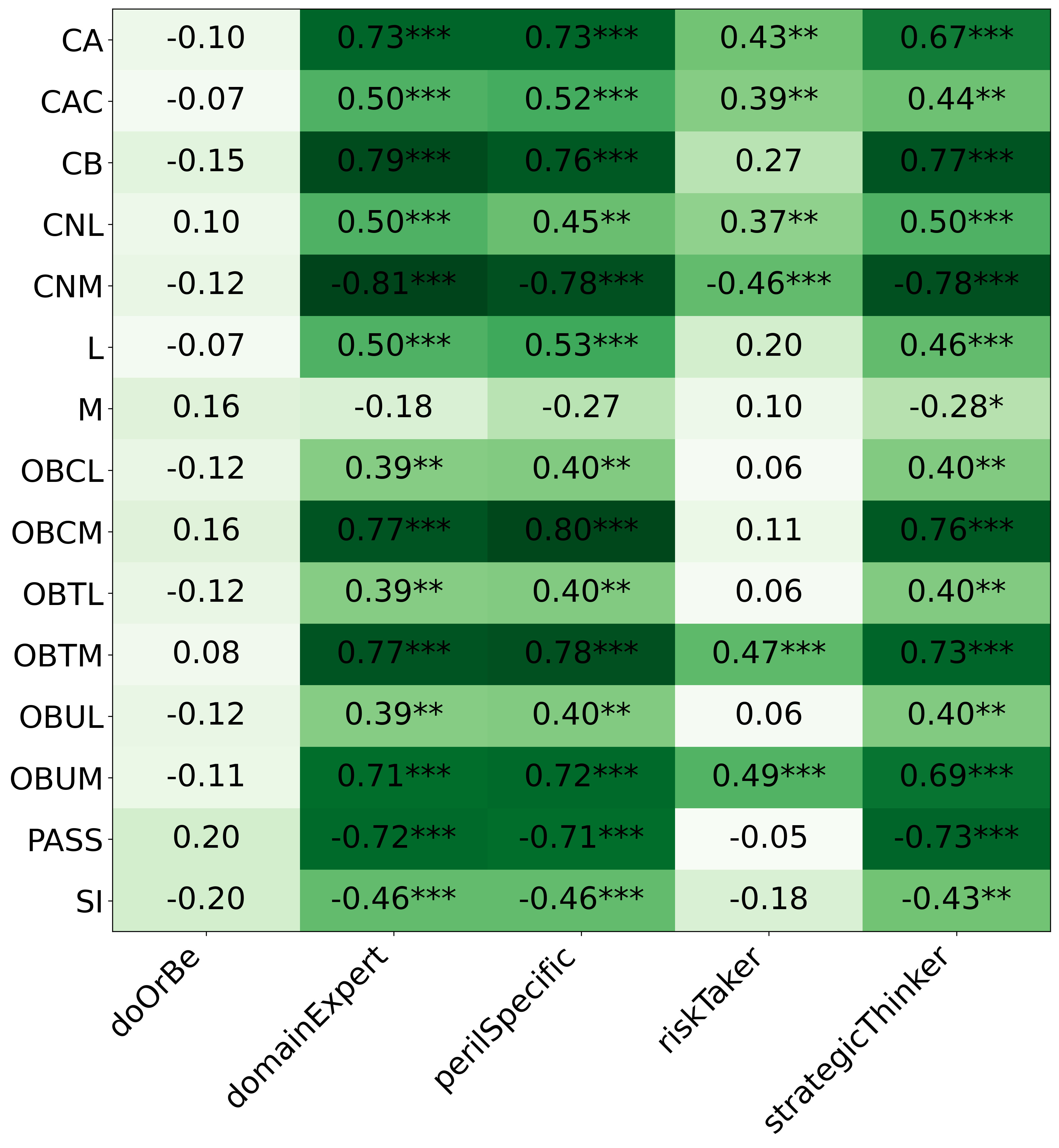}
        \caption{Heuristic Correlations - PI2 - LLaMA 4}
        \label{fig:inventoryHeuristics_l4_pi_2}
    \end{figure}

    This section expands upon the relationship between the annotated assessments in section 3.2 and the generated heuristics (both direct and inventory) as described in section 3.3. The figures show the correlation between each personality feature and heuristic weights, as chosen by players using the direct and inventory heuristic methods. Each figure caption indicates the heuristic (DH/PI) used and its generation batch (1/2).  All batches other than GPT4 were generated twice. The figures are ordered by phase from top to bottom: Phase 1 heuristics (initialization / deployment), Phase 2 (attack), Phase 3 (redeployment). Additionally, the statistical significance of each entry is indicated by asterisks "*" as follows: $*=(p\leq0.05), **=(p\leq0.01), ***=(p\leq0.005)$. Heuristic appearing as `nan' are due to not having been selected by the LLM. Relevant Figures: \ref{fig:inventoryHeuristics_gpt_dh_1} - \ref{fig:inventoryHeuristics_l4_pi_2}.
\raggedbottom

\clearpage
\subsection{Persona Descriptions}
    This section contains tables similar to Table \ref{tab:elo_personas}. These tables show the top and bottom five performing personas based on their final TrueSkill ratings after 1200 tournament games. Relevant Tables: \ref{tab:elo_personas_l3_dh_1_1} - \ref{tab:elo_personas_ms_pi_2_2}.

    \begin{table}[h]
        \footnotesize
        \centering
        \begin{tabularx}{\columnwidth}{|l|X|}
        \hline
        \rowcolor{lightgray} \textbf{Rating} & \textbf{Persona Description} \\ 
        \hline
29.5914 & A nanny who works with the widower to provide additional support and care for the children \\ \hline
29.1766 & A literary critic who dismisses the Madam Tulip series as shallow and predictable \\ \hline
28.7519 & A military historian writing a book on the impact of military leaders on shaping discipline and leadership \\ \hline
28.5374 & A startup CTO who emphasizes rapid deployment and market responsiveness over meticulous code craftsmanship \\ \hline
28.2439 & A recently promoted Major General in the Nigerian Army \\ \hline
        \hline
18.3503 & A U.S. Army veteran who served alongside the retired interpreter and shares a deep bond of mutual trust and respect \\ \hline
20.1616 & A military historian specializing in the Napoleonic Wars, with a focus on the Peninsular Campaign and the strategies employed by both the Allied and French forces \\ \hline
20.6356 & An amateur genealogist tracing family histories with possible connections to noble lineages in Scotland and England \\ \hline
21.2993 & A struggling high school student who has no interest in biology \\ \hline
21.3212 & A supply chain and logistics consultant for aviation and defense industries, aiming to analyze the effectiveness of Axis supply lines and Allied interdiction efforts \\ \hline
        \end{tabularx}
        \caption{TrueSkill for LLaMA 3 - DH1 - Run 1}
        \label{tab:elo_personas_l3_dh_1_1}
    \end{table}

\begin{table}[h]
    \footnotesize
    \centering
    \begin{tabularx}{\columnwidth}{|l|X|}
    \hline
    \rowcolor{lightgray} \textbf{Rating} & \textbf{Persona Description} \\
    \hline
29.1663 & A retired naval officer with experience in aircraft carrier operations and a deep knowledge of US Navy history and traditions. \\ \hline
28.9431 & A military historian writing a book on the impact of military leaders on shaping discipline and leadership. \\ \hline
28.7413 & A retired military general with extensive experience in national defense, providing insights on border security strategies. \\ \hline
27.5964 & A competitive collegiate football player always seeking for custom-designed team merchandise \\ \hline
27.3898 & An ancient deity known for their wisdom in cosmic affairs and mastery of temporal phenomena \\ \hline
    \hline
18.7251 & A U.S. Army veteran who served alongside the retired interpreter and shares a deep bond of mutual trust and respect \\ \hline
19.4858 & A military historian specializing in the Napoleonic Wars, with a focus on the Peninsular Campaign and the strategies employed by both the Allied and French forces. \\ \hline
21.9361 & A photographer skilled at capturing detailed images of the artifacts. \\ \hline
21.9867 & A geopolitical strategist who often appears on different networks presenting an alternative viewpoint on policies and events \\ \hline
22.0864 & A 7-year-old child diagnosed with autism spectrum disorder, seeking support in managing their social behaviors \\ \hline
    \end{tabularx}
    \caption{TrueSkill for LLaMA 3 - DH1 - Run 2}
    \label{tab:elo_personas_l3_dh_1_2}
\end{table}

\begin{table}[h]
    \footnotesize
    \centering
    \begin{tabularx}{\columnwidth}{|l|X|}
    \hline
    \rowcolor{lightgray} \textbf{Rating} & \textbf{Persona Description} \\
    \hline
30.2914 & An elderly relative who relies on the journalist's explanations to stay up-to-date on the latest technology trends. \\ \hline
27.9751 & A teenager struggling with anxiety and looking for coping mechanisms \\ \hline
27.7713 & A government agency using GIS analysis to plan efficient land use and infrastructure development \\ \hline
27.7561 & A military strategist in the Department of Defense, interested in the use of real-time satellite data to support decision-making and improve mission effectiveness. \\ \hline
27.6661 & A struggling high school student who has no interest in biology. \\ \hline
    \hline
19.4921 & A young child who comes to the shop every day with their parents to buy Kinder Eggs \\ \hline
20.6314 & A young child who laughs uncontrollably at the street performer's antics \\ \hline
20.8708 & A curious toddler who eagerly explores and enjoys playing with the DIY toys \\ \hline
21.3603 & A startup CTO who emphasizes rapid deployment and market responsiveness over meticulous code craftsmanship \\ \hline
21.3628 & A grassroots activist advocating for defunding the police and investing in alternative community-based solutions \\ \hline
    \end{tabularx}
    \caption{TrueSkill for LLaMA 3 - PI1 - Run 1}
    \label{tab:elo_personas_l3_pi_1_1}
\end{table}

\begin{table}[h]
    \footnotesize
    \centering
    \begin{tabularx}{\columnwidth}{|l|X|}
    \hline
    \rowcolor{lightgray} \textbf{Rating} & \textbf{Persona Description} \\ 
    \hline
28.9910 & A retired intelligence officer who had previously worked for the CIA. \\ \hline
28.0686 & A 7-year-old child diagnosed with autism spectrum disorder, seeking support in managing their social behaviors \\ \hline
27.8405 & A retired military general with extensive experience in national defense, providing insights on border security strategies. \\ \hline
27.6526 & A struggling high school student who has no interest in biology. \\ \hline
27.5939 & A wargaming enthusiast who designs and simulates historical military scenarios to explore the decision-making processes, strategies, and outcomes of various engagements, including the Capture of Guam. \\ \hline
    \hline
20.3536 & A startup CTO who emphasizes rapid deployment and market responsiveness over meticulous code craftsmanship \\ \hline
21.4205 & A young child who comes to the shop every day with their parents to buy Kinder Eggs \\ \hline
21.9011 & A person who struggles with Discardia – a fear of throwing things away. \\ \hline
22.3848 & A young child who laughs uncontrollably at the street performer's antics \\ \hline
22.5022 & A nanny who works with the widower to provide additional support and care for the children \\ \hline
    \end{tabularx}
    \caption{TrueSkill for LLaMA 3 - PI1 - Run 2}
    \label{tab:elo_personas_l3_pi_1_1}
\end{table}

\begin{table}[h]
    \footnotesize
    \centering
    \begin{tabularx}{\columnwidth}{|l|X|}
    \hline
    \rowcolor{lightgray} \textbf{Rating} & \textbf{Persona Description} \\ 
    \hline
29.5645 & An elderly woman who relies on the apothecary for her herbal remedies and trusts their expertise. \\ \hline
29.0417 & A genealogist studying the origins and variations of Arabic-language surnames, including Al-Marri and its related surnames such as Marri and Al Murrah. \\ \hline
28.7365 & A young child fascinated by Disney movies and loves to hear stories from their collection \\ \hline
28.3787 & A genealogist researching family histories connected to Biddeford, Maine. \\ \hline
28.0665 & A military historian writing a book on the impact of military leaders on shaping discipline and leadership. \\ \hline
    \hline
16.0860 & A recently promoted Major General in the Nigerian Army \\ \hline
17.3984 & A rare bird species critically affected by habitat loss \\ \hline
19.1394 & An amateur genealogist tracing family histories with possible connections to noble lineages in Scotland and England. \\ \hline
20.2137 & A traditional comedian who believes in adhering to mainstream comedy and disapproves of pushing boundaries \\ \hline
20.5467 & A U.S. Army veteran who served alongside the retired interpreter and shares a deep bond of mutual trust and respect \\ \hline
    \end{tabularx}
    \caption{TrueSkill for LLaMA 3 - DH2 - Run 1}
    \label{tab:elo_personas_l3_dh_2_1}
\end{table}

\begin{table}[h]
    \footnotesize
    \centering
    \begin{tabularx}{\columnwidth}{|l|X|}
    \hline
    \rowcolor{lightgray} \textbf{Rating} & \textbf{Persona Description} \\ 
    \hline
30.0065 & A young child who comes to the shop every day with their parents to buy Kinder Eggs \\ \hline
29.1161 & A 7-year-old child diagnosed with autism spectrum disorder, seeking support in managing their social behaviors \\ \hline
28.8581 & A nanny who works with the widower to provide additional support and care for the children \\ \hline
28.8233 & A genealogist researching family histories connected to Biddeford, Maine. \\ \hline
28.6546 & A highly respected admiral known for their strategic thinking and ability to inspire and motivate sailors \\ \hline
    \hline
14.7580 & A rare bird species critically affected by habitat loss \\ \hline
15.9706 & A traditional comedian who believes in adhering to mainstream comedy and disapproves of pushing boundaries \\ \hline
16.1548 & A recently promoted Major General in the Nigerian Army \\ \hline
18.7521 & An amateur genealogist tracing family histories with possible connections to noble lineages in Scotland and England. \\ \hline
20.2914 & A seasoned military strategist sharing stories from their covert operations and offering guidance in a fast-paced world of intelligence \\ \hline
    \end{tabularx}
    \caption{TrueSkill for LLaMA 3 - DH2 - Run 2}
    \label{tab:elo_personas_l3_dh_2_2}
\end{table}

\begin{table}[h]
    \footnotesize
    \centering
    \begin{tabularx}{\columnwidth}{|l|X|}
    \hline
    \rowcolor{lightgray} \textbf{Rating} & \textbf{Persona Description} \\ 
    \hline
30.4708 & An Indian military personnel who served his country for 30 years and received the 50th Independence Anniversary Medal. \\ \hline
28.5833 & A confused person who is not familiar with iGEM and Synthetic Biology \\ \hline
27.7430 & A retired Navy veteran who now works as a military consultant and shares practical knowledge \\ \hline
27.4178 & An elderly woman who relies on the apothecary for her herbal remedies and trusts their expertise. \\ \hline
27.3379 & A retired intelligence officer who had previously worked for the CIA. \\ \hline
    \hline
19.7349 & A young child fascinated by Disney movies and loves to hear stories from their collection \\ \hline
20.3243 & A supply chain and logistics consultant for aviation and defense industries, aiming to analyze the effectiveness of Axis supply lines and Allied interdiction efforts. \\ \hline
21.2134 & A young child who laughs uncontrollably at the street performer's antics \\ \hline
21.3314 & A curious toddler who eagerly explores and enjoys playing with the DIY toys \\ \hline
21.5132 & A startup CTO who emphasizes rapid deployment and market responsiveness over meticulous code craftsmanship \\ \hline
    \end{tabularx}
    \caption{TrueSkill for LLaMA 3 - PI2 - Run 1}
    \label{tab:elo_personas_l3_pi_2_1}
\end{table}

\begin{table}[h]
    \footnotesize
    \centering
    \begin{tabularx}{\columnwidth}{|l|X|}
    \hline
    \rowcolor{lightgray} \textbf{Rating} & \textbf{Persona Description} \\ 
    \hline
29.4089 & A military tactician specializing in special operations, particularly in counter-terrorism and hostage situations, with a focus on strategic planning and equipment selection for elite units. \\ \hline
29.3058 & An elderly woman who relies on the apothecary for her herbal remedies and trusts their expertise. \\ \hline
28.9895 & A military historian writing a book on the impact of military leaders on shaping discipline and leadership. \\ \hline
28.0034 & A traditional comedian who believes in adhering to mainstream comedy and disapproves of pushing boundaries \\ \hline
27.9695 & A retired military general with extensive experience in national defense, providing insights on border security strategies. \\ \hline
    \hline
18.5998 & A young child who laughs uncontrollably at the street performer's antics \\ \hline
19.2635 & A grassroots activist advocating for defunding the police and investing in alternative community-based solutions \\ \hline
19.3120 & A startup CTO who emphasizes rapid deployment and market responsiveness over meticulous code craftsmanship \\ \hline
21.0551 & A person who struggles with Discardia – a fear of throwing things away. \\ \hline
21.3481 & A young child fascinated by Disney movies and loves to hear stories from their collection \\ \hline
    \end{tabularx}
    \caption{TrueSkill for LLaMA 3 - PI2 - Run 2}
    \label{tab:elo_personas_l3_pi_2_2}
\end{table}

\begin{table}[h]
    \footnotesize
    \centering
    \begin{tabularx}{\columnwidth}{|l|X|}
    \hline
    \rowcolor{lightgray} \textbf{Rating} & \textbf{Persona Description} \\ 
    \hline
29.4089 & A military tactician specializing in special operations, particularly in counter-terrorism and hostage situations, with a focus on strategic planning and equipment selection for elite units. \\ \hline
29.3058 & An elderly woman who relies on the apothecary for her herbal remedies and trusts their expertise. \\ \hline
28.9895 & A military historian writing a book on the impact of military leaders on shaping discipline and leadership. \\ \hline
28.0034 & A traditional comedian who believes in adhering to mainstream comedy and disapproves of pushing boundaries \\ \hline
27.9695 & A retired military general with extensive experience in national defense, providing insights on border security strategies. \\ \hline
    \hline
18.5998 & A young child who laughs uncontrollably at the street performer's antics \\ \hline
19.2635 & A grassroots activist advocating for defunding the police and investing in alternative community-based solutions \\ \hline
19.3120 & A startup CTO who emphasizes rapid deployment and market responsiveness over meticulous code craftsmanship \\ \hline
21.0551 & A person who struggles with Discardia – a fear of throwing things away. \\ \hline
21.3481 & A young child fascinated by Disney movies and loves to hear stories from their collection \\ \hline
    \end{tabularx}
    \caption{TrueSkill for LLaMA 4 - DH1 - Run 1}
    \label{tab:elo_personas_l4_dh_1_1}
\end{table}

\begin{table}[h]
    \footnotesize
    \centering
    \begin{tabularx}{\columnwidth}{|l|X|}
    \hline
    \rowcolor{lightgray} \textbf{Rating} & \textbf{Persona Description} \\ 
    \hline
29.0400 & A retired intelligence officer who had previously worked for the CIA. \\ \hline
28.5427 & A young child who comes to the shop every day with their parents to buy Kinder Eggs \\ \hline
28.4108 & A genealogist helping clients trace their family roots, particularly those with connections to the Somme department in France. \\ \hline
27.7573 & A military historian writing a book on the impact of military leaders on shaping discipline and leadership. \\ \hline
27.6677 & A teenager struggling with anxiety and looking for coping mechanisms \\ \hline
    \hline
16.8731 & A curious toddler who eagerly explores and enjoys playing with the DIY toys \\ \hline
20.3059 & A nanny who works with the widower to provide additional support and care for the children \\ \hline
20.6517 & A struggling high school student who has no interest in biology. \\ \hline
21.2158 & A photographer skilled at capturing detailed images of the artifacts. \\ \hline
21.2810 & A wargaming enthusiast who enjoys designing and playing strategic simulations of historical military conflicts, particularly those involving Chinese forces. \\ \hline
    \end{tabularx}
    \caption{TrueSkill for LLaMA 4 - DH1 - Run 2}
    \label{tab:elo_personas_l4_dh_1_2}
\end{table}

\begin{table}[h]
    \footnotesize
    \centering
    \begin{tabularx}{\columnwidth}{|l|X|}
    \hline
    \rowcolor{lightgray} \textbf{Rating} & \textbf{Persona Description} \\ 
    \hline
28.4912 & A competitive collegiate football player always seeking for custom-designed team merchandise \\ \hline
28.3990 & A retired naval officer with experience in aircraft carrier operations and a deep knowledge of US Navy history and traditions. \\ \hline
27.7828 & An Indian military personnel who served his country for 30 years and received the 50th Independence Anniversary Medal. \\ \hline
27.6983 & A geopolitical strategist who often appears on different networks presenting an alternative viewpoint on policies and events \\ \hline
27.5775 & A retired intelligence officer who had previously worked for the CIA. \\ \hline
    \hline
20.3577 & A graphic designer seeking advice on how to prevent repetitive strain injuries \\ \hline
21.1707 & A curious toddler who eagerly explores and enjoys playing with the DIY toys \\ \hline
21.6260 & A 7-year-old child diagnosed with autism spectrum disorder, seeking support in managing their social behaviors \\ \hline
21.6414 & A young child who laughs uncontrollably at the street performer's antics \\ \hline
21.7782 & A photographer skilled at capturing detailed images of the artifacts. \\ \hline
    \end{tabularx}
    \caption{TrueSkill for LLaMA 4 - PI1 - Run 1}
    \label{tab:elo_personas_l4_pi_1_1}
\end{table}

\begin{table}[h]
    \footnotesize
    \centering
    \begin{tabularx}{\columnwidth}{|l|X|}
    \hline
    \rowcolor{lightgray} \textbf{Rating} & \textbf{Persona Description} \\ 
    \hline
29.4422 & A retired military general with extensive experience in national defense, providing insights on border security strategies. \\ \hline
29.1101 & A retired Navy veteran who now works as a military consultant and shares practical knowledge \\ \hline
28.5151 & A struggling high school student who has no interest in biology. \\ \hline
27.9740 & A genealogist helping clients trace their family roots, particularly those with connections to the Somme department in France. \\ \hline
27.8130 & A geopolitical strategist who often appears on different networks presenting an alternative viewpoint on policies and events \\ \hline
    \hline
19.0300 & A young child who laughs uncontrollably at the street performer's antics \\ \hline
19.6486 & A startup CTO who emphasizes rapid deployment and market responsiveness over meticulous code craftsmanship \\ \hline
19.9231 & A graphic designer seeking advice on how to prevent repetitive strain injuries \\ \hline
20.9841 & A 7-year-old child diagnosed with autism spectrum disorder, seeking support in managing their social behaviors \\ \hline
21.1041 & A curious toddler who eagerly explores and enjoys playing with the DIY toys \\ \hline
    \end{tabularx}
    \caption{TrueSkill for LLaMA 4 - PI1 - Run 2}
    \label{tab:elo_personas_l4_pi_1_2}
\end{table}

\begin{table}[h]
    \footnotesize
    \centering
    \begin{tabularx}{\columnwidth}{|l|X|}
    \hline
    \rowcolor{lightgray} \textbf{Rating} & \textbf{Persona Description} \\ 
    \hline
30.0335 & A grassroots activist advocating for defunding the police and investing in alternative community-based solutions \\ \hline
28.3202 & A healthcare blogger who spreads misinformation about vaccines and challenges the nurse's beliefs \\ \hline
28.1103 & A retired naval officer with experience in aircraft carrier operations and a deep knowledge of US Navy history and traditions. \\ \hline
27.9493 & A wargaming enthusiast who enjoys designing and playing strategic simulations of historical military conflicts, particularly those involving Chinese forces. \\ \hline
27.7971 & An elderly woman who relies on the apothecary for her herbal remedies and trusts their expertise. \\ \hline
    \hline
16.9432 & A teenager struggling with anxiety and looking for coping mechanisms \\ \hline
19.2549 & A traditional comedian who believes in adhering to mainstream comedy and disapproves of pushing boundaries \\ \hline
20.7565 & A confused person who is not familiar with iGEM and Synthetic Biology \\ \hline
21.7609 & A struggling high school student who has no interest in biology. \\ \hline
22.7316 & A retired Navy veteran who now works as a military consultant and shares practical knowledge \\ \hline
    \end{tabularx}
    \caption{TrueSkill for LLaMA 4 - DH2 - Run 1}
    \label{tab:elo_personas_l4_dh_2_1}
\end{table}

\begin{table}[h]
    \footnotesize
    \centering
    \begin{tabularx}{\columnwidth}{|l|X|}
    \hline
    \rowcolor{lightgray} \textbf{Rating} & \textbf{Persona Description} \\ 
    \hline
28.1846 & A supply chain and logistics consultant for aviation and defense industries, aiming to analyze the effectiveness of Axis supply lines and Allied interdiction efforts. \\ \hline
27.8466 & A geopolitical strategist who often appears on different networks presenting an alternative viewpoint on policies and events \\ \hline
27.8174 & A startup CTO who emphasizes rapid deployment and market responsiveness over meticulous code craftsmanship \\ \hline
27.5886 & A grassroots activist advocating for defunding the police and investing in alternative community-based solutions \\ \hline
27.5369 & An elderly woman who relies on the apothecary for her herbal remedies and trusts their expertise. \\ \hline
    \hline
21.3166 & A confused person who is not familiar with iGEM and Synthetic Biology \\ \hline
21.5359 & A struggling high school student who has no interest in biology. \\ \hline
21.7931 & A disaster relief coordinator working on preparedness and emergency response plans for towns in Tornado Alley \\ \hline
21.9628 & A traditional comedian who believes in adhering to mainstream comedy and disapproves of pushing boundaries \\ \hline
22.9009 & A genealogist researching family histories connected to Biddeford, Maine. \\ \hline
    \end{tabularx}
    \caption{TrueSkill for LLaMA 4 - DH2 - Run 2}
    \label{tab:elo_personas_l4_dh_2_2}
\end{table}

\begin{table}[h]
    \footnotesize
    \centering
    \begin{tabularx}{\columnwidth}{|l|X|}
    \hline
    \rowcolor{lightgray} \textbf{Rating} & \textbf{Persona Description} \\ 
    \hline
28.4403 & A nanny who works with the widower to provide additional support and care for the children \\ \hline
28.0372 & A geopolitical strategist who often appears on different networks presenting an alternative viewpoint on policies and events \\ \hline
27.4786 & A highly respected admiral known for their strategic thinking and ability to inspire and motivate sailors \\ \hline
27.3847 & A rare bird species critically affected by habitat loss \\ \hline
27.2912 & A retired naval officer with experience in aircraft carrier operations and a deep knowledge of US Navy history and traditions. \\ \hline
    \hline
20.3037 & A healthcare blogger who spreads misinformation about vaccines and challenges the nurse's beliefs \\ \hline
20.7256 & A curious toddler who eagerly explores and enjoys playing with the DIY toys \\ \hline
21.2932 & A startup CTO who emphasizes rapid deployment and market responsiveness over meticulous code craftsmanship \\ \hline
21.6368 & An elderly French woman who shares stories of her youth in Paris and offers traditional baking tips. \\ \hline
21.6622 & A young child who laughs uncontrollably at the street performer's antics \\ \hline
    \end{tabularx}
    \caption{TrueSkill for LLaMA 4 - PI2 - Run 1}
    \label{tab:elo_personas_l4_pi_2_1}
\end{table}

\begin{table}[h]
    \footnotesize
    \centering
    \begin{tabularx}{\columnwidth}{|l|X|}
    \hline
    \rowcolor{lightgray} \textbf{Rating} & \textbf{Persona Description} \\ 
    \hline
30.2238 & A competitive collegiate football player always seeking for custom-designed team merchandise \\ \hline
29.2156 & A nanny who works with the widower to provide additional support and care for the children \\ \hline
28.2542 & A teenager struggling with anxiety and looking for coping mechanisms \\ \hline
27.9331 & A wargaming enthusiast who designs and simulates historical military scenarios to explore the decision-making processes, strategies, and outcomes of various engagements, including the Capture of Guam. \\ \hline
27.9087 & A retired naval officer with experience in aircraft carrier operations and a deep knowledge of US Navy history and traditions. \\ \hline
    \hline
19.4226 & A curious toddler who eagerly explores and enjoys playing with the DIY toys \\ \hline
20.1626 & A 7-year-old child diagnosed with autism spectrum disorder, seeking support in managing their social behaviors \\ \hline
21.2523 & A government agency using GIS analysis to plan efficient land use and infrastructure development \\ \hline
21.5645 & A startup CTO who emphasizes rapid deployment and market responsiveness over meticulous code craftsmanship \\ \hline
21.6047 & A healthcare blogger who spreads misinformation about vaccines and challenges the nurse's beliefs \\ \hline
    \end{tabularx}
    \caption{TrueSkill for LLaMA 4 - PI2 - Run 2}
    \label{tab:elo_personas_l4_pi_2_2}
\end{table}

\begin{table}[h]
    \footnotesize
    \centering
    \begin{tabularx}{\columnwidth}{|l|X|}
    \hline
    \rowcolor{lightgray} \textbf{Rating} & \textbf{Persona Description} \\ 
    \hline
29.8093 & An elderly relative who relies on the journalist's explanations to stay up-to-date on the latest technology trends. \\ \hline
27.7350 & A teenager struggling with anxiety and looking for coping mechanisms \\ \hline
27.6069 & A grassroots activist advocating for defunding the police and investing in alternative community-based solutions \\ \hline
27.5729 & A retired intelligence officer who had previously worked for the CIA. \\ \hline
27.3498 & A 7-year-old child diagnosed with autism spectrum disorder, seeking support in managing their social behaviors \\ \hline
    \hline
21.0631 & A risk management consultant advising organizations on safety measures in high-risk regions, including Afghanistan. \\ \hline
21.3249 & A confused person who is not familiar with iGEM and Synthetic Biology \\ \hline
21.4664 & An elderly woman who relies on the apothecary for her herbal remedies and trusts their expertise. \\ \hline
21.8820 & A genealogist studying the origins and variations of Arabic-language surnames, including Al-Marri and its related surnames such as Marri and Al Murrah. \\ \hline
22.1924 & A graphic designer seeking advice on how to prevent repetitive strain injuries \\ \hline
    \end{tabularx}
    \caption{TrueSkill for Mistral Small - DH1 - Run 1}
    \label{tab:elo_personas_ms_dh_1_1}
\end{table}

\begin{table}[h]
    \footnotesize
    \centering
    \begin{tabularx}{\columnwidth}{|l|X|}
    \hline
    \rowcolor{lightgray} \textbf{Rating} & \textbf{Persona Description} \\ 
    \hline
29.3532 & A teenager struggling with anxiety and looking for coping mechanisms \\ \hline
28.5395 & A military strategist in the Department of Defense, interested in the use of real-time satellite data to support decision-making and improve mission effectiveness. \\ \hline
28.5144 & An elderly relative who relies on the journalist's explanations to stay up-to-date on the latest technology trends. \\ \hline
28.4752 & A 7-year-old child diagnosed with autism spectrum disorder, seeking support in managing their social behaviors \\ \hline
28.0034 & A young child fascinated by Disney movies and loves to hear stories from their collection \\ \hline
    \hline
20.0316 & A risk management consultant advising organizations on safety measures in high-risk regions, including Afghanistan. \\ \hline
20.0726 & A person who struggles with Discardia – a fear of throwing things away. \\ \hline
22.0895 & A young child who laughs uncontrollably at the street performer's antics \\ \hline
22.1544 & A startup CTO who emphasizes rapid deployment and market responsiveness over meticulous code craftsmanship \\ \hline
22.2729 & A healthcare blogger who spreads misinformation about vaccines and challenges the nurse's beliefs \\ \hline
    \end{tabularx}
    \caption{TrueSkill for Mistral Small - DH1 - Run 2}
    \label{tab:elo_personas_ms_dh_1_2}
\end{table}

\begin{table}[h]
    \footnotesize
    \centering
    \begin{tabularx}{\columnwidth}{|l|X|}
    \hline
    \rowcolor{lightgray} \textbf{Rating} & \textbf{Persona Description} \\ 
    \hline
30.7565 & A seasoned military strategist sharing stories from their covert operations and offering guidance in a fast-paced world of intelligence \\ \hline
29.9239 & A military historian specializing in the Napoleonic Wars, with a focus on the Peninsular Campaign and the strategies employed by both the Allied and French forces. \\ \hline
28.7850 & A nanny who works with the widower to provide additional support and care for the children \\ \hline
28.3912 & A government agency using GIS analysis to plan efficient land use and infrastructure development \\ \hline
28.1505 & A retired Navy veteran who now works as a military consultant and shares practical knowledge \\ \hline
    \hline
19.6633 & A young child who laughs uncontrollably at the street performer's antics \\ \hline
20.8942 & A genealogist researching family histories connected to Biddeford, Maine. \\ \hline
21.5948 & A curious toddler who eagerly explores and enjoys playing with the DIY toys \\ \hline
21.6085 & A young child who comes to the shop every day with their parents to buy Kinder Eggs \\ \hline
21.8016 & A struggling high school student who has no interest in biology. \\ \hline
    \end{tabularx}
    \caption{TrueSkill for Mistral Small - PI1 - Run 1}
    \label{tab:elo_personas_ms_pi_1_1}
\end{table}

\begin{table}[h]
    \footnotesize
    \centering
    \begin{tabularx}{\columnwidth}{|l|X|}
    \hline
    \rowcolor{lightgray} \textbf{Rating} & \textbf{Persona Description} \\ 
    \hline
28.9724 & Another military strategist from a rival nation, constantly attempting to outwit and outmaneuver \\ \hline
28.1851 & A government agency using GIS analysis to plan efficient land use and infrastructure development \\ \hline
28.1572 & A wargaming enthusiast who enjoys designing and playing strategic simulations of historical military conflicts, particularly those involving Chinese forces. \\ \hline
28.0173 & A recently promoted Major General in the Nigerian Army \\ \hline
27.4219 & A wargaming enthusiast who designs and simulates historical military scenarios to explore the decision-making processes, strategies, and outcomes of various engagements, including the Capture of Guam. \\ \hline
    \hline
18.5387 & A young child who laughs uncontrollably at the street performer's antics \\ \hline
19.4294 & A struggling high school student who has no interest in biology. \\ \hline
19.7970 & A curious toddler who eagerly explores and enjoys playing with the DIY toys \\ \hline
22.2818 & A genealogist researching family histories connected to Biddeford, Maine. \\ \hline
22.3399 & A young child who comes to the shop every day with their parents to buy Kinder Eggs \\ \hline
    \end{tabularx}
    \caption{TrueSkill for Mistral Small - PI1 - Run 2}
    \label{tab:elo_personas_ms_pi_1_2}
\end{table}

\begin{table}[h]
    \footnotesize
    \centering
    \begin{tabularx}{\columnwidth}{|l|X|}
    \hline
    \rowcolor{lightgray} \textbf{Rating} & \textbf{Persona Description} \\ 
    \hline
31.4374 & A nanny who works with the widower to provide additional support and care for the children \\ \hline
29.8648 & A struggling high school student who has no interest in biology. \\ \hline
28.6752 & A teenager struggling with anxiety and looking for coping mechanisms \\ \hline
28.5229 & A retired military general with extensive experience in national defense, providing insights on border security strategies. \\ \hline
28.3828 & A wargaming enthusiast who enjoys designing and playing strategic simulations of historical military conflicts, particularly those involving Chinese forces. \\ \hline
    \hline
17.7759 & a confused person who is not familiar with iGEM and Synthetic Biology \\ \hline
17.8619 & A young child who laughs uncontrollably at the street performer's antics \\ \hline
18.0121 & A person who struggles with Discardia – a fear of throwing things away. \\ \hline
18.8108 & A literary critic who dismisses the Madam Tulip series as shallow and predictable. \\ \hline
20.9075 & A genealogist researching family histories connected to Biddeford, Maine. \\ \hline
    \end{tabularx}
    \caption{TrueSkill for Mistral Small - DH2 - Run 1}
    \label{tab:elo_personas_ms_dh_2_1}
\end{table}

\begin{table}[h]
    \footnotesize
    \centering
    \begin{tabularx}{\columnwidth}{|l|X|}
    \hline
    \rowcolor{lightgray} \textbf{Rating} & \textbf{Persona Description} \\ 
    \hline
29.0947 & An elderly relative who relies on the journalist's explanations to stay up-to-date on the latest technology trends. \\ \hline
28.8704 & A 7-year-old child diagnosed with autism spectrum disorder, seeking support in managing their social behaviors \\ \hline
28.5627 & A young child who comes to the shop every day with their parents to buy Kinder Eggs \\ \hline
28.1359 & A retired military general with extensive experience in national defense, providing insights on border security strategies. \\ \hline
27.6784 & A nanny who works with the widower to provide additional support and care for the children \\ \hline
    \hline
15.6622 & A person who struggles with Discardia – a fear of throwing things away. \\ \hline
17.5323 & A young child who laughs uncontrollably at the street performer's antics \\ \hline
17.8052 & a confused person who is not familiar with iGEM and Synthetic Biology \\ \hline
18.3785 & A literary critic who dismisses the Madam Tulip series as shallow and predictable. \\ \hline
20.2392 & A risk management consultant advising organizations on safety measures in high-risk regions, including Afghanistan. \\ \hline
    \end{tabularx}
    \caption{TrueSkill for Mistral Small - DH2 - Run 2}
    \label{tab:elo_personas_ms_dh_2_2}
\end{table}

\begin{table}[h]
    \footnotesize
    \centering
    \begin{tabularx}{\columnwidth}{|l|X|}
    \hline
    \rowcolor{lightgray} \textbf{Rating} & \textbf{Persona Description} \\ 
    \hline
30.0935 & A seasoned military strategist sharing stories from their covert operations and offering guidance in a fast-paced world of intelligence \\ \hline
29.1228 & A wargaming enthusiast who enjoys designing and playing strategic simulations of historical military conflicts, particularly those involving Chinese forces. \\ \hline
29.0169 & A geopolitical strategist who often appears on different networks presenting an alternative viewpoint on policies and events \\ \hline
28.1472 & A wargaming enthusiast who designs and simulates historical military scenarios to explore the decision-making processes, strategies, and outcomes of various engagements, including the Capture of Guam. \\ \hline
27.3938 & A government agency using GIS analysis to plan efficient land use and infrastructure development \\ \hline
    \hline
20.5554 & A young child who laughs uncontrollably at the street performer's antics \\ \hline
20.5604 & A genealogist researching family histories connected to Biddeford, Maine. \\ \hline
21.3149 & A curious toddler who eagerly explores and enjoys playing with the DIY toys \\ \hline
21.3640 & A young child who comes to the shop every day with their parents to buy Kinder Eggs \\ \hline
22.0657 & A literary critic who dismisses the Madam Tulip series as shallow and predictable. \\ \hline
    \end{tabularx}
    \caption{TrueSkill for Mistral Small - PI2 - Run 1}
    \label{tab:elo_personas_ms_pi_2_1}
\end{table}

\begin{table}[h]
    \footnotesize
    \centering
    \begin{tabularx}{\columnwidth}{|l|X|}
    \hline
    \rowcolor{lightgray} \textbf{Rating} & \textbf{Persona Description} \\ 
    \hline
28.0526 & A 7-year-old child diagnosed with autism spectrum disorder, seeking support in managing their social behaviors \\ \hline
28.0013 & A competitive collegiate football player always seeking for custom-designed team merchandise \\ \hline
27.7001 & A military historian specializing in the Napoleonic Wars, with a focus on the Peninsular Campaign and the strategies employed by both the Allied and French forces. \\ \hline
27.6578 & An Indian military personnel who served his country for 30 years and received the 50th Independence Anniversary Medal. \\ \hline
27.4933 & A wargaming enthusiast who designs and simulates historical military scenarios to explore the decision-making processes, strategies, and outcomes of various engagements, including the Capture of Guam. \\ \hline
    \hline
20.2685 & A young child who laughs uncontrollably at the street performer's antics \\ \hline
21.0522 & A young child who comes to the shop every day with their parents to buy Kinder Eggs \\ \hline
21.1988 & A healthcare blogger who spreads misinformation about vaccines and challenges the nurse's beliefs \\ \hline
21.8348 & a confused person who is not familiar with iGEM and Synthetic Biology \\ \hline
22.1529 & A struggling high school student who has no interest in biology. \\ \hline
    \end{tabularx}
    \caption{TrueSkill for Mistral Small - PI2 - Run 2}
    \label{tab:elo_personas_ms_pi_2_2}
\end{table}

\clearpage

\subsection{Opposite Value Consistency}
    As discussed in Table \ref{tbl:modelOppositeHeuristicDiffs}, the direct heuristic method almost always results in values that are significantly higher than those generated by the inventory prompts.
    Relevant Tables: \ref{tbl:heuristicsCombined_gpt4_1} - \ref{tbl:heuristicsCombined_l4_2}.

\begin{table}[H]
\footnotesize
    \centering
    \begin{tabular}{|c|c|c|c|}
    \hline
    \rowcolor{lightgray}\textbf{Phase} & \textbf{Categories} & \textbf{DH avg.} & \textbf{PI avg.} \\ \hline
        1 & PTM, PTL   & 5.02  & \textbf{7.63} \\ \hline
                & PUM, PUL   & 4.45  & \textbf{6.08} \\ \hline
                & PCM, PCL   & 6.32 & \textbf{19.35} \\ \hline
                & ETE, ETN   & 6.84 & \textbf{10.84} \\ \hline
                & EACM, EACL & 4.58 & \textbf{17.58} \\ \hline
        2 & PTM, PTL   & 3.26  & \textbf{8.46} \\ \hline
                & PUM, PUL   & 4.09  & \textbf{8.29} \\ \hline
                & PCM, PCL   & 4.04 & \textbf{51.91} \\ \hline
                & ONM, ONL   & 9.99 & \textbf{10.60} \\ \hline
                & ICD, ICS   & 2.01 & \textbf{16.63} \\ \hline
        3 & OBTM, OBTL & 3.10 & \textbf{18.43} \\ \hline
                & OBUM, OBUL & 3.49 & \textbf{14.77} \\ \hline
                & OBCM, OBCL & \textbf{4.64} & 3.28  \\ \hline
                & CNM, CNL   & 3.66 & \textbf{41.40} \\ \hline
                & M, L       & 2.73 & \textbf{21.57} \\ \hline
    \end{tabular}
\caption{Direct heuristics (DH avg.) and inventory heuristics (PI avg.) by heuristic pair and phase for the first heuristic set generated with GPT4.}
    \label{tbl:heuristicsCombined_gpt4_1}
\end{table}

\begin{table}[H]
\footnotesize
\centering
\begin{tabular}{|c|c|c|c|}
\hline
\rowcolor{lightgray} \textbf{Phase} & \textbf{Heuristic Pair} & \textbf{DH avg.} & \textbf{PI avg.} \\ \hline
0 & EACM-EACL & 31.51 & \textbf{11.60} \\ \hline
0 & ETE-ETN & 10.43 & \textbf{7.03} \\ \hline
0 & PCM-PCL & 32.27 & \textbf{4.20} \\ \hline
0 & PTM-PTL & 26.52 & \textbf{4.44} \\ \hline
0 & PUM-PUL & 28.87 & \textbf{3.83} \\ \hline
1 & ICD-ICS & 16.81 & \textbf{9.80} \\ \hline
1 & ONM-ONL & 31.07 & \textbf{13.61} \\ \hline
1 & PCM-PCL & 22.06 & \textbf{11.25} \\ \hline
1 & PTM-PTL & 24.36 & \textbf{3.66} \\ \hline
1 & PUM-PUL & 22.32 & \textbf{3.53} \\ \hline
2 & CNM-CNL & 5.74 & \textbf{12.35} \\ \hline
2 & M-L & \textbf{5.82} & 10.69 \\ \hline
2 & OBCM-OBCL & 16.99 & \textbf{3.50} \\ \hline
2 & OBTM-OBTL & 15.35 & \textbf{7.15} \\ \hline
2 & OBUM-OBUL & 9.26 & \textbf{6.08} \\ \hline
\end{tabular}
\caption{Direct heuristics (DH avg.) and inventory heuristics (PI avg.) by heuristic pair and phase for the first heuristic set generated with Mistral Small.}
\label{tbl:heuristicsCombined_ms_1}
\end{table}

\begin{table}[H]
\footnotesize
\centering
\begin{tabular}{|c|c|c|c|}
\hline
\rowcolor{lightgray} \textbf{Phase} & \textbf{Heuristic Pair} & \textbf{DH avg.} & \textbf{PI avg.} \\ \hline
0 & EACM-EACL & 27.89 & \textbf{11.48} \\ \hline
0 & ETE-ETN   & 11.82 & \textbf{7.37}  \\ \hline
0 & PCM-PCL   & 28.82 & \textbf{4.54}  \\ \hline
0 & PTM-PTL   & 28.45 & \textbf{4.50}  \\ \hline
0 & PUM-PUL   & 26.47 & \textbf{3.87}  \\ \hline
1 & ICD-ICS   & 20.75 & \textbf{10.27} \\ \hline
1 & ONM-ONL   & 47.20 & \textbf{12.23} \\ \hline
1 & PCM-PCL   & 25.98 & \textbf{11.10} \\ \hline
1 & PTM-PTL   & 25.82 & \textbf{3.54}  \\ \hline
1 & PUM-PUL   & 21.25 & \textbf{3.87}  \\ \hline
2 & CNM-CNL   & \textbf{10.32} & 13.66 \\ \hline
2 & M-L       & \textbf{8.10}  & 10.21 \\ \hline
2 & OBCM-OBCL & 18.57 & \textbf{3.65}  \\ \hline
2 & OBTM-OBTL & 7.94  & \textbf{7.49}  \\ \hline
2 & OBUM-OBUL & 8.20  & \textbf{6.19}  \\ \hline
\end{tabular}
\caption{Direct heuristics (DH avg.) and inventory heuristics (PI avg.) by heuristic pair and phase for the second heuristic set generated with Mistral Small.}
\label{tbl:heuristicsCombined_ms_2}
\end{table}

\begin{table}[H]
\footnotesize
\centering
\begin{tabular}{|c|c|c|c|}
\hline
\rowcolor{lightgray} \textbf{Phase} & \textbf{Heuristic Pair} & \textbf{DH avg.} & \textbf{PI avg.} \\ \hline
0 & EACM-EACL & 47.19 & \textbf{10.94} \\ \hline
0 & ETE-ETN & 15.04 & \textbf{8.65} \\ \hline
0 & PCM-PCL & 22.27 & \textbf{4.57} \\ \hline
0 & PTM-PTL & 41.73 & \textbf{5.14} \\ \hline
0 & PUM-PUL & 24.31 & \textbf{4.38} \\ \hline
1 & ICD-ICS & 19.54 & \textbf{9.73} \\ \hline
1 & ONM-ONL & 20.37 & \textbf{11.78} \\ \hline
1 & PCM-PCL & 34.46 & \textbf{12.93} \\ \hline
1 & PTM-PTL & 13.68 & \textbf{1.97} \\ \hline
1 & PUM-PUL & 21.50 & \textbf{2.72} \\ \hline
2 & CNM-CNL & \textbf{7.46} & 9.28 \\ \hline
2 & M-L & 25.24 & \textbf{10.02} \\ \hline
2 & OBCM-OBCL & 45.90 & \textbf{4.90} \\ \hline
2 & OBTM-OBTL & 26.08 & \textbf{4.06} \\ \hline
2 & OBUM-OBUL & 43.79 & \textbf{3.11} \\ \hline
\end{tabular}
\caption{Direct heuristics (DH avg.) and inventory heuristics (PI avg.) by heuristic pair and phase for the first heuristic set generated with LLaMA 3.}
\label{tbl:heuristicsCombined_l3_1}
\end{table}

\begin{table}[H]
\footnotesize
\centering
\begin{tabular}{|c|c|c|c|}
\hline
\rowcolor{lightgray} \textbf{Phase} & \textbf{Heuristic Pair} & \textbf{DH avg.} & \textbf{PI avg.} \\ \hline
0 & EACM-EACL & 50.77 & \textbf{10.52} \\ \hline
0 & ETE-ETN & 18.50 & \textbf{7.26} \\ \hline
0 & PCM-PCL & 21.96 & \textbf{4.74} \\ \hline
0 & PTM-PTL & 13.96 & \textbf{5.64} \\ \hline
0 & PUM-PUL & 29.62 & \textbf{4.79} \\ \hline
1 & ICD-ICS & 48.57 & \textbf{7.79} \\ \hline
1 & ONM-ONL & 18.75 & \textbf{11.88} \\ \hline
1 & PCM-PCL & 26.71 & \textbf{14.26} \\ \hline
1 & PTM-PTL & 12.53 & \textbf{1.86} \\ \hline
1 & PUM-PUL & 12.91 & \textbf{2.21} \\ \hline
2 & CNM-CNL & \textbf{3.35} & 11.37 \\ \hline
2 & M-L & 16.16 & 8.17 \\ \hline
2 & OBCM-OBCL & 29.11 & \textbf{3.82} \\ \hline
2 & OBTM-OBTL & 35.31 & \textbf{4.35} \\ \hline
2 & OBUM-OBUL & 40.81 & \textbf{3.43} \\ \hline
\end{tabular}
\caption{Direct heuristics (DH avg.) and inventory heuristics (PI avg.) by heuristic pair and phase for the second heuristic set generated with LLaMA 3.}
\label{tbl:heuristicsCombined_l3_2}
\end{table}

\begin{table}[H]
\footnotesize
\centering
\begin{tabular}{|c|c|c|c|}
\hline
\rowcolor{lightgray} \textbf{Phase} & \textbf{Heuristic Pair} & \textbf{DH avg.} & \textbf{PI avg.} \\ \hline
0 & EACM-EACL & 18.17 & \textbf{11.53} \\ \hline
0 & ETE-ETN & 44.36 & \textbf{7.08} \\ \hline
0 & PCM-PCL & 43.85 & \textbf{4.85} \\ \hline
0 & PTM-PTL & 23.07 & \textbf{2.96} \\ \hline
0 & PUM-PUL & 42.09 & \textbf{2.63} \\ \hline
1 & ICD-ICS & \textbf{2.05} & 5.51 \\ \hline
1 & ONM-ONL & 83.72 & \textbf{14.55} \\ \hline
1 & PCM-PCL & 27.81 & \textbf{14.07} \\ \hline
1 & PTM-PTL & 14.19 & \textbf{4.62} \\ \hline
1 & PUM-PUL & 15.94 & \textbf{2.41} \\ \hline
2 & CNM-CNL & 52.89 & \textbf{15.15} \\ \hline
2 & M-L & 30.02 & \textbf{12.77} \\ \hline
2 & OBCM-OBCL & 29.95 & \textbf{4.92} \\ \hline
2 & OBTM-OBTL & 34.30 & \textbf{4.72} \\ \hline
2 & OBUM-OBUL & 24.43 & \textbf{3.68} \\ \hline
\end{tabular}
\caption{Direct heuristics (DH avg.) and inventory heuristics (PI avg.) by heuristic pair and phase for the first heuristic set generated with LLaMA 4.}
\label{tbl:heuristicsCombined_l4_1}
\end{table}

\begin{table}[H]
\footnotesize
\centering
\begin{tabular}{|c|c|c|c|}
\hline
\rowcolor{lightgray} \textbf{Phase} & \textbf{Heuristic Pair} & \textbf{DH avg.} & \textbf{PI avg.} \\ \hline
0 & EACM-EACL & 16.34 & \textbf{9.95} \\ \hline
0 & ETE-ETN & 47.17 & \textbf{6.19} \\ \hline
0 & PCM-PCL & 25.00 & \textbf{5.11} \\ \hline
0 & PTM-PTL & 19.76 & \textbf{2.78} \\ \hline
0 & PUM-PUL & 30.67 & \textbf{2.49} \\ \hline
1 & ICD-ICS & \textbf{3.58} & 4.21 \\ \hline
1 & ONM-ONL & 86.18 & \textbf{12.95} \\ \hline
1 & PCM-PCL & 20.96 & \textbf{13.90} \\ \hline
1 & PTM-PTL & 19.83 & \textbf{3.88} \\ \hline
1 & PUM-PUL & 17.00 & \textbf{2.27} \\ \hline
2 & CNM-CNL & 52.39 & \textbf{13.15} \\ \hline
2 & M-L & 28.34 & \textbf{11.74} \\ \hline
2 & OBCM-OBCL & 21.32 & \textbf{3.99} \\ \hline
2 & OBTM-OBTL & 15.83 & \textbf{3.94} \\ \hline
2 & OBUM-OBUL & 21.14 & \textbf{3.47} \\ \hline
\end{tabular}
\caption{Direct heuristics (DH avg.) and inventory heuristics (PI avg.) by heuristic pair and phase for the second heuristic set generated with LLaMA 4.}
\label{tbl:heuristicsCombined_l4_2}
\end{table}

\clearpage

\end{document}